\documentclass[jair,twoside,11pt]{article}
\input epsf %% at top of file

\usepackage{jair}
\usepackage{theapa}

\jairheading{30}{2007}{273-320}{02/07}{10/07}
\ShortHeadings{Reasoning with Very Expressive Fuzzy Description
Logics}{Stoilos, Stamou, Pan, Tzouvaras \& Horrocks}

\firstpageno{273}

\usepackage{amsmath}
\usepackage{amssymb}

\newfont{\mymathtt}{cmtt10 scaled 1095}

\newcommand{\alc}{%                             ALC
  \ensuremath{\mathcal{ALC}}\xspace}
\newcommand{\alcn}{%                            ALCN
  \ensuremath{\mathcal{ALCN}\xspace}}
\newcommand{\s}{%                               S
        \ensuremath{\mathcal{S}}\xspace}

\newcommand{\si}{%                              SI
  \ensuremath{\mathcal{SI}}\xspace}

\newcommand{\shi}{%                             SHI
  \ensuremath{\mathcal{SHI}}\xspace}
\newcommand{\alch}{%                            ALCH_R+ (SH)
  \alchrp}

\newcommand{\shif}{\ensuremath{\mathcal{SHI}\mathbf{\textit{f}}}\xspace} %SHIf

\newcommand{\shin}{%                             SHIN
  \ensuremath{\mathcal{SHIN}}\xspace}

\newcommand{\shiq}{%                             SHIQ
  \ensuremath{\mathcal{SHIQ}}\xspace}

\newcommand{\PDL}%                              PDL
        {PDL\xspace}
\newcommand{\CPDL}%                             converse-PDL
        {\emph{converse}-PDL\xspace}

\newfont{\bigmathxx}{cmsy10 scaled 1440}
\newfont{\smallmathxx}{cmsy10 scaled 720}

\newcommand{\I}{%                               I (caligraphic)
        \ensuremath{\mathcal{I}}\xspace}

\newcommand{\ifunc}{%                           ^I (caligraphic)
        \ensuremath{^\mathcal{I}}\xspace}

\newcommand{\deltai}{%                          Delta^I
        \ensuremath{\Delta\ifunc}\xspace}

\newcommand{\Exptime}{\textsc{Exptime}\xspace}
\newcommand{\Nexptime}{\textsc{Nexptime}\xspace}
\newcommand{\Pspace}{\textsc{Pspace}\xspace}

\newcommand{\Concepts}{\ensuremath{\mathbf{C}}\xspace}

\newcommand{\Roles}{\ensuremath{\mathbf{R}}\xspace}

\newcommand{\Individuals}{\ensuremath{\mathbf{I}}\xspace}

\newcommand{\Inv}{\ensuremath{\mathop{\mathsf{Inv}}}\xspace}
\newcommand{\Tr}{\ensuremath{\mathop{\mathsf{Trans}}}\xspace}

\newcommand{\kb}{\ensuremath{\Sigma}\xspace}

\newcommand{\K}%                                Modal K
        {\ensuremath{\mathbf{K}}\xspace}
\newcommand{\KT}%                               Modal KT
        {\ensuremath{\mathbf{KT}}\xspace}
\newcommand{\Kfour}%                            Modal K4
        {\ensuremath{\mathbf{K4}}\xspace}
\newcommand{\Sfour}%                            Modal S4
        {\ensuremath{\mathbf{S4}}\xspace}
\newcommand{\Km}%                               Modal Km
        {\ensuremath{\K_{(\mathbf{m})}}\xspace}
\newcommand{\KTm}%                              Modal KTm
        {\ensuremath{\KT_{(\mathbf{m})}}\xspace}
\newcommand{\Kfourm}%                           Modal K4m
        {\ensuremath{\mathbf{K4}_{(\mathbf{m})}}\xspace}
\newcommand{\Sfourm}%                           Modal S4m
        {\ensuremath{\mathbf{S4}_{(\mathbf{m})}}\xspace}

\newcommand{\con}[1]{%                  a concept
        \ensuremath{\mathsf{#1}}}

\newcommand{\role}[1]{%                 a role
        \ensuremath{\mathsf{\textsl{$#1$}}}}

\newcommand{\indv}[1]{%                 an individual (as an instance)
        \ensuremath{\mbox{{\tt #1}}}}

\newcommand{\Top}{\ensuremath{\top}}
\newcommand{\Bot}{\ensuremath{\bot}\xspace}
\newcommand{\Or}{\ensuremath{\mathbin\sqcup}\xspace}
\newcommand{\Sub}{\ensuremath{\subseteq}\xspace}

\renewcommand{\And}{\ensuremath{\sqcap}\xspace}
\newcommand{\Int}[1]{\ensuremath{#1^{\mathcal{I}}}}
\newcommand{\Issub}{\ensuremath{\sqsubseteq}}

\newcommand{\Tail}{\mathop{\mathsf{Tail}}}

\newcommand{\Not}{\neg}
\newcommand{\some}[2]{%
  \ensuremath{\exists #1 . #2}}
\newcommand{\Some}{\some}
\newcommand{\all}[2]{%
  \ensuremath{\forall #1 . #2}}
\newcommand{\All}[2]{\all #1 #2}

\newcommand{\sss}{\ensuremath{{\mathrel{\kern.25em{\sqsubseteq}
\kern-.5em \mbox{{\scriptsize *}}\kern.25em}}}\xspace}
\newcommand{\nsss}{\ensuremath{{\mathrel{\kern.25em{\sqsubseteq}
\kern-.5em \mbox{{\scriptsize *}}\kern-.5em /\kern.3em}}}\xspace}

\newcommand{\R}{\ensuremath{\mathcal{R}}\xspace}

\newcommand{\ndoteq}{\ensuremath{\mathrel{\not\doteq}}}
\newcommand{\Paths}{\ensuremath{\mathsf{Paths}}\xspace}

\newcommand{\set}[1]{\{{#1}\}}

\RequirePackage{xspace}

\newcommand{\shoiq}{\ensuremath{\mathcal{SHOIQ}}\xspace}
\newcommand{\shoin}{\ensuremath{\mathcal{SHOIN}}\xspace}

\newcommand{\shoindp}{\ensuremath{\mathcal{SHOIN({\bf D^+})}}\xspace}
\newcommand{\shifdp}{\ensuremath{\mathcal{SHIF({\bf D^+})}}\xspace}

\newcommand{\alcf}{\ensuremath{\mathcal{ALCF}}\xspace}
\newcommand{\alcd}{\ensuremath{\mathcal{ALC(\Donly)}}\xspace}

\newcommand{\Nonly}{\ensuremath{\mathcal{N}}\xspace}
\newcommand{\Sonly}{\ensuremath{\mathcal{S}}\xspace}

\newcommand{\Donly}{\ensuremath{\mathcal{D}}\xspace}
\newcommand{\Oonly}{\ensuremath{\mathcal{O}}\xspace}

\newcommand{\Ionly}{\ensuremath{\mathcal{I}}\xspace}

\newcommand{\Aonly}{\ensuremath{\mathcal{A}}\xspace}
\newcommand{\A}{\Aonly}

\newcommand{\Tonly}{\ensuremath{\mathcal{T}}\xspace}

\newcommand{\Honly}{\ensuremath{\mathcal{H}}\xspace}

\newcommand{\T}{\Tonly}

\newcommand{\nlist}[1]{\ensuremath{#1_1,\dots,#1_n}}

\newcommand{\tup}[1]{\ensuremath{\langle #1\rangle}\xspace}

\newcommand{\Forest}{\ensuremath{\mathbf{\mathtt{F}}\xspace}}

\newcommand{\wrt}{\ensuremath{\textrm{w.r.t.}}\xspace}

\newcommand{\rcn}[3]{\ensuremath{#1_{#2,#3}}\xspace}
\newcommand{\fkdalc}{%                              f_{KD}-ALC
  \ensuremath{\mbox{f}_{KD}\mbox{-}\mathcal{ALC}}\xspace}
\newcommand{\fkdsi}{%                              f_{KD}-SI
  \ensuremath{\mbox{f}_{KD}\mbox{-}\mathcal{SI}}\xspace}
\newcommand{\fkdshi}{%                              f_{KD}-SI
  \ensuremath{\mbox{f}_{KD}\mbox{-}\mathcal{SHI}}\xspace}
\newcommand{\fkdshin}{%                              f_{KD}-SHIN
  \ensuremath{\mbox{f}_{KD}\mbox{-}\mathcal{SHIN}}\xspace}
\newcommand{\fkdshoin}{%                              f_{KD}-SHOIN
  \ensuremath{\mbox{f}_{KD}\mbox{-}\mathcal{SHOIN}}\xspace}
\newcommand{\fkdshoiq}{%                              f_{KD}-SHOIQ
  \ensuremath{\mbox{f}_{KD}\mbox{-}\mathcal{SHOIQ}}\xspace}

\newcommand{\falc}[1]{%                              f_{L}-ALC
  \ensuremath{\mbox{f}_{#1}\mbox{-}\mathcal{ALC}}\xspace}

\newcommand{\bS}{\textbf{S}\xspace}
\newcommand{\bR}{\textbf{R}\xspace}

\newcommand{\lan}{\ensuremath{\langle}\xspace}
\newcommand{\ran}{\ensuremath{\rangle}\xspace}
\newcommand{\cL}{\ensuremath{\mathcal{L}}\xspace}
\newcommand{\cE}{\ensuremath{\mathcal{E}}\xspace}

\newcommand{\cV}{\ensuremath{\mathcal{V}}\xspace}

\newcommand{\alcq}{\ensuremath{\mathcal{ALCQ}}\xspace}

\newcommand{\J}{\ensuremath{\mathcal{J}}\xspace}

\renewcommand{\Forest}{\ensuremath{\mathcal{F}_\A}\xspace}

\renewcommand{\bigcup}{\mathop{\mathop{\mbox{\bigmathxx\symbol{91}}}}\limits}
\newcommand{\undermax}{\mathop{\mathop{\mbox{max}}}\limits}

\newcommand{\fkd}{f$_{KD}$\xspace}

\newcommand{\bow}{\mathcal{\bowtie}\xspace}
\newcommand{\bowref}{\ensuremath{\mathcal{\bowtie^-}\xspace}}

\def\squareforqed{\hbox{\rlap{$\sqcap$}$\sqcup$}}
\def\qed{\ifmmode\squareforqed\else{\unskip\nobreak\hfil
\penalty50\hskip1em\null\nobreak\hfil\squareforqed
\parfillskip=0pt\finalhyphendemerits=0\endgraf}\fi}

\def\diae{\hfill{$\diamondsuit$}   % qed with full box
  \ifdim\lastskip<\medskipamount \removelastskip\penalty55\medskip\fi}

\newtheorem{theorem}{Theorem}[section]
\newtheorem{corollary}[theorem]{Corollary}
\newtheorem{lemma}[theorem]{Lemma}

\newtheorem{definition}[theorem]{Definition}
\newtheorem{example}[theorem]{Example}
\newtheorem{rem}[theorem]{Remark}

\renewcommand{\shif}{\ensuremath{\mathcal{SHIF}}\xspace}
\renewcommand{\alch}{\ensuremath{\mathcal{ALCH}}\xspace}

\begin{document}

%\title{Fuzzy Description Logics with Transitive and Inverse Roles}

\title{Reasoning with Very Expressive Fuzzy Description Logics}

%Jeff 29/07/05
%\author{Giorgos Stoilos\inst{1}, Giorgos Stamou\inst{1}, Jeff Z. Pan\inst{2}, Vassilis
%Tzouvaras\inst{1},\\ and Ian Horrocks\inst{2}}
%
%\institute{Department of Electrical and Computer Engineering,
%National Technical University of Athens, Zographou 15780,
%Greece\\[10pt] \and School of Computer Science, The University of Manchester\\
%Manchester, M13 9PL, UK}

\author{\name Giorgos Stoilos \email gstoil@image.ece.ntua.gr \\
       \name Giorgos Stamou \email gstam@softlab.ece.ntua.gr \\
       \addr Department of Electrical and Computer Engineering,\\
       National and Technical University of Athens,\\
       Zographou 15780, Athens, GR
       \AND
       \name Jeff Z. Pan \email jpan@csd.abdn.ac.uk \\
       \addr Department of Computing Science,\\
       The University of Aberdeen, UK\\
       \AND
       \name Vassilis Tzouvaras \email tzouvaras@image.ece.ntua.gr \\
       \addr Department of Electrical and Computer Engineering,\\
       National and Technical University of Athens,\\
       Zographou 15780, Athens, GR
       \AND
       \name Ian Horrocks \email horrocks@cs.man.ac.uk \\
       \addr School of Computer Science, The University of Manchester\\
       Manchester, M13 9PL, UK}

\maketitle

%Jeff 30/11/05
%George 22/09/05
\begin{abstract}
It is widely recognized today that the management of %uncertainty
imprecision and vagueness will yield more intelligent and
realistic knowledge-based applications. Description Logics (DLs)
are a family of knowledge representation languages that have
gained considerable attention the last decade, mainly due to their
decidability and the existence of empirically high performance of
reasoning algorithms. In this paper, we extend the well known
fuzzy \alc DL to the fuzzy \shin DL, which extends the fuzzy \alc
DL with transitive role axioms (\Sonly), inverse roles (\Ionly),
role hierarchies (\Honly) and number restrictions (\Nonly). We
illustrate why transitive role axioms are difficult to handle in
the presence of fuzzy interpretations and how to handle them
properly. Then we extend these results by adding role hierarchies
and finally number restrictions. The main contributions of the
paper are the decidability proof of the fuzzy DL languages
fuzzy-\si and fuzzy-\shin, as well as decision procedures for the
knowledge base satisfiability problem of the fuzzy-\si and
fuzzy-\shin.
\end{abstract}

\section{Introduction}\label{sec:intro}
%Jeff 30/11/05

Nowadays, many applications and domains  use some form of
knowledge representation language in order to improve their
capabilities. Encoding human knowledge and providing means to
reason with it can benefit applications a lot, by enabling them
%Stoilos 13/07/06 to deduce
provide intelligent answers to complex user defined tasks.
Examples of modern applications that have recently adopted
knowledge representation languages are the World Wide Web
\cite{Tim01,Baader02}, where knowledge is used to improve the %George 22/09/05
abilities of agents and the interoperability between disparate
systems, multimedia processing applications
\cite{Alejandro03,Benitez00}, which use knowledge in order to
bridge the ``gap" between human perception of the objects that
exist within multimedia documents, and computer ``perception" of
pixel values, configuration applications \cite{McGuiness03a}, etc.
% and many more.
 Unfortunately,
%it is often the case
there are occasions where traditional knowledge representation
languages fail to accurately represent the concepts that appear in
a domain of interest.
%This is mainly due to the
%fact that knowledge is quite often
For example, this is particularly the case when domain knowledge
is inherently imprecise or vague.
%Examples of such
Concepts like that of a ``near" destination \cite{Tim01}, a
``highQuality" audio system \cite{McGuiness03a}, ``many" children,
a ``faulty" reactor \cite{Horrocks99},
%George 13/07/05 corrected "soom" to "soon"
``soon" and many more, require special modelling features. In the
past many applications of various research areas, like decision
making, image processing, robotics and medical diagnosis have
adopted special mathematical frameworks that are intended for
modelling such types of concepts
\cite{Zimmermann87,Larsen93,Krishnapuram92}. One such a
mathematical framework is \emph{fuzzy set theory} \cite{Zadeh65}.
Though fuzzy extensions of various logical formalisms, like
propositional, predicate or modal logics have been investigated in
the past \cite{Hajek98}, such a framework is %less popular in
%formalisms like
is not yet well developed for Description Logics and much research
work needs to be done. More precisely, there is the need for
reasoning in very expressive fuzzy Description Logics.

In order to achieve knowledge reusability and high
interoperability, modern applications
%like the Semantic Web, multimedia
%knowledge representation or configuration
often use the concept of an ``ontology" \cite{Tim01} to represent
the knowledge that exists within their domain. Ontologies are
created by encoding the full knowledge we possess for a specific
entity of our world using a knowledge representation language. A
logical formalism that has gained considerable attention the last
decade is Description Logics \cite{Baader03a}. Description Logics
(DLs) are a family of class-based (concept-based) knowledge
representation formalisms, equipped with well defined
model-theoretic semantics~\cite{Tars56}. They are characterized by
the use of
%George 13/07/05 changed class to concept
various constructors to build complex concept descriptions from
simpler ones, an emphasis on the decidability of key reasoning
problems, and by the provision of sound, complete and empirically
tractable reasoning services. Both the well-defined semantics and
the powerful reasoning tools that exist for Description Logics
makes them ideal for encoding knowledge in many applications like
the Semantic Web \cite{Baader02,Pan04}, multimedia applications
\cite{Meghini01}, medical applications \cite{Rector97}, databases
\cite{Calvanese98} and many more.
%For example widely used
Interestingly, the current standard for Semantic Web ontology
languages, OWL \cite{Bechhofer04}, is based on
%highly expressive
Description Logics %\footnote{Without regarding annotation
%properties of OWL, they are equivalent to the \shoindp and \shiq
%DLs \cite{Horrocks03b}.}
to represent knowledge and support a wide range of reasoning
services. More precisely, without regarding annotation properties
of OWL, the OWL Lite species of OWL is equivalent to the \shifdp
DL, while OWL DL is equivalent to \shoindp \cite{Horrocks03c}.
Although DLs provide considerable expressive power, they feature
%expressive
limitations regarding their ability to represent vague (fuzzy)
knowledge. As obvious, in order to make applications that use DLs
able to cope with such
%and uncertain
information we have to extend them with a theory capable of
representing such kind of information. One such important theory
is fuzzy set theory. Fuzzy Description Logics are very interesting
logical formalisms as they can be used in numerous domains like
multimedia and information retrieval \cite{Fagin98,Meghini01} to
provide ranking degrees, geospatial \cite{Chen05} to cope with
vague concepts like ``near", ``far" and many more.

In order to make the need to handle vagueness knowledge more
evident and the application of fuzzy set theory more intuitive,
let us consider an example. Suppose that we are creating a
knowledge-based image processing application. In such application
the task is to (semi)automatically detect and recognize image
objects. Suppose also that the content of the images represents
humans or animals. For such a domain one can use standard features
of Description Logics to encode knowledge. For example, a
knowledge base describing human bodies could contain the following
entities

\[\mathsf{Arm}\sqsubseteq \exists\mathsf{isPartOf}.\con{Body}\]
\[\mathsf{Body}\sqsubseteq \exists\mathsf{isPartOf.Human}\]

\noindent where $\sqsubseteq$ is a subsumption relation and
$\mathsf{isPartOf}$ is obviously a transitive relation. This
knowledge can be captured with the aid of the \s DL
\cite{Sattler96}. Moreover, one might want to capture the
knowledge that the role $\mathsf{hasPart}$ is the inverse of the
role $\mathsf{isPartOf}$, writing $\mathsf{hasPart}:=
\mathsf{isPartOf^-}$, thus being able to state that something that
is a body and has a tail is also an animal as,

\[\mathsf{Body}\sqcap \exists\mathsf{hasPart.Tail}\sqsubseteq \mathsf{Animal}.\]

\noindent For this new feature one would require the \si DL
\cite{Horrocks99}. The new axiom gives us the ability to recognize
that the concept $\mathsf{Arm}\sqcap \mathsf{Tail}$ is subsumed by
$\exists \mathsf{isPartOf.Animal}$. Finally, the \si DL can be
further extended with role hierarchies and number restrictions.
Hence, one is able to capture the fact that the role
$\con{hasDirectPart}$ is a sub-role of the role $\con{hasPart}$,
by writing $\con{isDirectPartOf}\sqsubseteq \con{isPartOf}$, while
we can also provide a more accurate definition of the concept
$\con{Body}$ by giving the axiom,

\[\mathsf{Body}\sqsubseteq \exists\mathsf{isDirectPartOf.Human}\sqcap \leq 2\con{hasArm}\sqcap\geq 2\con{hasArm}\]

\noindent stating that the body is a direct part of a human and it
also has exactly two arms.

Up to now we have only used standard Description Logic features.
Now suppose that we run an image analysis algorithm. Such
algorithms usually segment the image into regions and try to
annotate them with appropriate semantic labels using low level
image features. This process involves a number of vague concepts
since an image region might be red, blue, circular, small or
smooth textured to some degree or two image regions might not be
totally but only to some degree adjacent (since not all of their
pixels are adjacent), one contained within the other, etc. Hence
we can only decide about the membership of a region to a specific
concept only to a certain degree \cite{Athanasiadis07}. For
example, in our case we could have that the object $\indv o_1$
$\mathsf{isPartOf}$ the object $\indv o_2$ to a degree of $0.8$,
that $\indv o_2$ $\mathsf{isPartOf}$ $\indv o_3$ to a degree of
$0.9$, that $\indv o_1$ is an $\mathsf{Arm}$ to a degree of $0.75$
and that $\indv o_2$ is a $\mathsf{Body}$ to a degree of 0.85.
From that fuzzy knowledge one could deduce that $\indv o_3$
belongs to the concept $\exists \mathsf{hasPart}.
\mathsf{Body}\sqcap \exists \mathsf{hasPart}.\mathsf{Arm}$ to a
degree of 0.75. This together with a definition of the form
$\mathsf{Human}\equiv \exists \mathsf{hasPart}.
\mathsf{Body}\sqcap \exists \mathsf{hasPart}.\mathsf{Arm}$, where
$\equiv$ represents equivalence, means that there is a good chance
that $\indv o_3$ is a $\mathsf{Human}$. Observe, that in this
definition, in order for someone to be a human, we do not force a
$\mathsf{Body}$ to explicitly have a part that is an
$\mathsf{Arm}$. This is a reasonable choice in the present
application, because depending on the \emph{level} of the
segmentation, there might be several segmented regions between
$\indv o_2$ and $\indv o_3$. As it is obvious is such applications
handling the inherent vagueness certainly benefits the specific
application.

%Jeff 30/11/05
In this paper we extend the well known fuzzy \alc (f-\alc)
DL~\cite{Straccia01} to the fuzzy \shin DL (f-\shin), which
extends the f-\alc DL
%present the extension of the description
%logic \si with fuzzy set theory. \si extends the well-known \alc
%DL \cite{Schmidt},
with the \emph{inverse} role constructor, \emph{transitive} role
axioms, \emph{role hierarchies} and the \emph{number restrictions}
constructor. Moreover, we prove the decidability of the f-\shin DL
by providing a tableaux algorithm for deciding the standard DL
inference problems. In order to provide such an algorithm we
proceed in two steps. First, we focus on the f-\si language
studying the properties of fuzzy transitive roles in value and
existential restrictions, as well as the applicability of the
techniques used in the classical \si language to ensure the
termination of the algorithm \cite{Horrocks99}. As we will see
there is great difficulty on handling such axioms on the context
of fuzzy DLs, but after finishing our investigation we will see
that similar notions as in classical \si language can be applied.
Secondly, we extend these results by adding role hierarchies and
number restrictions. We provide all the necessary extensions to
the reasoning algorithm of f-\si, thus providing a reasoning
algorithm for the f-\shin language. Discarding datatypes, \shin is
slightly more expressive than \shif (OWL Lite) and slightly less
expressive than \shoin (OWL DL). In order to achieve our goal we
again extend the techniques used for the classical \shin language
and which ensure correctness of the algorithm
\cite{Horrocks99,Horrocks00}. Finally, we prove the decidability
of the extended algorithm. There are many benefits on following
such an approach. On the one hand we provide a gradual
presentation to the very complex algorithm of f-\shin, while on
the other hand we provide a reasoning algorithm for a less
expressive, but more efficient fuzzy DL language, f-\si. The
classical \si language is known to be \Pspace-complete, in
contrast to the \Exptime-completeness of \shin \cite{Tobies01},
hence our algorithm for f-\si can be used for future research and
for providing efficient and optimized implementations.
%It has been argued
%\cite{Horrocks99} that these two features are crucial for many
%applications that use DLs.
%\si is a sub-language of the
%more highly expressive \shifdp and \shoindp DLs, which  are used
%as the underpinnings of the OWL-Lite and OWL DL ontology
%languages, respectively.
%As we will see, when it comes to reasoning with fuzzy description
%logics, there are many points that become even harder to deal
%with, compared to classical DLs. More precisely, the techniques
%used for termination of the reasoning algorithm \cite{Horrocks99}
%had to be proved to be applicable in the new framework, as well as
%the properties of the semantics of the fuzzy DL, when transitive
%and inverse fuzzy relations are present, had to be investigated
%thoroughly, in order to provide tractable decision procedures
%\cite{Vardi97}.

Please note that fuzzy DLs \cite{Straccia01} are complementary to
other approaches that extend DLs, like \emph{probabilistic} DLs
\cite{Koller97,Giugno02,Ding04}, or \emph{possibilistic} DLs
\cite{Hollunder94}. More precisely, these theories are meant to be
used for capturing different types of \emph{imperfect} information
and knowledge. Fuzziness is purposed for capturing vague (fuzzy)
knowledge, i.e. facts that are certain but which have degrees of
truth assigned to them, like for example the degree of truth of
someone being tall. On the other hand, possibilistic and
probabilistic logics are purposed for capturing cases where
knowledge is uncertain due to lack of information or knowledge
about a specific situation or a future event, like for example a
sensor reading or a weather forecast. These facts are assigned
degrees of possibility, belief or probability, rather than truth
degrees. \citeA{Dubois01} provide a comprehensive analysis on
these theories along with their different properties.

%George 13/07/05 Various changes. We had a new Section (3) to describe.
The rest of the paper is organized as follows. Section
\ref{sec:prelim} briefly introduces the DL \shin and provides some
preliminaries about the notion of a fuzzy set and how set
theoretic and logical operations have been extended to the fuzzy
set framework. Section \ref{sec:f-SI} introduces the syntax and
semantics of the fuzzy \shin DL, which we call
f-\shin.\footnote{In a previous approach to fuzzy DLs
%%%%%%%%%%%
%George 13/07/05 Footnote moved earlier!!
%George 21/06/05
the notation $f$\hspace*{-3pt}\alc is used \cite{Straccia04}, but
this notation is not so flexible to represent fuzzy DLs which use
different norm operations, as we will see later on. In some other
approaches \cite{Tresp98,Holldobler02} the naming $\alc_F$ is used
but this can easily be confused with \alcf (\alc
%Jeff 18/06/05
% plus
extended with \emph{functional restrictions}, \citeR{Horrocks99}),
when pronounced.} language. Section \ref{sec:transinvest} provides
an investigation on the semantics of fuzzy DLs when fuzzy
transitive relations participate in value and existential
restrictions.
%\cite{Stoilos05,Stoilos05a,Stoilos05d,Stoilos05e,Pan06a}
%Jeff 29/07/05
%In section \ref{blockserving} we examine the applicability of
%special techniques used for reasoning in the classical \si
%language in the context of f-\si.
In section \ref{sec:sireasoning} we give a detailed presentation
of the reasoning algorithm for deciding the
%George 13/07/05 satisfiability->consistency
consistency of a fuzzy-\si ABox and we provide the proofs for the
termination, soundness and completeness of the procedure. Then, in
section \ref{sec:f-SHIN} we extend the previous results by adding
role hierarchies and number restrictions. More precisely, the
results of section \ref{sec:transinvest} are enriched by
considering transitive roles and roles hierarchies in value and
existential restrictions. Using this results we extend the
algorithm of section \ref{sec:sireasoning} to handle with these
new feature and finally we prove its soundness, completeness and
termination. %In section \ref{sec:reduction} we study a reduction
%technique from fuzzy DL languages to crisp DLs, proposed in the
%literature, in the case of f-\si, and
At last, in section \ref{sec:relwork} we review some previous work
on fuzzy Description Logics while section \ref{sec:conclution}
concludes the paper.

\section{Preliminaries}\label{sec:prelim}
In the current section we will briefly introduce classical DLs and
fuzzy set theory, recalling some mathematical properties of fuzzy
set theoretic operators.
%Jeff 18/06/05: we need some background for DL, in particular \si, and
%because we extend \si to f-\si, it is natural to introduce \si first.
\subsection{Description Logics and the \shin DL}
\label{sec:dl}

Description Logics (DLs)~\cite{Baader03a} are a family of
logic-based knowledge representation formalisms designed to
represent and reason about the knowledge of an application domain
in a structured and well-understood way. They are based on a
common family of languages, called \emph{description languages},
which provide a set of constructors to build   concept (class) and
role (property) descriptions. Such descriptions can be used in
axioms and assertions of DL knowledge bases  and can be reasoned
about with respect to (\wrt) DL knowledge bases by DL systems.

%Jeff 30/11/05 remove \Cexp etc
In this section, we will briefly introduce the \shin DL, which
will be extended to the f-\shin DL later. A description language
consists of an alphabet of distinct concept names (\Concepts),
role names (\Roles) and individual (object) names (\Individuals);
together with a set of constructors to construct concept and role
descriptions.
%For the \si DL, we call \Cexp \si
%and \Rexp \si the set of concepts and roles of \si, respectively.

%Jeff 30/11/05 explain descriptions
Now we define the notions of  \shin-roles and \shin-concepts.

\begin{definition}
Let $\role{RN}\in\Roles$ be a role name and $R$ a \shin-role.
\shin-role descriptions (or simply \shin-roles) are defined by the
abstract syntax: $S::=\role{RN}\mid R^-$. The inverse relation of
roles is symmetric, and to avoid considering
%George 13/07/05
%the
roles such as $R^{--}$, we define a function \Inv which returns
the inverse of a role, more precisely,
\begin{eqnarray*}
\Inv(R):=\left\{
\begin{array}{ll}
\role{RN}^- & \mbox{if $R=\role{RN}$,}\\
\role{RN}& \mbox{if $R=\role{RN}^-$.}\\
\end{array}\right.
\end{eqnarray*}
 % \shin-concepts are defined by the abstract syntax:
%$$C: : = \top\,\mid\,\bot\,\mid\,\con{CN}\,\mid\,\neg C\,\mid\,
%C\sqcap D\,\mid\,C\sqcup D\,\mid\,\some R C \,\mid\,\all{R}{C}
%$$%
%The set of \fkdshin-roles is $\bR\cup\{R^-|R\in\bR\}$.
The set of \shin-concept descriptions (or simply \shin-concepts)
is the smallest set such that:
\begin{enumerate}
    \item every concept name $\con{CN}\in \Concepts$ is a
    \shin-concept,\vspace{0.0cm}

    \item if $C$ and $D$ are \shin-concepts and $R$ is a
    \shin-role, then  $\neg C$, $C\sqcup D$, $C\sqcap D$, $\forall
    R.C$ and $\exists R.C$ are also \shin-concepts, called
    \emph{general negation} (or simply \emph{negation}),
    \emph{disjunction}, \emph{conjunction}, \emph{value
    restrictions} and \emph{existential restriction},
    respectively, and\vspace{0.0cm}

    \item if $S$ a \emph{simple}\footnote{A role is called \emph{simple}
    if it is neither transitive nor has any transitive sub-roles.
    This is crucial in order to get a decidable logic
    \cite{Horrocks99j}.} \shin-role and $n\in \mathbb{N}$, then
    $(\geq nS)$ and $(\leq nS)$ are also \shin-concepts, called
    \emph{at-most} and \emph{at-least} number restrictions.

\end{enumerate}
\end{definition}
By removing point 3 of the above definition we obtain the set of
\si-concepts.

Description Logics have a model-theoretic semantics, which is
defined in terms of interpretations. An \emph{interpretation}
(written as \I) consists of a \emph{domain} (written as \deltai)
and an \emph{interpretation function} (written as $\cdot\ifunc$),
where the domain is a nonempty set of objects and the
interpretation function maps each individual name
$\indv{a}\in\Individuals$ to an element $\indv{a}\ifunc \in
\deltai$, each concept name $\con{CN}\in\Concepts$ to a subset
$\con{CN} \ifunc \subseteq \Delta \ifunc$, and each role name
$\role{RN}\in\Roles$ to a binary relation $\role{RN} \ifunc
\subseteq \deltai \times \deltai$. The interpretation function can
be extended to give semantics to concept and role descriptions.
These are given in Table~\ref{table:si-sem}.

\begin{table}[t]
\begin{center}
\begin{tabular}{|l|c|l|}
\hline
\hspace*{18pt} Constructor & \hspace*{8pt}Syntax   &  \hspace*{55pt}   Semantics     \\
\hline  top & \Top & \deltai\\
\hline bottom & \Bot & $\emptyset$\\
\hline
concept name & \con{CN} & $\Int{\con{CN}}\subseteq\Int{\Delta}$\\
\hline
general negation   & $\Not{C}$ & $\Int{\Delta}\setminus\Int{C}$\\
\hline
conjunction &  $C\And D$ & $\Int{C}\cap\Int{D}$\\
\hline
disjunction   & $C\Or D$ & $\Int{C}\cup\Int{D}$\\
\hline
exists restriction & \Some{R}{C}& $\set{x\in\deltai \mid \exists y. \tup{x,y}\in\Int{R}\land y\in\Int{C}}$\\
\hline
value restriction & \All{R}{C} & $\set{x\in\deltai \mid \forall y.\tup{x,y} \in\Int{R}\rightarrow y\in\Int{C}}$\\
\hline
at-most restriction & $\leq nR$ & $\set{x\in\deltai \mid \sharp \set{y\in\deltai \mid R^\I(x,y)}\leq n}$\\
\hline
at-least restriction & $\geq nR$ & $\set{x\in\deltai \mid \sharp \set{y\in\deltai \mid R^\I(x,y)}\geq n}$\\
\hline
\end{tabular}
\caption{Semantics of \shin-concepts} \label{table:si-sem}
\end{center}
\end{table}

%George 13/07/05 added (KB)
A \shin knowledge base (KB) consists of a TBox, an RBox and an
ABox. A \shin \emph{TBox} is a finite set of \emph{concept
inclusion axioms} of the form $C\Issub D$, or \emph{concept
equivalence axioms} of the form $C\equiv D$, where $C,D$ are
\shin-concepts. An interpretation \I satisfies $C\Issub D$ if
$C\ifunc\Sub D\ifunc$ and it satisfies $C\equiv D$ if $C^\I=D^\I$.
Note that concept inclusion axioms of the above form are called
general concept inclusions~\cite{Horrocks99,Baader90}. A \shin
\emph{RBox} is a finite set of \emph{transitive role} axioms
($\Tr(R)$), and \emph{role inclusion} axioms ($R\sqsubseteq S$).
An interpretation \I satisfies $\Tr(R)$ if, for all
$x,y,z\in\deltai$, $\set{\tup{x,y},\tup{y,z}}\subseteq
R\ifunc\rightarrow\tup{x,z}\in R\ifunc$, and it satisfies $R
\sqsubseteq S$ if $R^\I \subseteq S^\I$. A set of role inclusion
axioms defines a \emph{role hierarchy}. For a role hierarchy we
introduce \sss as the transitive-reflexive closure of
$\sqsubseteq$. At last, observe that if $R\sqsubseteq S$, then the
semantics of role inclusion axioms imply that $\Inv(R)^\I\subseteq
\Inv(S)^\I$, while the semantics of inverse roles imply that
$\Tr(\Inv(R))$. A \si RBox is obtained by a \shin RBox if we
disallow role inclusion axioms. A \shin \emph{ABox} is a finite
set of \emph{individual axioms} (or \emph{assertions}) of the form
$\indv{a}:C$, called \emph{concept assertions}, or
$\tup{\indv{a},\indv{b}}:R$, called \emph{role assertions}, or of
the form $\indv{a}\ndoteq \indv{b}$. An interpretation \I
satisfies $\indv{a}:C$ if $\indv{a}\ifunc\in C\ifunc$, it
satisfies $\tup{\indv{a},\indv{b}}:R$ if
$\tup{\indv{a}\ifunc,\indv{b}\ifunc}\in R\ifunc$, and it satisfies
$\indv{a}\ndoteq \indv{b}$ if $\indv{a}\ifunc\neq\indv{b}\ifunc$.
A \si ABox is obtained by a \shin ABox by disallowing inequality
axioms $\indv{a}\ndoteq \indv{b}$. An interpretation \I satisfies
a \shin knowledge base \kb if it satisfies all the axioms in \kb.
\kb is \emph{satisfiable} (\emph{unsatisfiable}) iff there exists
(does not exist) such an interpretation \I that satisfies \kb. Let
$C,D$ be \shin-concepts, $C$ is \emph{satisfiable}
(\emph{unsatisfiable}) \wrt \kb iff there
%exist
exists (does not exist) an interpretation \I of \kb s.t.
$C\ifunc\not=\emptyset$; $C$ subsumes $D$ \wrt \kb iff for every
interpretation \I of \kb we have $C\ifunc\Sub D\ifunc$. Given a
concept axiom, a role axiom, or an assertion $\Psi$, \kb
\emph{entails} $\Psi$, written as $\kb\models\Psi$, iff for all
models \I of \kb we have $\I$ satisfies $\Psi$.

%\si concept
%  satisfiability and subsumption problems, as well as the \si
%  knowledge base satisfiability problem can be solved by tableaux
%  algorithms~\cite{HST98,HST00}.

%The semantics of Description Logic languages are based on
%model-theoretic semantics \cite{Tars56} which use the notion of an
%interpretation to assign meaning to concepts and the connectives
%that are used. An interpretation $\I$ consists of the pair
%$\I=(\Delta^\I,\cdot^\I)$, where $\Delta^\I$ is a non-empty set of
%objects (the domain of interpretation) and $\cdot^\I$ an
%interpretation function which assigns to every atomic concept $A$
%a set $A^\I\subseteq \Delta^\I$ and to every atomic role $R$ a
%binary relation $R^\I\subseteq \Delta^\I\times \Delta^\I$. Using
%the notion of a characteristic function ($\chi_{A^\I}$)\cite{},
%which defines a subset $A^\I$ of a universal set $\Delta^\I$
%($\Delta^\I\times \Delta^\I$), we can alternatively view an
%interpretation function as a function which assigns to atomic
%concepts and roles the characteristic function that defines the
%sets $A^\I$ and $R^\I$, respectively.

\subsection{Fuzzy Sets}
Fuzzy set theory and fuzzy logic are widely used today for
capturing the inherent vagueness (the lack of distinct boundaries
of sets) that exists in real life applications \cite{Klir95}. The
notion of a fuzzy set was first introduced by \citeA{Zadeh65}.
While in classical set theory an element either belongs to a set
or not, in fuzzy set theory elements belong only to a certain
degree. More formally, let \textit{X} be a collection of elements
(called \emph{universe of discourse}) i.e \textit{\(X=\{
x_{1},x_{2}, \ldots \} \)}. A crisp subset \textit{A} of
\textit{X} is any collection of elements of \textit{X} that can be
defined with the aid of its \emph{characteristic function}
\textit{\(\chi_{A}\)(x)} that assigns any \textit{\(x \in X\)} to
a value 1 or 0 if this element belongs to \textit{X} or not,
respectively. On the other hand, a fuzzy subset \textit{A} of
\textit{X}, is defined by a \emph{membership function}
\textit{\(\mu_{A}\)(x)}, or simply \textit{A(x)},
%Jeff 18/06/05
for each \textit{\(x \in X\)}. This membership function assigns
any \textit{\(x \in X\)} to a value between 0 and 1 that
represents the degree in which this element belongs to
%Jeff 18/06/05
%\textit{X}.
$A$. Additionally, a \emph{fuzzy binary relation} $R$ over two
crisp sets $X$ and $Y$ is a function $R:X\times
Y\rightarrow[0,1]$. For example, one can say that $Tom$ belongs to
the set of $Tall$ people to a degree of $0.8$, writing
$Tall(Tom)=0.8$, or that the object $o_1$ is part of the object
$o_2$ to a degree of 0.6, writing $isPartOf(o_1,o_2)=0.6$. Several
properties of fuzzy binary relations have been investigated in the
literature \cite{Klir95}. For example, a binary fuzzy relation is
called \emph{$\sup$-$\min$ transitive} if $R(x,z)\geq \sup_{y\in
Y}\{\min(R(x,y),R(y,z))\}$, while the inverse of a relation $R$ is
defined as $R^{-1}(y,x)=R(x,y)$ \cite{Klir95}.

Using the above idea, the most important operations defined on
crisp sets and relations, like
%Jeff 18/06/05
the boolean operations (complement, union,
%Jeff 18/06/05
and intersection etc.), are extended in order to cover fuzzy sets
and fuzzy relations.
%Jeff 18/06/05
%Thus,
%Apart from it ability to deal with vagueness, fuzzy set theory can
%also be used to formalize possibility theory
%\cite{Klir95,Dubois94}. Possibility theory is another important
%theory, intended in capturing uncertainty that comes from the
%incompleteness (lack) of information \cite{Dubois94b}. In fact,
%possibility theory can be seen as a measure-theoretic counterpart
%of fuzzy set theory \cite{Klir95}.
Accordingly, a sound and complete mathematical framework that
plays an important role in the management of %uncertain
imprecise and vague information has been defined and used in a
wide set of scientific areas including expert systems and decision
making \cite{Zimmermann87}, pattern recognition \cite{Kandel82},
image analysis and computer
vision \cite{Krishnapuram92}, medicine \cite{Oguntade82}, %economics
%\cite{Billot92},
control \cite{Sugeno85}, etc.

%[Todo: we need to add some references for the above application
%domains.]

\subsection{Fuzzy Set Theoretic Operations}\label{sec:norms}
%Jeff 18/06/05
In this section, we will explain how to extend boolean operations
and logical implications in the context of fuzzy sets and fuzzy
logics. These operations are now performed by mathematical
functions over the unit interval.

%George 22/09/05
The operation of complement is performed by a unary operation,
$c:[0,1]\rightarrow [0,1]$, called \emph{fuzzy complement}. In
order to provide meaningful fuzzy complements, such functions
should satisfy certain properties. More precisely, they should
satisfy the \emph{boundary conditions}, $c(0)=1$ and $c(1)=0$, and
be \emph{monotonic decreasing}, for $a\leq b$, $c(a)\geq c(b)$. In
the current paper we will use the Lukasiewicz negation,
$c(a)=1-a$, which additionally is \emph{continuous} and
$\emph{involutive}$, for each $a\in[0,1]$, $c(c(a))=a$ holds. In
the cases of \emph{fuzzy intersection} and \emph{fuzzy union} the
mathematical functions used are binary over the unit interval.
These functions are usually called \emph{norm} operations referred
to as t-norms ($t$), in the case of fuzzy intersection, and
t-conorms (or s-norms) ($u$), in the case of fuzzy union \cite{Klement04a}.
Again these operations should satisfy certain mathematical
properties. More precisely, a t-norm (t-conorm) satisfies the
\emph{boundary condition}, $t(a,1)=a$ $(u(a,0)=a)$, is
\emph{monotonic increasing}, for $b\leq d$ then $t(a,b)\leq
t(a,d)$ $(u(a,b)\leq u(a,d))$, \emph{commutative}, $t(a,b)=t(b,a)$
$(u(a,b)=u(b,a))$, and \emph{associative},
$t(a,t(b,c))=t(t(a,b),c)$ $(u(a,u(b,c))=u(u(a,b),c))$. Though
there is a wealth of such operations in the literature
\cite{Klir95} we restrict our attention to specific ones. More
precisely, we are using the G\"odel t-norm, $t(a,b)=\min(a,b)$ and
the G\"odel t-conorm, $u(a,b)=\max(a,b)$. Additionally to the
aforementioned properties, these operations are also
\emph{idempotent}, i.e. $\min(a,a)=a$ and $\max(a,a)=a$, hold.
Finally, a \emph{fuzzy implication} is performed by a binary
operation, of the form $\J:[0,1]\times[0,1]\rightarrow [0,1]$. In
the current paper we use the Kleene-Dienes fuzzy implication which
is provided by the equation, $\J(a,b)=\max(c(a),b)$. The reason
for restricting our attention to these operations would be made
clear in section \ref{sec:tableauxproc}. We now recall a property
of the $\max$ norm operation that we are going to use in the
investigation of the properties of transitive relations under the
framework of fuzzy set theory.

\begin{lemma}\label{thm:props}
\cite{Hajek98} For any $a,b\in [0,1]$, where $j$ takes values from
the index set $J$, the $\max$ operation satisfies the following
property:
\begin{itemize}
%Jeff 05/07/05
%    \item [(1)] $\sup_{j\in J}\min(a,b_j)=\min(a,\sup_{j\in J}b_j)$,\label{eq:1}
    \item $\inf_{j\in J}\max(a,b_j)=\max(a,\inf_{j\in J}b_j)$.\label{eq:2}
\end{itemize}
\end{lemma}

%Jeff 18/06/05
%Clearly the characteristic function of a crisp set assigns a value
%of either 1 or 0 and thereby discriminating between members and
%nonmembers of the crisp set.

%Jeff 18/06/05
\section{The \fkdshin DL}\label{sec:f-SI}

%Jeff 18/06/05
In this section, we introduce
%the \fkd-\si DL, which is
a fuzzy extension of the \shin DL presented in
Section~\ref{sec:dl}. Following \citeA{Stoilos05}, since we are
using the Kleene-Dienes (KD) fuzzy implication in our language, we
call it \fkdshin. This presentation follows the standard syntax
and semantics of fuzzy DLs, that has been introduced in the
literature \cite{Straccia01,Holldobler02,Sanchez04}. More
precisely, \fkdshin was first presented by \citeA{Straccia05}. For
completeness reasons we will also present the language \fkdshin
here. Please also note that our presentation differs from that of
\citeA{Straccia05} in the semantics of concept and role inclusion
axioms.

%Jeff 18/06/05
As usual, we consider an alphabet of distinct concept names
(\Concepts), role names (\Roles) and individual names
(\Individuals). The abstract syntax of \fkdshin-concepts and
\fkdshin-roles (and respectively of \fkdsi-concepts and
\fkdsi-roles) is the same as their \shin counterparts; however,
their semantics is based on fuzzy interpretations (see below).
Similarly, \fkdshin keeps the same syntax of concept and role
axioms as their counterparts in \shin. Interestingly, \fkdshin
extends \shin individual axioms (assertions) into fuzzy individual
axioms, or fuzzy assertions (following, \citeR{Straccia01}), where
membership degrees can be asserted.

%Jeff 18/06/05
Firstly, by using membership functions that range over the
interval $[0,1]$, classical interpretations can be extended to the
concept of \emph{fuzzy interpretations} \cite{Straccia01}.
%Now
Here we abuse the symbols and define a fuzzy interpretation as a
pair $\I=(\deltai,\cdot\ifunc)$,\footnote{In the rest of the
paper, we use $\I=(\deltai,\cdot\ifunc)$ to represent fuzzy
interpretations instead of crisp interpretations.} where the
domain \deltai is a non-empty set
%George 13/07/05 added "of objects"
of objects and $\cdot\ifunc$ is a
%Jeff 18/06/05
\emph{fuzzy interpretation function}, which maps
\begin{enumerate}
    \item  an individual name $\indv a\in\Individuals$   to an element $\indv
    a\ifunc\in\deltai$,

    \item   a concept name  $\con A\in\Concepts$ to a membership function
    $\con A\ifunc:\deltai\rightarrow [0,1]$,

     \item   a role name  $R\in\Roles$ to a membership function
     $R\ifunc:\deltai\times\deltai\rightarrow [0,1]$.

\end{enumerate}
For example, if $o\in\Delta^\I$ then $\con A^\I(o)$ gives the
degree that the object $o$ belongs to the fuzzy concept $\con A$,
e.g. $\con A^\I(o)=0.8$. By using the fuzzy set theoretic
operations  defined in section \ref{sec:norms}, the fuzzy
interpretation function can be extended to give semantics to
\fkdshin-concepts and \fkdshin-roles. For example, since we use
the $\max$ function for fuzzy union the membership degree of an
object $a$ to the fuzzy concept $(C\sqcup D)^\I$ is equal to
$\max(C^\I(a),D^\I(a))$. Moreover since, according to Table
\ref{table:si-sem}, a value restriction $\all R C$ is an
implication of the form, $\forall y(R(x,y)\rightarrow C(y))$, we
can interpret $\forall$ as $\inf$ \cite{Hajek98}, and
$\rightarrow$ as the Kleene-Dienes fuzzy implication and finally
have the equation,
$\inf_{b\in\Delta^\I}\{\max(1-R^\I(a,b),C^\I(b))\}$. The complete
set of semantics is depicted in Table \ref{table:fkdsi-sem}. We
have to note that there are many proposals for semantics of number
restrictions in fuzzy DLs \cite{Sanchez04,Straccia05}. We choose
to follow the semantics proposed by \citeA{Straccia05} since they
are based in the First-Order interpretation of number restrictions
\cite{Baader03a}. Moreover, as it is shown by \citeA{Stoilos05c}
and as we will see in section \ref{sec:f-SHIN}, under these
semantics all inference services of \fkdshin stay decidable and
reasoning can be reduced to a simple counting problem, yielding an
efficient algorithm.
%Jeff 30/11/05
Note that,
%George 01/12/05
although
most of the above semantics have been presented elsewhere
\cite{Sanchez04,Straccia05}, %but for sake of readability
we include them here simply for the sake of completeness.
% The
%interpretation of value and existential restrictions is the usual
%one found in literature \cite{Hajek98}. They result
%%by
%from the classical predicate formulas $\forall y(R(x,y)\rightarrow
%C(y))$ and $\exists yR(x,y)\wedge C(y)$, respectively,
%%George 22/09/05
% by interpreting $\forall$ as $\inf$ and $\exists$ as
%$\sup$ and then using the relevant fuzzy operations for each crisp
%connective.
%note that here we interpret $\forall$ as $\inf$ and $\exists$ as
%$\sup$.

%Jeff 18/06/05
\begin{table}[t]
\begin{center}
\begin{tabular}{|l|c|l|}
\hline
\hspace*{8pt} Constructor & \hspace*{3pt}Syntax   &  \hspace*{75pt}   Semantics     \\
\hline  top & \Top & $\Top^\I(a)=1$   \\
 \hline
 bottom & \Bot & $\Bot^\I(a)=0$
\\
%\hline
% concept name & \con{CN} & $\Int{\con{CN}}\subseteq\Int{\Delta}$  \\
\hline
general negation    & $\Not{C}$  & $(\neg C)^\I(a)=1-C^\I(a)$\\
\hline
conjunction &  $C\And D$ & $(C\sqcap D)^\I(a)=\min(C^\I(a),D^\I(a))$\\
\hline
disjunction   & $C\Or D$ & $(C\sqcup D)^\I(a)=\max(C^\I(a),D^\I(a))$\\
\hline
exists restriction & \Some{R}{C} & $(\exists R.C)^\I(a)=\sup_{b\in\Delta^\I}\{\min(R^\I(a,b),C^\I(b))\}$\\
\hline
value restriction & \All{R}{C} & $(\forall R.C)^\I(a)=\inf_{b\in\Delta^\I}\{\max(1-R^\I(a,b),C^\I(b))\}$\\
\hline
at-most & $\leq pR$ & $\inf_{b_1,\ldots,b_{p+1}\in\Delta^\I}\max^{p+1}_{i=1}\{1-R^\I(a,b_i)\}$\\
\hline
at-least & $\geq pR$ & $\sup_{b_1,\ldots,b_p\in\Delta^\I}\min^p_{i=1}\{R^\I(a,b_i)\}$\\
\hline
inverse role   & $R^-$  & $(R^{-})^\I(b,a)=R^\I(a,b)$ \\
\hline

\end{tabular}
\caption{Semantics of \fkdshin-concepts and \fkdshin-roles}
\label{table:fkdsi-sem}
\end{center}
\end{table}

An \fkdshin knowledge base consists of a TBox, an RBox and an
ABox. Let \con A be a concept name and $C$ an \fkdshin concept. An
\fkdshin TBox is a finite set of fuzzy concept axioms
%. Let \con
%A be a concept name, $C$ a \fkd-\si-concept. Fuzzy concept axioms
of the form $\con A\Issub C$,
% are
called \emph{fuzzy inclusion introductions}, and
% fuzzy concept axioms
of the form $\con A\equiv C$,
% are
called \emph{fuzzy equivalence introductions}. A fuzzy
interpretation \I satisfies $\con A\Issub C$ if $\forall
o\in\deltai, \con A\ifunc(o)\leq C\ifunc(o)$. A fuzzy
interpretation satisfies $\con A\equiv C$ if $\forall o\in\deltai,
\con A\ifunc(o)= C\ifunc(o)$. A fuzzy interpretation \I satisfies
an \fkdshin TBox \T iff it satisfies all fuzzy concept axioms in
\T; in this case, we say that  \I is a \emph{model} of \T.

%Jeff 18/06/05
There are two remarks here. Firstly,
%by the above definition of
%satisfiability of fuzzy concept inclusion axioms, we have defined
we give a crisp subsumption of fuzzy concepts here, which  is the
usual way subsumption is defined in the context of fuzzy sets
\cite{Klir95}. In contrast, \citeA{Straccia05} defines a fuzzy
subsumption of fuzzy concepts. As it was noted by
\citeA{Bobillo06} in f$_{KD}$-DLs fuzzy subsumption is sometimes
counterintuitive.
%\citeA{Straccia05} has introduced a
% a formula for defining
%fuzzy subsumption of fuzzy concepts,
% This definition
%which only applies in fuzzy DLs that use
%$R$-implications~\cite{Hajek98}, instead of $S$-implications that
%we use in this paper.
% to
%perform fuzzy implications,
%hence not in our context.
%Stoilos 13/07/06
Secondly, as we can see, we are only allowing for \emph{simple}
TBoxes. A TBox \T is called \emph{simple} if it neither includes
cyclic nor general concept inclusions, i.e. axioms are of the form
$A\sqsubseteq C$ or $A\equiv C$, where $A$ is a concept name that
is never defined by itself either directly or indirectly.
%have not defined
%\fkdsi does not support fuzzy
%general concept inclusions.
%~\cite{Horrocks99,Baader90};
%George 13/07/05 added word "directly" because the reduction to crisp SHI solves GCIs
%how to directly deal with such axioms still remains an open
%problem in fuzzy Description Logics.
A procedure to deal with cyclic and general TBoxes, in the context
of fuzzy DLs, has been recently developed by \citeA{Stoilos06ate},
while also in parallel a slightly different technique was
presented by \citeA{Li06a}. This process involves additional
expansion rules and a preprocessing step called
\emph{normalization}, which are not affected by the expressivity
of the underlying fuzzy DL. Hence, in order to keep our
presentation simple we will not consider general TBoxes in the
following, but we will focus on the decidability and reasoning of
\fkdsi and \fkdshin, which involve many technical details. At the
end of section \ref{sec:f-SHIN} we will comment more on the issue
of handling GCIs in the \fkdshin language.

%Jeff 18/06/05
An \fkdshin RBox is a finite set of fuzzy transitive role axioms
of the form $\Tr(R)$ and fuzzy role inclusion axioms of the form
$R\sqsubseteq S$, where $R,S$ are \fkdshin-roles. A fuzzy
interpretation \I satisfies $\Tr(R)$ if $\forall a,c\in\deltai$,
$R^\I(a,c)\geq \sup_{b\in\Delta^\I}\{ \min(
R^\I(a,b),R^\I(b,c))\}$, while it satisfies $R\sqsubseteq S$ if
$\forall a,b\in\deltai$, $R^\I(a,b)\leq S^\I(a,b)$. Note that the
semantics
%The
%semantics for transitivity of fuzzy relations
result
% by
from the definition of $\sup$-$\min$ transitive relations in fuzzy
set theory. A fuzzy interpretation \I satisfies an \fkdshin RBox
\R iff it satisfies all fuzzy transitive role axioms in \R; in
this case, we say that \I is a model of \R. Similarly with the
classical \shin language, the semantics of inverse roles and role
inclusion axioms of \fkdshin imply that from $\Tr(R)$ and
$R\sqsubseteq S$ it holds that $\Tr(\Inv(R))$ and
$\Inv(R)^-\sqsubseteq \Inv(S)^-$.

%Jeff 18/06/05
An \fkdshin ABox is a finite set of fuzzy
assertions~\cite{Straccia01}
%.
%Let $\Individuals=\{a,b,c,...\}$ be a set of individual names.
%A \emph{fuzzy assertion} \cite{Straccia01} is
of the form $(\indv a:C)\bow n$ or $(\tup{\indv a,\indv b}:R)\bow
n$, where $\bow$ stands for $\geq, >, \leq$ and $<$, and
$n\in[0,1]$ or of the form $\indv a\ndoteq \indv b$. Intuitively,
a fuzzy assertion of the form $(\indv a:C)\geq n$ means that the
membership degree of the individual $\indv a$ to the concept $C$
is at least equal to $n$. We call assertions defined by $\geq,>$
\emph{positive} assertions, while those defined by $\leq,<$
\emph{negative} assertions. Formally, given a fuzzy interpretation
\I,
\begin{center}
\begin{tabular}{rcl}
\I satisfies $(\indv a:C)\geq n$ &\hspace{0.1cm} if
\hspace*{0.1cm} & $C^\I(\indv a^\I)\geq n$,\vspace{0.05cm}\\
\I satisfies $(\indv a:C)\leq n$                & if & $C^\I(\indv a^\I)\leq n$,\vspace{0.05cm}\\
\I satisfies $(\tup{\indv a,\indv b}:R)\geq n$  & if & $R^\I(\indv a^\I,\indv b^\I)\geq n$,\vspace{0.05cm}\\
\I satisfies $(\tup{\indv a,\indv b}:R)\leq n$  & if & $R^\I(\indv a^\I,\indv b^\I)\leq n$,\vspace{0.05cm}\\
%\I satisfies $\indv{a}\doteq \indv{b}$          & if & $\indv{a}^\I\doteq \indv{b}^\I$.\vspace{0.1cm}\\
\I satisfies $\indv{a}\ndoteq \indv{b}$         & if & $\indv{a}^\I\neq \indv{b}^\I$.\\
\end{tabular}
\end{center}
The satisfiability of fuzzy assertions with $>,<$ is defined
analogously.
%George 26/05/05
Observe that,
%as first pointed out by \citeA{Holldobler02},
we can also simulate assertions of the form $(\indv a:C)= n$ by
considering two assertions of the form $(\indv a:C)\geq n$ and
$(\indv a:C)\leq n$ \cite{Holldobler02,Straccia01}. A fuzzy
interpretation \I satisfies an \fkdshin ABox \A iff it satisfies
all fuzzy assertions in \A; in this case, we say that  \I is a
model of \A.

%George 13/07/05 removed a strange character (>)
Furthermore, as it was noted by \citeA{Straccia01,Straccia05}, due
to the mathematical properties of the norm operations defined in
section \ref{sec:norms}, the following \fkdshin-concept
equivalences are satisfied: $\neg \top\equiv\bot$, $\neg \bot
\equiv \top$, $C\sqcap \top\equiv C$, $C\sqcup \bot\equiv C$,
$C\sqcup \top\equiv\top$ and $C\sqcap \bot\equiv\bot$.
Furthermore, since the Lukasiewicz complement is involutive it
holds that, $\neg \neg C\equiv C$. Moreover, the De Morgan laws:
%Jeff 17/07/05 removed "(" in front of C_1
$C_1\sqcap C_2\equiv \neg (\neg C_1\sqcup \neg C_2), C_1\sqcup
C_2\equiv \neg (\neg C_1\sqcap \neg C_2)$, are satisfied.
%George 08/09/05 Also removed the other parenthesis ")" :-)
%Jeff 05/07/05: (some R C) etc are DL descriptions, which haven't been introduced yet.
% as well as
%the classical duality between the existential $(\exists)$ and the
%universal $(\forall)$ quantifiers $(\exists R.C\equiv \neg \forall
%R.\neg C)$, hold.
As a consequence of the satisfiability of the De Morgan laws and
the use of the Kleene-Dienes fuzzy implication the following
concept equivalences also hold.
% the \emph{negation normal
%form} (NNF) \cite{Hollunder90} of a concept can be produced. We
%thus assume all concepts to be in their NNF form.
\begin{center}
\begin{tabular}{rclrcl}
$\neg\some{R}{C}$ & $\equiv$ & $\all{R}{(\neg C)}$, & $\neg\all{R}{C}$ & $\equiv$ & $\some{R}{(\neg C)}$,\\
$\neg\leq p_1R$ & $\equiv$ & $\geq (p_1+1)R$, & $\neg\geq p_1R$ & $\equiv$ & $\left\{%
\begin{array}{ll}
    \leq (p_1-1)R, & \hbox{$p_1\in \mathbb{N}^*$} \\
    \bot, & \hbox{$p_1=0$} \\
\end{array}%
\right.$\\
\end{tabular}
\end{center}
At last note that the classical laws of contradiction $(C\sqcap
\neg C\equiv \bot)$ and excluded middle $(C\sqcup \neg C\equiv
\top)$, do not hold.

%Jeff 30/11/05: individuals \indv; in particular \indv o_i
\begin{example}
%Consider for example
Let us revisit the fuzzy knowledge base ($\kb$) that  we
informally introduced in section \ref{sec:intro}. Formally, the
knowledge base can be defined as follows: $\kb=\tup{\T,\R,\A}$,
where
%written this fuzzy
%knowledge base is formed by a fuzzy TBox \T, a fuzzy RBox \R and a
%fuzzy ABox $\A$. Hence, we can write the following definitions:

\begin{center}
\begin{tabular}{lcl}
$\T$ & = & $\{\mathsf{Arm}\sqsubseteq\exists\mathsf{isPartOf.Body}$,\vspace{0.1cm}\\
     &   & $  \mathsf{Body}\sqsubseteq \exists\mathsf{isPartOf.Human}\}$,\vspace{0.1cm}\\
$\A$ & = & $\{(\tup{ \indv o_1, \indv o_2}:\mathsf{isPartOf})\geq 0.8$,$  (\tup{ \indv o_2, \indv o_3}:\mathsf{isPartOf})\geq 0.9$,\vspace{0.1cm}\\
%     &   & $  (\tup{ \indv o_2, \indv o_3}:\mathsf{isPartOf})\geq 0.9$,\vspace{0.1cm}\\
     &   & $  (\indv o_2:\mathsf{Body})\geq 0.85$, $(\indv o_1:\mathsf{Arm})\geq 0.75\}$,\vspace{0.1cm}\\
%     &   & $  (\indv o_1:\mathsf{Arm})\geq 0.75\}$,\vspace{0.1cm}\\

$\R$ & = & $ \{\Tr(\mathsf{isPartOf})\}.$\\
\end{tabular}
\end{center}
Now, in order for some fuzzy interpretation \I to be a model of
$\T$ it should hold that $\mathsf{Arm}^\I(\indv o_i^\I)\leq
(\exists \mathsf{isPartOf}.\mathsf{Body})^\I(\indv o_i^\I)$ and
$\mathsf{Body}^\I(\indv o_i^\I)\leq (\exists \mathsf{isPartOf}.
\mathsf{Body})^\I(\indv o_i^\I)$, $\forall \indv
o_i^\I\in\deltai$.
%Regarding the fuzzy Abox \A we have that
Furthermore, if $\mathsf{isPartOf}^\I(\indv o_1^\I,\indv
o_2^\I)\geq 0.8$, $\mathsf{isPartOf}^\I(\indv o_2^\I,\indv
o_3^\I)\geq 0.9$, $\mathsf{Body}^\I(\indv o_2^\I)\geq 0.85$  and
$\mathsf{Arm}^\I(\indv o_1^\I)\geq 0.75$,   then $\I$ is also a
model of $\A$.
%If
%additionally,
As a model of the RBox \R,
%and
%since $\mathsf{isPartOf}$ is transitive, we can deduce that,
\I should also satisfy that $\mathsf{isPartOf}^\I(\indv
o_1^\I,\indv o_3^\I)\geq\sup_{a \in\deltai}
\{\min(\mathsf{isPartOf}^\I(\indv o_1^\I,
a),\mathsf{isPartOf}^\I(a, \indv o_3^\I))\}=\sup\{\ldots,
\min(0.8$, $0.9),\ldots\}\geq 0.8$.
%as well as $\mathsf{hasPart}^\I(\indv o_3^\I,\indv
%o_1^\I)=\mathsf{isPartOf}^\I(\indv o_1^\I,\indv o_3^\I)$.

Now let us consider  the concept  $\exists \mathsf{hasPart}.
\mathsf{Body}\sqcap \exists \mathsf{hasPart}.\mathsf{Arm}$ that we
mentioned in Section~\ref{sec:intro}. Due to the semantics of
existential restrictions  presented in Table
\ref{table:fkdsi-sem}, we have that

\begin{center}
\begin{tabular}{lcl}
$(\exists \mathsf{hasPart}. \mathsf{Body})^\I(\indv o_3^\I)$ & = &
$\sup_{a\in\deltai}\{\min(\mathsf{(isPartOf^-)}^\I(\indv
o_3^\I,a),\mathsf{Body}^\I(a))\}\geq 0.85,$\vspace{0.1cm}\\

$(\exists \mathsf{hasPart}. \mathsf{Arm})^\I(\indv o_3^\I)$ & = &
$\sup_{a\in\deltai}\{\min(\mathsf{(isPartOf^-)}^\I(\indv
o_3^\I,a),\mathsf{Arm}^\I(a))\}\geq 0.75.$
\end{tabular}
\end{center}

\noindent Hence $\indv o_3$ would belong in the intersection of
the two concepts with the minimum membership degree which is
greater or equal than 0.75, as we claimed in Section
\ref{sec:intro}. \diae
\end{example}

%A fuzzy knowledge base \kb is a triple \tup{\T,\R,\A}, where \T is
%a fuzzy $TBox$, \R is a fuzzy RBox and \A is a fuzzy $ABox$.

%A fuzzy $TBox$ \T together with a fuzzy $ABox$ \A, define a fuzzy
%knowledge base (KB) $\Sigma=(\T, \A)$.

%Jeff 18/06/05
%In
Following \citeA{Straccia01}, we introduce the concept of
conjugated pairs of fuzzy assertions
%has been introduced, in order
to represent pairs of assertions that form a contradiction. The
possible conjugated pairs are defined in Table \ref{conjugated},
where $\phi$ represents a \si assertion. For example, if $\phi=
\indv a:C$, then the assertions $(\indv a:C)\geq 0.7$ and $(\indv
a:C)<0.7$ conjugate.
\begin{table}[ht]
\begin{center}
\begin{tabular}{|c|c|c|}
\hline
& $ \phi<m $ & $\phi\leq m $\\
\hline
$ \phi\geq n$ & $n\geq m$ & $n>m$\\
\hline
$ \phi>n$ & $n\geq m$ & $n\geq m$\\
\hline
\end{tabular}
\end{center}
\caption{Conjugated pairs of fuzzy assertions}\label{conjugated}
\end{table}
Furthermore, due to the presence of inverse roles and role
inclusion axioms the definition should be slightly extended from
that of \citeA{Straccia01} and hence, one should also take under
consideration possible inverse roles or a role hierarchy when
checking for conjugation two role assertions. For example, if
$R\sss S$, then the assertion $(\tup{\indv a,\indv b}:R)\geq 0.9$,
conjugates with $(\tup{\indv b,\indv a}:\Inv(S))\leq 0.4$;
similarly for the rest of the inequalities.

%By using the fuzzy set theoretic operations, defined in section
%\ref{sec:norms}, a fuzzy interpretation can be extended to
%interpret \fkd-\si-concepts. The interpretation of these fuzzy
%concepts are depicted in Table \ref{semantics}. The interpretation
%of value and existential restrictions is the usual one found in
%literature \cite{Hajek98}. They result by the classical predicate
%formulas, $\forall y(R(x,y)\rightarrow C(y))$ and $\exists
%yR(x,y)\wedge C(y)$, respectively, by using the relevant fuzzy
%operations and by interpreting $\forall$ as $\inf$ and $\exists$
%as $\sup$. The semantics for transitivity of fuzzy relations
%results by the definition of transitive relations in fuzzy set
%theory. A fuzzy relation $R$, defined over the domain $X\times X$,
%is called \emph{sup-\min transitive} iff $R(x,z)\geq \sup_{y\in
%X}\min(R(x,y),R(y,z))$.

Now, we will define the reasoning problems of the \fkdshin DL.

A fuzzy interpretation \I satisfies an \fkdshin knowledge base \kb
if it satisfies all  axioms in \kb; in this case, \I is called a
\emph{model} of \kb. An \fkdshin knowledge base \kb is
\emph{satisfiable} (\emph{unsatisfiable}) iff there exists (does
not exist) a fuzzy interpretation \I which satisfies all axioms in
\kb. An \fkdshin-concept $C$ is \emph{satisfiable}
(\emph{unsatisfiable})
%Stoilos 13/07/06
\wrt an RBox \R and a TBox \T iff there
exists (does not exist) some model \I of \R and \T for which there
is some $a\in\Delta^\I$ such that $C^\I(a)=n$, and $n\in(0,1]$. In
this case, $C$ is called \emph{n-satisfiable}
%Stoilos 13/07/06
\wrt \R and \T \cite{Navara00}.
%George 13/04/06
Let $C$ and $D$ be two \fkdshin-concepts. We say that $C$ is
\emph{subsumed} by $D$
%Stoilos 13/07/06
\wrt \R and \T if for every model \I of \R and \T it holds that,
$\forall x\in\deltai.C^\I(x)\leq D^\I(x)$. Furthermore, an
\fkdshin ABox \A is \emph{consistent} \wrt \R and \T if there
exists a model \I of \R and \T that that is also a model of \A.
Moreover, given a fuzzy concept axiom or a fuzzy assertion
$\Psi\in\{C\sqsubseteq D,C\equiv D,\phi \bow n\}$, an \fkdshin
knowledge base \kb \emph{entails} $\Psi$, written $\kb \models
\Psi$, iff all models of \kb also satisfy $\Psi$.

%Jeff 05/07/05
%In absence of fuzzy general concept axioms and cyclic
%terminologies,

%[Todo: Explain the fact the KB satisfiability can be reduced to
%ABox consistency \wrt an RBox.]

%
%George 26/05/05 Adding the glb lub definitions
%Jeff 18/06/05
Furthermore, by studying Table \ref{conjugated}, we can conclude
that an \fkdshin ABox \A can contain a number of positive or
negative assertions without forming a contradiction. Therefore, it
is useful
%It is in our interest
to compute
%which is the $\phi$'s best
 lower
and upper bounds of truth-values.
%In \cite{Straccia01} the concept
Given an \fkdshin knowledge base \kb and an assertion $\phi$,
% of
the  \emph{greatest lower bound} of $\phi$ w.r.t. $\Sigma$
%was
%defined as
is $glb(\Sigma,\phi)=\sup\{n:\Sigma\models  \phi\geq n \}$, where
$\sup \emptyset=0$. Similarly, the
% and
%that of a
\emph{least upper bound}
%as,
of $\phi$ w.r.t. $\Sigma$ is
$lub(\Sigma,\phi)=\inf\{n:\Sigma\models \phi\leq n \}$,
% Observe that
where $\inf \emptyset=1$. A decision procedure to solve the best
truth-value bound was provided by \citeA{Straccia01}. In that
procedure the membership degrees that appear in a \fkdshin ABox,
together with their complemented values and the degrees 0, 0.5 and
1, were collected in a set of membership degrees $N^\Sigma$ and
subsequently the entailment of a fuzzy assertions $\phi \geq n$
and $\phi \leq n$, for all $n\in N^\Sigma$ was tested, thus
determining glb and lub. Obviously this procedure is independent
of the expressivity of the DL language, and thus also applicable
in our context.
%Such a
%procedure can also be used in our framework although we are
%dealing with a far more expressive language than that in
%\cite{Straccia01}.
%[Todo: Explain why/how this decision procedure works for \fkdsi as
%well.]

%Stoilos 13/07/06
\begin{rem}\label{rem:remark}
From Table \ref{table:fkdsi-sem} we see that the semantics of the
value and existential restrictions in fuzzy DLs are defined with
the aid of an infimum and a supremum operation. This means that we
can construct an infinite interpretation \I, i.e. an
interpretation where $\deltai=\{b_1,b_2,\ldots\}$ contains
infinite number of objects, for which $\forall R.C$ is
n-satisfiable ($(\forall R.C)^\I(a)=n$ for some $a\in\deltai$) but
for all $b_i\in\deltai$, $\max(1-R^\I(a,b_i),C^\I(b_i))>n$. This
is possible since although the maximum of the membership degrees
involved for each individual object $b_i$ is strictly greater than
$n$ the limit of the infinite sequence could converge to $n$. This
fact was first noted for fuzzy DLs by \citeA{Hajeck05},
introducing the notion of \emph{witnessed} model for fuzzy DLs. A
model is called \emph{witnessed} if for $(\forall R.C)^\I(a)=n$
there is some $b\in\deltai$ such that either $R^\I(a,b_i)=1-n$ or
$C^\I(b_i)=n$, i.e. there is some $b\in\deltai$ that witnesses the
membership degree of $a$ to $\forall R.C$. Fortunately, there are
fuzzy logics that have an infinite model if and only if they have
a \emph{witnessed} model. More precisely, Hajek proves this
property for the Lukasiewicz fuzzy logic\footnote{Lukasiewicz
fuzzy logic uses the t-norm $t(a,b)=\max(0,a+b-1)$, the t-conorm
$u(a,b)=\min(1,a+b)$, the Lukasiewicz complement and the fuzzy
implication $\J(a,b)=\min(1,1-a+b)$}. He then concludes that the
same proofs can be modified to apply to the fuzzy logic defined by
the fuzzy operators we are using in the current paper. That is
because these operators are definable in the Lukasiewicz logic
\cite{Mostert57}. For the rest of the paper, without loss of
generality, we are going to consider only witnessed models.
\end{rem}

In this paper, we will provide an algorithm to decide the fuzzy
ABox consistency problem \wrt an RBox in very expressive fuzzy
DLs. Many other reasoning problems can be reduced to this problem.
Firstly, concept satisfiability for a fuzzy concept $C$ can be
reduced to consistency checking of the fuzzy ABox \set{$(\indv a:
C)>0$}. Secondly, in this paper, we only consider unfoldable
TBoxes, where KB satisfiability can be reduced to ABox consistency
\wrt an RBox. A TBox is \emph{unfoldable} if it contains no cycles
and contains only \emph{unique introductions}, i.e., concept
axioms with only concept names appearing on the left hand side
and, for each concept name \con{A}, there is at most one axiom in
\T of which \con{A} appears on the left side. A knowledge base
with an unfoldable TBox can be transformed into an equivalent one
with an empty TBox by a transformation called \emph{unfolding}, or
\emph{expansion} \cite{Nebel90}:
%This is accomplished by \emph{eliminating} the TBox with the aid
%of a procedure known as \emph{expansion} of a TBox
%\cite{Baader03}. In an expanded TBox fuzzy
Concept inclusion introductions $\con{A}\sqsubseteq C$ are
replaced by concept equivalence introductions $\con{A}\equiv
\con{A}'\sqcap C$, where $\con{A}'$ is a new concept name, which
stands for the qualities that distinguish the elements of
$\con{A}$ from the other elements of $C$. Subsequently, if $C$ is
a complex concept expression, which is defined in terms of concept
names, defined
%elsewhere
in the TBox, we replace their definitions in $C$.
% such that we
%encounter only concept names.
It can be proved that the initial TBox with the expanded one are
equivalent.

%George 26/05/50
%Jeff 18/06/05

%It should be noted that
Moreover, the problem of entailment can be reduced to the problem
of fuzzy knowledge base satisfiability \cite{Straccia01}. More
precisely, for $\kb=\tup{\T,\R,\A}$,
% then,
$\kb\models   \phi\ \bow\ n $ iff $\kb'=\tup{\T,\R,\A\cup\{ \phi\
\neg\ \bow\ n\}}$ is unsatisfiable. With $\neg\ \bow$, we denote
the ``negation" of
% a type of
inequalities;
%. For example
e.g., if $\bow\equiv\ \geq$ then $\neg\ \bow \equiv\ <$, while if
$\bow\equiv\ <$ then  $\neg\ \bow\equiv\ \geq$.
%At last
Finally, the subsumption problem of two fuzzy concepts $C$ and $D$
\wrt a TBox can also be reduced to the fuzzy knowledge base
satisfiability problem.
%for $ABoxes$..
More formally, \citeA{Straccia01},
% was
proved that $\tup{\T,\emptyset,\A}\models C\sqsubseteq D$ iff
$\tup{\T,\emptyset,\{(\indv a:C)\geq n,(\indv a:D)<n \}}$,
%with
for both $n\in\{n_1,n_2\}$, $n_1\in(0,0.5]$ and $n_2\in(0.5,1]$,
is unsatisfiable.
%George 22/06/05
%Stoilos 13/07/06
The above reduction can be extended in order for a fuzzy knowledge
base to also include an RBox. Please note that, in crisp DLs, in
order to check if a concept $C$ is subsumed by a concept $D$ we
check for the unsatisfiability of the concept, $C\sqcap \neg D$.
This reduction to unsatisfiability is not applicable to \fkd-DLs
since the fuzzy operations that we use do not satisfy the laws of
contradiction and
%Jeff 17/07/05
% the laws of
excluded middle.

We conclude the section with an example.

%Jeff 30/11/05: individuals \indv; in particular \indv o_i
\begin{example}\label{example:reduction}
Consider again our sample knowledge base (\kb). By applying the
transformation of unfolding, defined earlier, one would obtain the
following expanded fuzzy TBox:

\begin{center}
\begin{tabular}{lcl}
$\T'$ & = & $\{\mathsf{Arm}\equiv\mathsf{Arm'}\sqcap\exists\mathsf{isPartOf.Body}$,\vspace{0.1cm}\\
      &   & $  \mathsf{Body}\equiv \mathsf{Body'}\sqcap\exists\mathsf{isPartOf.Human}\}$\vspace{0.1cm}\\
\end{tabular}
\end{center}
while the respective fuzzy assertions would be transformed to

\begin{center}
\begin{tabular}{lcl}
$\A'$ & =  & $\{  (\indv o_2:\mathsf{Body'}\sqcap\exists\mathsf{isPartOf.Human})\geq 0.85\ran$,
\vspace{0.1cm}\\
      &   & $  (\indv o_1:\mathsf{Arm'}\sqcap\exists\mathsf{isPartOf.Body})\geq
      0.75\ran\}$\\
\end{tabular}
\end{center}
Now, let us formally specify the query we introduced in section
\ref{sec:intro}. If $\kb'=\tup{\T',\R,\A'}$ is our modified
knowledge base, after unfolding, the query would have the form
$\kb'\models (\indv o_3:\exists \mathsf{hasPart}.
\mathsf{Body}\sqcap \exists \mathsf{hasPart}.\mathsf{Arm})\geq
0.75$. According to our previous discussion in order to check for
the entailment of such a query one should check for the
%George 13/04/06 unsatisfiability
consistency of the fuzzy ABox $\A'\cup\{(\indv o_3:\exists
\mathsf{hasPart}. \mathsf{Body}\sqcap \exists
\mathsf{hasPart}.\mathsf{Arm})<0.75\}$, \wrt the RBox \R, since
after the expansion we can remove $\T'$. Our task in the following
sections is to provide a procedure that decides the
%George 13/04/06 satisfiability
consistency of a fuzzy ABox \wrt an RBox.

\diae
\end{example}

\section{Transitivity in Fuzzy Description Logics}\label{sec:transinvest}
%Jeff 05/07/05
In classical DLs,  a role $R$ is transitive iff for all
$a,b,c\in\Delta^\I$, $\tup{a,b}\in R^\I$ and $\tup{b,c}\in R^\I$
imply $\tup{a,c}\in R^\I$. \citeA{Sattler96} shows that, for
$a,b,\nlist c\in\Delta^\I$, if $R$ is transitive, $b$ is \emph{an}
$R$-successor of $a$, \nlist c are \emph{the} $R$-successors of
$b$,
 and $a\in(\all R C)^\I$,
then all $R$-successors of $a$ should be instances of $(\all R
C)^\I$, e.g.,  $b \in (\all R C)\ifunc$ because: (i)
$\tup{a,c_i}\in R^\I$ (as $R$ is transitive), (ii) $c_i\in
C\ifunc$ (as $a\in(\all R C)^\I$) and (iii) $b\in (\all R C)^\I$
(due to the semantics of \all R C). In other words, this means
that the following concept subsumption holds, $\forall
R.C\sqsubseteq \forall R.(\forall R.C)$.

The above property suggests that
%This means that if there exist such objects $a,b,c\in\deltai$ and
%$a\in(\forall R.C)^\I$,
% with $\Tr(R)$, then
%due to the transitivity of $R$ and the semantics of the $\forall$
%constructor,
%we have $c\in C^\I$ and
%. But, since $c$ must belong to
%$C^\I$ whenever the relation $R^\I(b,c)$, holds it means that $b$
%must also belong to
%$b\in(\forall R.C)^\I$. This property was identified in
%\cite{Sattler96} where
value restrictions on transitive relations (\all R C) are
% being
propagated along the path of individuals. This propagation is
crucial for reasoning algorithms in order to retain the
\emph{tree-model property} \cite{Baader03a}, which is a property
that leads to
%Stoilos 13/07/06 robustly
decidable decision procedures \cite{Vardi97}. Our goal in the rest
of the section is to investigate this property in the context of
fuzzy Description Logics that allow for transitive role axioms. We
have to determine if similar propagation occurs and if it is, to
find out the membership degree that the propagation carries to
subsequent objects. This is the first time that such an
investigation is presented in the literature.

%Jeff 05/07/05
In fuzzy
%concept languages,
DLs, objects are instances of all possible fuzzy concepts in some
%now that
%since membership
degree, ranging over the interval $[0,1]$. As we have shown in
Section 3, a fuzzy role
 $R$ is transitive iff, for all $a,c\in\deltai, R\ifunc(a,c)\geq \sup_{b\in
\deltai}\min(R\ifunc(a,b),R\ifunc(b,c))$.
%George 08/09/05
Since this holds for the supremum, for an arbitrary $b\in\deltai$
we have that, $R\ifunc(a,c)\geq \min(R\ifunc(a,b),R\ifunc(b,c))$
and by applying fuzzy complement in both sides we get,
$c(R\ifunc(a,c))\leq c(\min(R\ifunc(a,b),R\ifunc(b,c)))$. In what
follows, we will show that not only value restrictions, but also
existential restrictions on transitive roles (\some R C) are being
propagated, so as to  satisfy infimum and supremum restrictions.
%Let $a,b \in\Delta^\I$ be
%%be an
%objects in \deltai and  $R$ a transitive role.
%% $b\in\Delta^\I$ an
%%$R$-successor of $a$ and \nlist c $\in\Delta^\I$ the
% %   $R$-successors of $b$.
% If $(\some R C)\ifunc(a)\leq e_a$,
% %Firstly, due to the semantic of fuzzy transitive roles, we
% %have $R\ifunc(a, c_i)\geq \min(R\ifunc(a,b),R\ifunc(b,c_i))$, for
% %all $1\leq i\leq n$. Secondly, due to the semantics of $(\some R
% %C)\ifunc(a)$, we have $e_a\geq \min(R\ifunc(a,b),C\ifunc(b))$ and $e_a\geq
% %\min(R\ifunc(a,c_i),C\ifunc(c_i))$, for
%% all $1\leq i\leq n$. Similarly, due to the semantics of $(\some R
% %C)\ifunc(b)$, we have $(\some R
% %C)\ifunc(b)\geq
%% \min(R\ifunc(b,c_i),C\ifunc(c_i))$, for
% %all $1\leq i\leq n$.
% we have\\
%%\begin{center}
%\begin{tabular}{ll}
%(1) $\sup_{d\in \deltai}\min(R^\I(a,d),C\ifunc(d))\leq e_a$ & $\Rightarrow_{monotonicity}$\\
%(2) $\sup_{d\in \deltai}\min(\min(R\ifunc(a,b),R\ifunc(b,d)),C\ifunc(d))\leq e_a$ & $\Rightarrow_{associativity}$\\
%(3) $\sup_{d\in \deltai}\min(R\ifunc(a,b),\min(R\ifunc(b,d),C\ifunc(d)))\leq e_a$ & $\Rightarrow_{Lemma\ref{thm:props}:(1)}$\\
%(4) $\min(R\ifunc(a,b),\sup_{d\in \deltai}\min(R\ifunc(b,d),C\ifunc(d)))\leq e_a$ & $\Rightarrow$\\
%(5) $\min(R\ifunc(a,b),(\some R C)\ifunc(b))\leq e_a$,
%\end{tabular}\\
%%\end{center}
%\noindent which means that either $R\ifunc(a,b)\leq e_a$ or
%$(\some R C)\ifunc(b)\leq e_a$. In other words, if
%$R^\I(a,b)>e_a$,   we have
%%the existential restriction
% $(\some R
%C)\ifunc(b)\leq e_a$.
%Jeff 05/07/05
Now we look at the value restrictions. Let $a,b \in\Delta^\I$ be
  objects in \deltai and  $R$ a transitive role.
  % $b\in\Delta^\I$   an
%$R$-successor of $a$ and \nlist c $\in\Delta^\I$ the
 %   $R$-successors of $b$.
 If $(\all R C)\ifunc(a)\geq v_a$, we have (note that $c$ below represents fuzzy complement)\\
\begin{tabular}{ll}
(1) $\inf_{d\in \deltai} \max(c(R^\I(a,d)),C\ifunc(d))\geq v_a$ &
$\Rightarrow_{
%R\ifunc(a,d)\geq \min(R\ifunc(a,b),R\ifunc(b,d))
%George 08/09/05
%R \mbox{\small \ is transitive}}$\\
monotonicity}$\vspace{0.1cm}\\
(2) $\inf_{d\in \deltai}\max(c(\min(R\ifunc(a,b),R\ifunc(b,d))),C\ifunc(d))\geq v_a$ & $\Rightarrow_{De~Morgan}$\vspace{0.1cm}\\
(3) $\inf_{d\in \deltai}\max(\max(c(R\ifunc(a,b)),c(R\ifunc(b,d))),C\ifunc(d))\geq v_a$ & $\Rightarrow_{associativity}$\vspace{0.1cm}\\
(4) $\inf_{d\in \deltai}\max(c(R\ifunc(a,b)),\max(c(R\ifunc(b,d)),C\ifunc(d)))\geq v_a$ & $\Rightarrow_{Lemma~\ref{thm:props}}$\vspace{0.1cm}\\
(5) $\max(c(R\ifunc(a,b)),\inf_{d\in \deltai}\max(c(R\ifunc(b,d)),C\ifunc(d)))\geq v_a$ & $\Rightarrow$\vspace{0.1cm}\\
(6) $\max(c(R\ifunc(a,b)),(\all R C)\ifunc(b))\geq v_a$,
\end{tabular}\\[3pt]
which means   either $c(R\ifunc(a,b))\geq v_a$ or $(\all R
C)\ifunc(b)\geq v_a$. There are some remarks here. Firstly, the
above $b$ is an arbitrary object in \deltai. In other words, for
any object $x\in\deltai$, if $c(R\ifunc(a,x))< v_a$, we have
$(\all R C)\ifunc(x)\geq v_a$. Similarly, if  $(\all R
C)\ifunc(a)> v_a\geq c(R\ifunc(a,x))$, we have $(\all R
C)\ifunc(x)> v_a$. Hence, the following result is obtained.

\begin{corollary}
If $(\forall R.C)^\I(a)\rhd n$ and $\Tr(R)$ then, in an \fkd-DL,
$(\forall R.(\forall R.C))^\I(a)\rhd n$ holds.
\end{corollary}

Now, let $a,b \in\Delta^\I$,
%Stoilos 13/07/06 be objects in \deltai,
$R$ a transitive role and consider the case $(\some R
C)\ifunc(a)\leq e_a$. By applying a fuzzy complement in both sides
of the inequation, and since fuzzy complements are monotonic
decreasing, we obtain $c((\some R C)\ifunc(a))\geq c(e_a)$. Based
on the semantics of the language this can be rewritten as
$(\neg(\some R C))\ifunc(a)\geq c(e_a)$ and by using the concept
equivalences presented in the previous section we have, $(\all R
(\neg C))\ifunc(a)\geq c(e_a)$. Hence, by using the above results
for value restrictions we can conclude that for any object
$x\in\deltai$, if $c(R\ifunc(a,x))< c(e_a)\Rightarrow
R\ifunc(a,x)> e_a$, then $(\all R (\neg C))\ifunc(x)\geq
c(e_a)\Rightarrow (\some R C)\ifunc(x)\leq e_a$ and similarly, if
$(\all R (\neg C))\ifunc(a)> c(e_a)\geq c(R\ifunc(a,x))$, i.e.
$(\some R C)\ifunc(a)<c(e_a)\leq R\ifunc(a,x)$, we have $(\all R
(\neg C))\ifunc(x)> c(e_a)$ and thus $(\some R C)\ifunc(x)< e_a$.
Hence, again we are able to show the next result.

\begin{corollary}
If $(\exists R.C)^\I(a)\lhd n$ and $\Tr(R)$ then, in an \fkd-DL,
$(\exists R.(\exists R.C))^\I(a)\lhd n$ holds.
\end{corollary}

\noindent The above results will be used in properties 11 and 12
in Definition~\ref{SItableau}.

\section{Reasoning with Transitive and Inverse Roles in \fkd-DLs}\label{sec:sireasoning}
In the current section we will show how to reason about transitive
and inverse roles in the context of fuzzy DLs, thus providing a
reasoning algorithm for the \fkdsi language. This algorithm can be
used in order to provide efficient implementations for
applications that do not need the expressive power of \fkdshin.

%George 29/06/05
%Jeff 05/07/05
In section \ref{sec:f-SI},  we  have shown that most inference
services of fuzzy DLs, like entailment and subsumption, can be
reduced to the problem of
%Jeff 05/07/05
ABox consistency checking
%for
%$ABoxes$
\wrt an RBox.
%Additionally, the satisfiability of a
%single concept $D$ can also be reduced to the consistency problem
%for $ABoxes$. More precisely, in classical DLs, a concept $D$ is
%satisfiable iff the $ABox$ $\{a:D\}$ is consistent. In our case we
%have fuzzy assertions which are of the form $\lan a:D\bow n\ran$,
%where $\bow$ stands for $\geq, >, \leq$ or $<$.
%Jeff 05/07/05
As other tableaux algorithms, our tableaux algorithm for checking
ABox consistency
%is usually checked with tableaux algorithms that try
tries to prove the satisfiability of an assertion by constructing,
for an \fkdsi ABox \A, a fuzzy tableau of \A, i.e., an abstraction
of a model of \A. Given the notion of a fuzzy tableau, it is quite
straightforward to prove the algorithm is a decision procedure for
ABox consistency. The fuzzy tableau we present here can be seen as
an extension of the tableau presented by \citeA{Horrocks00} to
handle with degrees. The first such extension was presented by
\citeA{Stoilos05}, but here we will revise that definition.
%This is accomplished by providing a set of
%decomposition rules which unfold the possibly complex concept
%expression represented by $D$. In the case of $ABox$ consistency,
%where various individuals exists and might be interconnected the
%model is represented by a so-called \emph{completion-forest}
%\cite{Horrocks00}, which is a collection of
%\emph{completion-trees}. In such trees each node corresponds to
%individuals in the model, and is labelled with a set of triples of
%the form $\lan D,\bow,n\ran$, which denote the concept, the type
%and the membership degree that the individual of the node has been
%asserted to belong to $D$.

%Jeff 05/07/05
Without loss of generality, we assume all concepts $C$ occurring
in \A to be in \emph{negation normal form} (NNF)
\cite{Hollunder90}; i.e., negations occur in front of concept
names only. A \fkdsi-concept can be transformed into an equivalent
one in NNF by pushing negations inwards using a combination of the
De Morgan laws (which are satisfied by the operations we defined
in section \ref{sec:norms}). Next, for a fuzzy concept $D$,
%Jeff 05/07/05
%For triples of a single node, the concepts of conjugated, positive
%and negative triples can be defined in the obvious way. Since the
%expansion rules decompose the initial concept, the concepts that
%appear in triples are sub-concepts of the initial concept.
we will denote by $sub(D)$ the set that contains $D$ and it is
closed under sub-concepts of $D$ \cite{Horrocks99}. The set of all
sub-concepts
%Jeff 05/07/05
of concepts that appear within an $ABox$ is denoted by $sub(\A)$.
%In the present paper we will extend the notions of a
%\emph{tableau} for an $ABox$ \A \cite{Horrocks00}, to a
%\emph{fuzzy tableau}, and we will prove the connection that holds
%between the consistency of an $ABox$ and the existence of a fuzzy
%tableau.

In the following, we use the symbols $\rhd$ and $\lhd$ as a
placeholder for the inequalities $\geq, >$ and $\leq, <$ and the
symbol $\bowtie$ as a placeholder for all types of inequalities.
Furthermore, we use the symbols $\bowref, \rhd^-$ and $\lhd^-$ to
denote their \emph{reflections};
%Jeff 05/07/05
% For example
e.g., the reflection of $\leq$ is $\geq$ and that of $>$ is $<$.
%The extension of the definition of a tableau for \A is the
%following:
%Finally, we use the notation $+$ to denote the
%\emph{strengthening} or \emph{weakening} of an inequation. For
%example, $+\geq\equiv >$, i.e. strengthens the inequality, while
%$+>\equiv \geq$,

\begin{definition}\label{SItableau}
If \A is an \fkdsi ABox,
%George 13/07/05 added "\R an \fkdsi RBox,"
\R an \fkdsi RBox, $\bR_{\A}$ is the set of roles occurring in \A
and \R together with their inverses and $\Individuals_\A$ is the
set of individuals in \A, a fuzzy tableau T for \A
%George 13/07/05 added "with respect to \R,"
with respect to \R, is defined to be a quadruple (\bS, \cL, \cE,
\cV) such that: \bS is a set of elements, $\cL:\bS\times
sub(\A)\rightarrow [0,1]$ maps each
%Jeff 18/06/05
%individual
element and concept, that is a member of $sub(\A)$, to the
membership degree of that element to the concept,
$\cE:\bR_{\A}\times \bS \times \bS\rightarrow [0,1]$ maps each
role of $\bR_{\A}$ and pair of elements to the membership degree
of the pair to the role, and $\cV:\Individuals_\A\rightarrow \bS$
maps individuals occurring in \A to elements in \bS.
%Jeff 29/07/05
%For the above triples, the concepts of conjugated, positive and
%negative triples can be defined in the obvious way.
For all $s,t\in\bS,$ $C, D\in sub(\A)$, $n\in[0,1]$ and
$R\in\bR_{\A}$, T satisfies:

%George 29/06/05 I did some stuff in the notation. Changed >= to \rhd
%and <= to \lhd and to places where there was a conflict I used \rhd' and \lhd'
\begin{enumerate}
    \item $\cL(s,\bot)=0$ and $\cL(s,\top)=1$ for all
    $s\in\bS$,\vspace{0.0cm}

    \item If $\cL(s,\neg A)\bow n$, then
    $\cL(s,A)\bowref 1-n$,\vspace{0.0cm}

    \item If $\cL(s,C\sqcap D)\rhd n$, then $\cL(s,C)\rhd n$ and
    $\cL(s,D)\rhd n$,\vspace{0.0cm}

    \item If $\cL(s,C\sqcup D)\lhd n$, then $\cL(s,C)\lhd n$ and
    $\cL(s,D)\lhd n$,\vspace{0.0cm}

    \item If $\cL(s,C\sqcup D)\rhd n$, then $\cL(s,C)\rhd n$ or
    $\cL(s,D)\rhd n$,\vspace{0.0cm}

    \item If $\cL(s,C\sqcap D)\lhd n$, then $\cL(s,C)\lhd n$ or
    $\cL(s,D)\lhd n$,\vspace{0.0cm}

    \item If $\cL(s,\forall R.C)\rhd n$, then $\cE(R,\lan
    s,t\ran)\rhd^- 1-n$ or $\cL(t,C)\rhd n$,\vspace{0.0cm}

    \item If $\cL(s,\exists R.C)\lhd n$, then $\cE(R,\lan
    s,t\ran)\lhd n$ or $\cL(t,C)\lhd n$,\vspace{0.0cm}

    \item If $\cL(s,\exists R.C)\rhd n$,
    then there exists $t\in\bS$ such that $\cE(R,\lan
    s,t\ran)\rhd n$ and $\cL(t,C)\rhd n$,\vspace{0.0cm}

    \item If $\cL(s,\forall R.C)\lhd n$, then there
    exists $t\in\bS$ such that $\cE(R,\lan
    s,t\ran)\lhd^-1-n$ and $\cL(t,C)\lhd n$,\vspace{0.0cm}

    \item If $\cL(s,\exists R.C)\lhd n$ and $\Tr(R)$, then $\cE(R,
    \lan s,t\ran)\lhd n$ or $\cL(t, \exists R.C)\lhd
    n$,\vspace{0.0cm}

    \item If $\cL(s,\forall R.C)\rhd n$, $\Tr(R)$, then
    $\cE(R,\lan s,t\ran)\rhd^-1-n$ or
    $\cL(t,\forall R.C)\rhd n$,\vspace{0.0cm}

    \item $\cE(R,\lan s,t\ran)\bow n$ iff $\cE(\Inv(R),
    \lan t,s\ran)\bow n$,\vspace{0.0cm}

   % \item There do not exist two conjugated triples in any set of
%    triples for any individual $x\in\bS$,\vspace{0.0cm}
%Giorgos 18/06/05
%    \item There does not exist an individual $x\in\bS$, which
%    contains two conjugated triples in its label,\vspace{0.1cm}
%Jeff 29/07/05
%Giorgos 19/10/05
    \item If $(a:C)\bow n\in\A$, then $\cL(\cV(a),C)\bow
    n$,\vspace{0.0cm}

%Jeff 29/07/05
%Giorgos 19/10/05
    \item If $(\tup{a,b}:R)\bow n\in\A$, then
    $\cE(R,\lan\cV(a),\cV(b)\ran)\bow n$\vspace{0.0cm}

\end{enumerate}
\end{definition}
%Analogous properties apply if we substitute $\geq$ by $>$ and
%$\leq$ by $<$.

%Jeff 05/07/05
%We can
%0bserve that
% when there is a
%for positive triples, properties of fuzzy tableau are
%much like
%similar to the properties of the classical tableaus
%\cite{Horrocks00}.
%On the other hand when we have
%For negative triples, however, the properties are somewhat
%inverted
%. This is
% because

%George Remark:correctly consider ALL models 28/09/05
There are some remarks regarding Definition \ref{SItableau}.
%Stoilos 13/07/06
First, observe that we use the notation $\cE(R,\lan s,t\ran)$
instead of simply $\cE(R,s,t)$ in order to distinguish between a
role $R$ and an ordered pair of nodes $\lan s,t\ran$. Moreover, in
the above definition we are based on the semantics of fuzzy
interpretations, presented in Table \ref{table:fkdsi-sem}, in
order to find properties of the fuzzy models according to what
relation holds between a membership degree, a specific value and
an inequality type. Then, based on these properties we would
develop tableaux expansion rules, which try to construct such an
abstracted model. For example, for property 3, due to the
semantics of $C\sqcap D$, we have that if $(C\sqcap D)^\I(s)\geq
n$, then $C^\I(s)=n_1$, $D^\I(s)=n_2$, with
$\min(n_1,n_2)=(C\sqcap D)^\I(s)\geq n$. Due to the properties of
the $\min$ norm we can conclude that both $n_1\geq n$ and $n_2\geq
n$, hold. Furthermore, for property 7, due to the semantics of
$\all R C$, if $(\forall R.C)^\I(s)\geq n$ we have,
$\max(1-R\ifunc(s,t),C\ifunc(s))\geq n$, hence either
$1-R\ifunc(s,t)\geq n\Rightarrow R\ifunc(s,t)\leq 1-n$ or
$C\ifunc(t)\geq n$. Similarly, we have constructed properties for
all possible relations between a node, an \fkdsi-concept and a
value of the unit interval. Properties 9 and 10 are based on the
fact that we assume the existence of only witnessed models.
Otherwise, no such assumption could be made. Hence, intuitively a
fuzzy tableau is an abstraction of the witnessed models of a fuzzy
ABox. Finally, property 14 means that if a fuzzy assertion of the
form $(a:C)>n$ exists in a fuzzy ABox, then the membership degree
of the node $\cV(a)$ to the concept $C$ in the fuzzy tableau,
should be strictly greater than $n$. Similarly, with the rest of
inequalities as well as with property 15.
%Property 10 is due to the fact that $(\forall R.C)\ifunc(s)\lhd n$
%is equivalent to $(\some R {(\nnnf C)})\ifunc(s) \lhd^- (1-n)$;
%hence, universal constraints are converted into existential
%constraints.
%Intuitively, a fuzzy
%tableau is an abstraction of \emph{all} the models of a fuzzy
%$ABox$, hence in Properties 2, 3, 8 and 9 we have not provided a
%partial solution to the fuzzy equations created. This is a crucial
%property in order to be able to create a fuzzy tableau from an
%arbitrary model of a fuzzy $ABox$. Please also note that in the
%rest of Properties the partial solution is sufficient enough and
%does not affect the creation of a tableau out of a model. This
%would be apparent in the proof of the following lemma.

%Jeff 05/07/05
% For example if property 10
%didn't require a role filler then for some $\lan \forall
%R.C,\leq,0.6\ran\in\cL(s)$ we would have $\max(1-0,0)=1\nleq 0.6$.
%Moreover, in order to give an intuitive example of rule 6 suppose
%that we have $\lan \forall R.C,\geq,0.6\ran\in\cL(s)$ and $\lan
%\lan s,t\ran,\geq,0.2\ran\in\cE(R)$. The last triple is not
%conjugated with $\lan \lan s,t\ran,\leq,0.4\ran$, thus for any
%value p for the triple $\lan C,\bow,p\ran\in\cL(t)$, we have that
%$\max(1-0.2,p)\geq 0.6$ is satisfied and the rule need not be
%applied. On the other hand for $\lan \exists
%R.C,<0.5\ran\in\cL(s)$ and $\lan \lan
%s,t\ran,\geq,0.5\ran\in\cE(R)$ the last triple is conjugated with
%$\lan \lan s,t\ran,<,0.5\ran$, we thus have that
%$\min(R(s,t),C(t))<0.5$, holds, only when $C(t)<0.5$. The same
%results hold for properties 7,10 and 11.

We now have to prove the lemma connecting ABox consistency and the
existence of a fuzzy tableau for \A.

%George 13/07/05 added "w.r.t. \R"
\begin{lemma} \label{SIsatisf}
An \fkdsi ABox \A is consistent w.r.t. \R, iff there exists a
fuzzy tableau for \A w.r.t. \R.
\end{lemma}
\begin{Proof}
For the \emph{if} direction if $T=(\bS,\cL,\cE,\cV)$ is a fuzzy
tableau for an ABox
%George 13/07/05 added "w.r.t. \R"
\A w.r.t. \R, we can construct a fuzzy interpretation
\I=$(\Delta^\I,\cdot^\I)$ that is a model of \A. %The construction
%of such a model has been described by \citeA{Straccia01}.
%%George 13/07/05 added new stuff
%Some additional points have to be taken under consideration
%%Jeff 17/07/05
%%cause
%because of transitive properties. More precisely we have to
%perform the transitive closure of transitive relations in order to
%build a correct model for them.
%
%For a set of triples of the form $\lan A,\geq,n_i\ran$, $i$ a
%positive integer, that might exist within a set of triples
%$\cL(x)$, the \maximum value of $n_i$'s is chosen as a membership
%degree of $x$ to the fuzzy set $A^\I$.
%%George 13/07/05
%%For triples of the form $\lan A,>,n\ran$ a small factor $\epsilon$
%%is added to the maximum.
%If the maximum value participates in a triple of the form $\lan
%A,>,n\ran$ a small factor $\epsilon$ is added to the maximum.
%Furthermore, when a positive triple does not exist, while a
%negative does, in a set $\cL(x)$, the membership degree is set to
%0. In cases where a value or existential restriction exists as
%well as a non conjugated relation, special care to the choice of
%$\epsilon$ has to be made in order not to choose a high value that
%causes a conjugation in the interpretation. Summing up the
%function that returns the maximum is denoted by $glb$
%\cite{Straccia01}. The existence of such a value is ensured by
%property 14 of the fuzzy tableau. Observe that we only construct
%the interpretation of primitive concepts. The interpretation of
%arbitrary concepts is defined by their primitive ones
%\cite{Straccia01}.

An interpretation can be defined as follows:

\begin{center}
\begin{tabular}{rcl}
$\Delta^\I$ & \hspace{0.1cm} = \hspace*{0.1cm} & \bS\vspace{0.1cm}\\
%George 08/09/05 Moved a\in\Individuals_\A to the other side, since a is an individual of the ABox
$a^\I$ & = & $\cV(a)$, $a\in\Individuals_\A$\vspace{0.1cm}\\
$\top^\I(s)$ & = & $\cL(s,\top)$ for all $s\in\bS$\vspace{0.1cm}\\
$\bot^\I(s)$ & = & $\cL(s,\bot)$ for all $s\in\bS$\vspace{0.1cm}\\
$A^\I(s) $ & = & $\cL(s,A)$ for all $s\in\bS$ and concept names $A$\vspace{0.1cm}\\
$R^\I(s,t)$ & = &  $\left\{\begin{tabular}{ll} $R^+_{\cE}(s,t)$ for all $\lan s,t\ran\in\bS\times\bS$ & if $\Tr(R)$\\
                                               $R_{\cE}(s,t)$ for all $\lan s,t\ran\in\bS\times\bS$ & otherwise\\
                            \end{tabular}\right.$\\
\end{tabular}
\end{center}
%%George 08/09/05 I removed the \epsilon[bla bla] cause I think that it is redundant
%\begin{center}{
%\begin{tabular}{rcl}
%$\Delta^\I$ & \hspace{0.1cm} = \hspace*{0.1cm} & \bS\vspace{0.1cm}\\
%%George 08/09/05 Moved a\in\Individuals_\A to the other side, since a is an individual of the ABox
%$a^\I$ & = & $\cV(a), a\in\Individuals_\A$\vspace{0.1cm}\\
%$A^\I(s) $ & = & $glb[\lan A,\bow,n'\ran]$ for all $\lan A,\bow,n'\ran\in\cL(s)$\vspace{0.1cm}\\
%$R^\I(s,t)$ & = & $\left\{\begin{tabular}{ll} $glb[\lan \lan s,t\ran,\bow,n'\ran]$ for all $\lan \lan s,t\ran,\bow,n'\ran\in \cE(R)^+$ & if $\Tr(R)$\\
%                                              $glb[\lan \lan s,t\ran,\bow,n'\ran]$ for all $\lan \lan s,t\ran,\bow,n'\ran\in\cE(R)$ & otherwise\\\end{tabular}\right.$\\
%\end{tabular}}
%\end{center}
where $R_{\cE}(s,t)$ is a binary fuzzy relation defined as
$R_{\cE}(s,t)=\cE(R,\lan s,t\ran)$ for all $\lan
s,t\ran\in\bS\times \bS$, and $R^+_{\cE}$ represents its
\emph{$\sup$-$\min$} transitive closure \cite{Klir95}.
%, for triples of the form $\lan \lan
%s,t\ran,\rhd,n\ran$, where $n$ equals to 0 when $\lan s,t\ran$
%appears only as a
%%George 13/07/05 an->a
%negative triple in $\cE(R)$. This closure preserves the form of an
%assertion, which is either $\geq$ or $>$, in order to correctly
%assign the membership degree of the pair in $R^\I$.
%%George 13/07/05
%If two pairs of nodes have the same membership degree the
%inequality $\geq$ prevails. For example the transitive closure of
%the triples $\lan \lan s_1, s_2\ran,>,n\ran$ and $\lan \lan s_2,
%s_3\ran,\geq, n\ran$, is $\lan \lan s_1, s_3\ran,\geq,n\ran$. For
%two positive triples of the same pair $\lan s,t\ran$, the one with
%the greatest membership degree participates in the closure, while
%for triples with the same membership degree the form $>$ prevails.
%%George 13/07/05 Removed "is".

To prove that \I is a model of \A, we show by induction on the
structure of concepts that $\cL(s,C)\bow n$ implies $C^\I(s)\bow
n$ for any $s\in\bS$. First, Property 1 ensures that the top and
bottom concepts are interpreted correctly. Together with
properties 14, 15, and the interpretation of individuals and
roles, this implies that \I satisfies each assertion in \A.
Without loss of generality, in the following, we will only show
the cases with $\cL(s,C)\geq n$. The rest of the inequalities can
be shown in a similar way.

\begin{enumerate}
    \item If $A$ is a concept name then by definition
    $n\bow\cL(s,A)=A^\I(s)$.\vspace{0.1cm}

    \item If $\cL(s,\neg A)\geq n$, then due to property 2
    $\cL(s,A)\leq 1-n$. By definition of \I, $A^\I(s)\leq 1-n$,
    hence $(\neg A)^\I(s)\geq c(1-n)=n$.\vspace{0.1cm}
    %Similar proof holds for
    %the rest of the inequalities.

    \item If $\cL(s,C\sqcap D)\geq n$, then $\cL(s,C)\geq n$ and
    $\cL(s,D)\geq n$. By induction, $C^\I(s)\geq n$, $D^\I(s)\geq
    n$, hence $(C\sqcap D)^\I(s)=\min(C^\I(s),D^\I(s))\geq
    n$.\vspace{0.1cm}
    %The
    %cases $\cL(s,C \sqcap D)>n$ and $\cL(s,C\sqcup D)\lhd n$ can be
    %shown similarly.

    \item If $\cL(s,C\sqcup D)\geq n$, then $\cL(s,C)\geq n$ or
    $\cL(s,D)\geq n$. By induction either $C^\I(s)\geq n$ or
    $D^\I(s)\geq n$ and $(C\sqcup
    D)^\I(s)=\max(C^\I(s),D^\I(s))\geq n$.\vspace{0.1cm}
    %The cases
    %$\cL(s,C \sqcup D)>n$ and $\cL(s,C\sqcap D)\lhd n$ can be
    %shown similarly.\vspace{0.1cm}

    \item If $\cL(s,\exists R.C)\geq n$, then there exists
    $t\in\bS$ such that, $\cE(R,\lan s,t\ran)\geq n$ and
    $\cL(t,C)\geq n$. By definition $R^\I(s,t)\geq n$ and by
    induction $C^\I(t)\geq n$. Hence, $(\exists
    R.C)^\I(s)=\sup_{t\in\deltai}\min(R^\I(s,t),C^\I(t))\geq
    n$.\vspace{0.1cm}
    %The cases $\cL(s,\exists R.C)>n$ and
    %$\cL(s,\forall R.C)\lhd n$ are shown in a similar way.

%George 30/05/05 Changed \geq in the R^I(s,t) to \rhd
    \item If $\cL(s,\forall R.C)\geq n$ and
    $R^\I(s,t)=p$, then either
    \begin{enumerate}
%George 30/05/05 Also changed \geq to \rhd
%%%%%%%%%
%George 13/07/05 Several corrections
        \item $\cE(R,\lan s,t \ran)=p$, or
        \item there exist several paths $l\geq 1$ of the form,
        $\cE(R,\lan s,s_{l_1}\ran)=p_{l_1}, \cE(R,\lan
        s_{l_1},s_{l_2}\ran)=p_{l_2},\ldots, \cE(R,\lan
        s_{l_m},t\ran)=p_{l_{m+1}}$.
        The membership degree $p$ of the pair $\lan s,t\ran$
        to the transitive closure of $R$,
        would be equal to the maximum degree (since we cannot have
        infinite number of different paths) of all the
        minimum degrees for each path. If that degree
        is such that it is not lower or equal to $1-n$ (since
        $\geq^-=\leq$) then there exists a path, $k$,
        where for all degrees:
        \[\cE(R,\lan
        s_{k_i},s_{k_{i+1}}\ran)=p_{k_i}, 0\leq i\leq m,
        s_{k_0}\equiv s, s_{k_{m+1}}\equiv t,\] it holds that
        $p_{k_i}>1-n$, because all $p_{k_i}$ would be greater
        or equal than the minimum degree of the path. Hence, due
        to Property 12 for all $s_{k_i}$ we have $\cL(s_{k_i},\forall
        R.C)\geq n$.
    \end{enumerate}
%George 13/07/05 Corrections continue
    In case $p\leq 1-n$ we have that $\max(1-p,C^\I(t))\geq n$.
    In case $p\not\leq 1-n$, then $\cL(t, C)\geq n$, so by
    induction $C^\I(t)\geq n$ and thus also
    $\max(1-R^\I(s,t),C^\I(t))\geq n$. In both cases we have that
    $(\forall R.C)^\I(s)\geq n$.
    %The cases $\cL(s,\forall R.C)>n$
    %and $\cL(s,\exists R.C)\lhd n$ can be shown similarly.
\end{enumerate}

For the converse, if \I=$(\Delta^\I,\cdot^\I)$ is a (witnessed)
model of \A \wrt \R, then a fuzzy tableau $T=(\bS,\cL,\cE,\cV)$
for \A \wrt R can be defined as:

\begin{center}
\begin{tabular}{rcl}
\bS & = & $\Delta^\I$\vspace{0.1cm}\\
$\cE(R,\lan s,t\ran)$   &   =   &   $R^\I(s,t)$ \vspace{0.1cm}\\
$\cL(s,C)$              &   =   &   $C^\I(s)$\vspace{0.1cm}\\
$\cV(a)$ & = & $a^\I$
\end{tabular}
\end{center}

\begin{enumerate}
    \item Property 1 is satisfied since \I is a fuzzy
    interpretation.\vspace{0.1cm}

%    \item Properties 2-10 of Definition \ref{SItableau} are
%    satisfied as a direct consequence of the semantics of
%    \fkdsi-concepts.\vspace{0.1cm}

    \item Let $\cL(s,\neg C)\rhd n$. The definition of $T$ implies
    that $(\neg C)^\I(s)=n'\rhd n\Rightarrow C^\I(s)=1-n'\rhd^-
    1-n$, so $\cL(s,C)\rhd^- 1-n$ and Property 2 is satisfied.
    Similarly with the inequalities
    $\lhd\in\{\leq,<\}$.\vspace{0.1cm}

    \item Let $\cL(s,C\sqcap D)\rhd n$. The definition of $T$
    implies that $(C\sqcap D)^\I(s)=n'\rhd n\Rightarrow
    \min(C^\I(s),D^\I(s))=n'\rhd n$. By definition,
    $\cL(s,C)\rhd n$ and $\cL(s,D)\rhd n$ and $T$ satisfies
    Property 3. Property 4 is proved in a similar way.
    \vspace{0.1cm}

    \item Let $\cL(s,C\sqcup D)\rhd n$. The definition of $T$
    implies that $(C\sqcup D)^\I(s)=n'\rhd n\Rightarrow
    \max(C^\I(s),D^\I(s))=n'\rhd n$. By definition of $T$, either
    $\cL(s,C)\rhd n$ or $\cL(s,D)\rhd n$, and  $T$ satisfies Property
    5. Property 6 is proved in a similar way.\vspace{0.1cm}

    \item Let $\cL(s,\forall R.C)\rhd n$. The definition of $T$
    implies that $(\forall R.C)^\I(s)=n'\rhd n\Rightarrow
    \inf_{y\in\deltai}\max(1-R^\I(s,y),C^\I(y))=n'\rhd n$.
    This means that for any $t\in\deltai$ either
    $1-R^\I(s,t)=n'\rhd n$ or $C^\I(t)=n'\rhd n$, and by
    definition either $\cE(R,\lan s,t\ran)\rhd^- 1-n$ or
    $\cL(t,C)\geq n$. Thus, $T$ satisfies Property 7. Property 8
    is proved in a similar way.\vspace{0.1cm}

    \item Let $\cL(s,\exists R.C)\rhd n$. The definition of $T$
    implies that $(\exists R.C)^\I(s)=n'\rhd n \Rightarrow
    \sup_{y\in\deltai}\min(R^\I(s,y),C^\I(y))=n'\rhd n$.
    This means that there exists some $t\in\deltai$ with
    $R^\I(s,t)=n'\rhd n$ and $C^\I(t)=n'\rhd n$. By definition
    $t\in\bS$ and $T$ satisfies Property 9. Property 10 is proved
    in a similar way.\vspace{0.1cm}

    \item Property 12 of definition \ref{SItableau} is satisfied
    as a result of the semantics of transitive roles and value
    restrictions that have been investigated in section
    \ref{sec:transinvest}. Hence, if $(\forall R.C)^\I(s)\geq n$,
    $\Tr(R)$ then either $R^\I(s,t)\leq 1-n$, or $(\forall
    R.C)^\I(t)\geq n$ holds, otherwise if $(\forall R.C)^\I(s)>n$,
    $\Tr(R)$ then either $R^\I(s,t)<1-n$ or $(\forall
    R.C)^\I(t)>n$ holds. By definition of $T$ if $\cL(s,\forall
    R.C)\rhd n$, $\Tr(R)$ then either $\cE(R,\lan s,t\ran)\rhd^-
    1-n$ or $\cL(t,\forall R.C)\rhd n$.
    For similar reasons Property 11, holds.\vspace{0.1cm}

    \item $T$ satisfies Property 13 in Definition
    \ref{SItableau} as a direct consequence of the semantics of
    inverse relations.\vspace{0.1cm}

    \item $T$ satisfies Properties 14 and 15 in Definition
    \ref{SItableau} because \I is a model of \A.

\end{enumerate}
\qed
\end{Proof}

\subsection{An Algorithm for Constructing an \fkdsi Fuzzy
Tableau}\label{sec:tableauxproc}

%Jeff 17/07/05

Now we present a tableaux   algorithm  that tries to construct,
given an
  \fkdsi ABox \A and an \fkdsi RBox \R, a fuzzy tableau for \A \wrt
  \R. We prove that this algorithm construct a fuzzy tableau for
  \A and \R iff there exists a fuzzy tableau for \A and \R, and
  thus decides consistency of \fkdsi ABoxes \wrt RBoxes.

% As it is obvious in order to
%decide ABox consistency a procedure that constructs a fuzzy
%tableau for an \fkd-\si $ABox$ has to be determined. As mentioned
%above such a procedure will be based on tableaux algorithms. In
%the current section we will provide the technical details for
%constructing a correct tableaux algorithm.

%Jeff 17/07/05
Like the  tableaux algorithm presented  by \citeA{Horrocks00}, our
algorithm
%to  decide  consistency of an  \fkdsi ABox \A \wrt an
%RBox \R
 works on \emph{completion-forests} rather than on
\emph{completion-trees},
% This is because
since an ABox might contain several individuals with arbitrary
roles connecting them. Due to the presence of transitive roles,
the termination of the algorithm is ensured by the use of
\emph{blocking}, where
%the
an expansion is terminated when two individuals on the same path
are asserted to belong to the same concepts. As \fkdsi provides
both inverse roles and transitive role axioms, our algorithm uses
\emph{dynamic blocking} \cite{Horrocks99};  i.e., blocked nodes
(and their sub-branches) can be un-blocked and blocked again
later. As it was noted by \citeA{Horrocks99} this un-blocking and
re-blocking technique is crucial in the presence of inverse roles
since information might be propagated up the completion-forest and
affect other branches. For example consider the nodes $x$, $y$ and
$z$, the edges $\lan x,y\ran$ and $\lan x,z\ran$ and suppose that
$x$ blocks $y$. In the presence of inverse roles it is possible
that $z$ adds information to node $x$, although $z$ is a successor
of $x$. In that case the block on $y$ must be broken. Finally,
even in cases where a node is blocked and un-blocking does not
occur it is necessary to allow some expansion to be performed. For
example node $y$ might contain inverse information that if allowed
to be propagated upwards can render the completion-forest
unsatisfiable. Thus, dynamic blocking uses the notions of
\emph{directly} and \emph{indirectly blocked} nodes.

%Jeff 17/07/05
\begin{definition}[Completion-Forest]\label{SIforest}
A completion-forest $\Forest$ for an \fkdsi ABox \A  is a
collection of trees
%that
whose distinguished roots are arbitrarily  connected by edges.
Each node $x$ is labelled with a set $\cL(x)=\{\lan
C,\bow,n\ran\}$, where $C\in sub(\A)$, $\bow\in\{\geq, >, \leq,
<\}$ and $n\in[0,1]$. Each edge \tup{x,y} is labelled with a set
$\cL(\lan x,y\ran)=\{\lan R,\bow,n\ran\}$, where $R\in\bR_\A$ are
(possibly inverse) roles occurring in \A. Intuitively, each triple
$\lan C,\bow,n\ran$ $(\lan R,\bow,n\ran)$, called \emph{membership
triple}, represents the membership degree and the type of
assertion of each node (pair of nodes) to a concept $C\in sub(\A)$
(role $R\in\bR_\A)$.

%of which correspond to the individuals in the ABox.
% The
%Other nodes of the forest correspond to the individuals that have
%been generated in order to satisfy positive existential and
%negative value restrictions and the edges between two nodes to the
%relations that connect two individuals. Nodes are labelled with a
%set of triples $\cL(x)$ (\emph{node triples}), which contain
%concepts that are subsets of $sub(\A)$, augmented with the
%membership degree and the type of assertion that the node belongs
%to the specific concept. More precisely we define $\cL(x)=\{\lan
%C,\bow,n\ran\}$, where $C\in sub(\A)$, $\bow\in\{\geq, >, \leq,
%<\}$ and $n\in[0,1]$. Furthermore, edges $\lan x,y \ran$ are
%labelled with a set $\cL(\lan x,y\ran)$ (\emph{edge triples})
%defined as
% $\cL(\lan x,y\ran)=\{\lan R,\bow,n\ran\}$, where
%$R\in\bR_{A}$. The algorithm expands the
% tree
%forest either by expanding the set $\cL(x)$, of a node $x$ with
%new triples, or by adding new leaf nodes.

%Jeff 17/07/05
%Jeff 29/07/05
If nodes $x$ and $y$ are connected by an edge $\lan x,y \ran$ with
$\tup{R,\bow,n}\in\cL(\tup{x,y})$, then $y$ is called an \rcn R
\bow n-\emph{successor} of $x$ and $x$ is called an \rcn R \bow
n-\emph{predecessor} of $y$. If $y$ is an \rcn R \bow n-successor
or an \rcn {\Inv(R)} \bow n-\emph{predecessor} of $x$, then $y$ is
called an \rcn R \bow n-neighbour of $x$. Let $y$ be an \rcn R >
n-neighbour  of $x$, the edge \tup{x,y} \emph{conjugates} with
triples \tup{R, \lhd, m} if $n\geq m$. Similarly, we can extend it
to the cases of  \rcn R \geq n-, \rcn R < n- and  \rcn R \leq
n-neighbours.

A node $x$ is an $R$-successor (resp. $R$-predecessor or
$R$-neighbour) of $y$ if it is an \rcn R \bow n-successor (resp.
\rcn R \bow n-predecessor or \rcn R \bow n-neighbour) of $y$ for
some role $R$. A node $x$ is a \emph{positive} (resp.
\emph{negative}) successor (resp. predecessor or neighbour) of $y$
if $\bow\in\set{>,\geq}$ (resp. $\bow\in\set{<,\leq}$).
%In
%this case, we say the edge \tup{x,y} connects $x$ and $y$ to a
%degree $n$.
As usual, \emph{ancestor} is the transitive closure of
\emph{predecessor}.

%A node $x$ is called an $R-neighbour$ of a node $x$ if either $y$
%is a successor of $x$ and $\cL(\lan x,y \ran)=\lan R,\bow,n\ran$
%or $y$ is a predecessor of $x$ and $\cL(\lan y,x\ran)=\lan
%\Inv(R),\bow,n\ran$. We then say that the edge triple
%\emph{connects} $x$ and $y$ to a degree of n. If we replace $\bow$
%with $\rhd$ we get the notion of a \emph{positive} $R$-neighbour
%and if by $\lhd$ we get that of a \emph{negative} $R$-neighbour.

%Jeff 17/07/05
A node $x$ is \emph{blocked} iff it is not a root node and it is
either directly or indirectly blocked. A node $x$ is
\emph{directly blocked} iff none of its ancestors are blocked, and
it has an ancestor $y$ such that $\cL(x)=\cL(y)$. In this case, we
say $y$ directly blocks $x$. A node $x$ is \emph{indirectly
blocked} iff one of its predecessor is blocked.

%, for some ancestor $y$ of $x$, $y$ is blocked or $\cL(x)=\cL(y)$.
%A
%blocked
%node $x$ is \emph{indirectly blocked} if its predecessor
%is blocked, otherwise it is \emph{directly blocked}. If $x$ is
%directly blocked, it has a unique ancestor $y$ that blocks it.

%Jeff 17/07/05
%In description logics the notion of a \emph{clash} is used in
%order to denote that a contradiction has occurred in the
%completion-forest. In our framework
A node $x$ is said to contain a \emph{clash} iff there
%exists
exist two conjugated triples,
%within
%a single node,
 or one of the following triples within
 %a node:
 $\cL(x)$:

\begin{center}
\begin{tabular}{l}
$\lan \bot,\geq,n\ran$, $\lan \top,\leq,n\ran$, for $n>0$, $n<1$
respectively\\
$\lan \bot,>,n\ran$, $\lan \top,<,n\ran$\\
$\lan C,<,0\ran$, $\lan C,>,1\ran$
\end{tabular}
\end{center}
Moreover, for an edge $\lan x,y\ran$, $\cL(\lan x,y\ran)$ is said
to contain a clash iff there exist two conjugated triples in
$\cL(\lan x,y\ran)$, or if $\cL(\lan x,y\ran)\cup
\{\lan\Inv(R),\bow,n\ran\mid \lan R,\bow,n\ran\in\cL(\lan
y,x\ran)\}$, and $x,y$ are root nodes, contains two conjugated
triples.
\end{definition}
The definition of a completion-forest is quite intuitive. Since a
fuzzy ABox contains fuzzy assertions of the form $(a:C)\bow n$ and
$(\tup{a,b}:R)\bow n$, then the nodes and edges of the forest must
contain the information about the concept, the type of inequality
and the membership degree for every individual, which in the
forest is represented by a node.

\begin{table*}[t]
\begin{center}{ \footnotesize\hspace*{-20pt}
\begin{tabular}{crl}
\hline
\hspace*{15pt}Rule\hspace{15pt} & \hspace*{25pt}  & \hspace*{90pt}Description      \\
\hline

%George 21/10/05
%Jeff 17/07/05
$(\neg_\bow)$
%$(\sim)$
& if 1.& $\lan \neg
%\sim
C,\bow,n\ran\in\cL(x)$\\
                        &    2.& and $\lan C,\bowref,1-n\ran\not\in\cL(x)$\\
                        & then & $\cL(x)\rightarrow \cL(x)\cup \{\lan C,\bowref,1-n\ran\}$\vspace{0.2cm}\\

$(\sqcap_{\rhd})$  & if 1.& $\lan C_1\sqcap C_2,\rhd,n\ran\in\cL(x)$, $x$ is not indirectly blocked, and\\
                        &    2.& $\{\lan C_1,\rhd,n\ran,\lan C_2,\rhd,n\ran\}\not\subseteq\cL(x)$\\
                        & then & $\cL(x)\rightarrow\cL(x)\cup\{\lan C_1,\rhd,n\ran,\lan C_2,\rhd,n\ran\}$ \vspace{0.2cm}\\

$(\sqcup_{\lhd})$  & if 1.& $\lan C_1\sqcup C_2,\lhd,n\ran\in\cL(x)$, $x$ is not indirectly blocked, and\\
                        &    2.& $\{\lan C_1,\lhd,n\ran,\lan C_2,\lhd,n\ran\}\not\subseteq\cL(x)$\\
                        & then & $\cL(x)\rightarrow\cL(x)\cup\{\lan C_1,\lhd,n\ran,\lan C_2,\lhd,n\ran\}$\vspace{0.2cm}\\

$(\sqcup_{\rhd})$  & if 1.& $\lan C_1\sqcup C_2,\rhd,n\ran\in\cL(x)$, $x$ is not indirectly blocked, and\\
                        &    2.& $\{\lan C_1,\rhd,n\ran,\lan C_2,\rhd,n\ran\}\cap\cL(x)=\emptyset$\\
                        & then & $\cL(x)\rightarrow\cL(x)\cup\{C\}$ for some
                         $C\in\{\lan C_1,\rhd,n\ran,\lan C_2,\rhd,n\ran\}$\vspace{0.2cm}\\

$(\sqcap_{\lhd})$  & if 1.& $\lan C_1\sqcap C_2,\lhd,n\ran\cL(x)$, $x$ is not indirectly blocked, and\\
                        &    2.& $\{\lan C_1,\lhd,n\ran,\lan C_2,\lhd,n\ran\}\cap\cL(x)=\emptyset$\\
                        & then & $\cL(x)\rightarrow\cL(x)\cup\{C\}$ for some
                        $C\in\{\lan C_1,\lhd,n\ran,\lan C_2,\lhd,n\ran\}$\vspace{0.2cm}\\

$(\exists_{\rhd})$ & if 1.& $\lan\exists R.C,\rhd,n\ran\in\cL(x)$, $x$ is not blocked,\\
%Jeff 29/07/05
                        &    2.& $x$ has no \rcn R
\rhd n-neighbour $y$
%Jeff 17/07/05
                        % connected with a triple
                        %with $\cL(\tup{x,y})=\lan
%Jeff 17/07/05
                        %R^*,
                        %R, \rhd, n\ran$
                        and $\lan C,\rhd,n\ran\in\cL(y)$\\
                        & then & create a new node $y$ with $\cL(\lan x,y\ran)
                        =\{\lan R,\rhd,n\ran\}$, $\cL(y)=\{\lan C,\rhd,n\ran\}$\vspace{0.2cm}\\
%George 29/06/05
$(\forall_{\lhd})$ & if 1.& $\lan \forall
R.C,\lhd,n\ran\in\cL(x)$, $x$ is
not blocked,\\
                        &    2.& $x$ has no
%Jeff 29/07/05
                        \rcn R
{\lhd^-} {1-n}-neighbour $y$
%Jeff 17/07/05
                        %connected
                        %with
                        %a triple
                        %$\cL(\tup{x,y})=\lan
                        %R^*
                       % R,\lhd^-,1-n\ran$
                       and $\lan C,\lhd,n\ran\in\cL(y)$\\
                        & then & create a new node $y$ with $\cL(\lan x,y\ran)=\{\lan R,
                        \lhd^-,1-n\ran\}$, $\cL(y)=\{\lan C,\lhd,n\ran\}$\vspace{0.2cm}\\

$(\forall_{\rhd})$ & if 1.& $\lan \forall R.C,\rhd,n\ran\in\cL(x)$, $x$ is not indirectly blocked, and\\
                      &    2.& $x$ has an
%Jeff 29/07/05
                      \rcn R {\rhd'} {n_1}-neighbour $y$ with $\lan C,\rhd,n\ran\not\in\cL(y)$
                      and\\
                      &    3.& $\tup{x,y}$
                      %R^*,
                      conjugates with
                      $\lan R, \rhd^-,1-n\ran$\\
                      %a positive triple
%Jeff 17/07/05
                      %that connects $x$ and $y$
                      %in $\cL(\tup{x,y})$\\
                      & then & $\cL(y)\rightarrow\cL(y)\cup \{\lan C,\rhd,n\ran\}$\vspace{0.2cm}\\

$(\exists_{\lhd})$ & if 1.& $\lan \exists R.C,\lhd,n\ran\in\cL(x)$, $x$ is not indirectly blocked and\\
%Jeff 29/07/05
                      &    2.& $x$ has an \rcn R \rhd {n_1}-neighbour $y$ with $\lan C,\lhd,n\ran\not\in\cL(y)$ and\\
                      &    3.& $\tup{x,y}$
                      % R^*,
                      conjugates with
                      %a positive triple
%Jeff 17/07/05
                       %that connects $x$ and $y$\\
                       %in $\cL(\tup{x,y})$\\
                      $\lan R,\lhd,n\ran$\\
                      & then & $\cL(y)\rightarrow\cL(y)\cup \{\lan C,\lhd,n\ran\}$\vspace{0.2cm}\\

$(\forall_{+})$ & if 1.& $\lan \forall R.C,\rhd,n\ran\in\cL(x)$
%Jeff 17/07/05
with
%\Tr(R),
$\Tr(R)$, $x$ is not indirectly blocked, and\\
                      &    2.& $x$ has an
%Jeff 29/07/05
                      \rcn R {\rhd'} {n_1}-neighbour $y$ with $\lan \forall R.C,\rhd,n\ran\not\in\cL(y)$ and\\
                      &    3.& $\tup{x,y}$
                      %R^*,
                      conjugates with
                      $\lan R, \rhd^-,1-n\ran$ \\
                      %a positive triple
                      %that connects $x$ and $y$\\
                      %in $\cL(\tup{x,y})$\\
                      & then & $\cL(y)\rightarrow\cL(y)\cup \{\lan \forall R.C,\rhd,n\ran\}$\vspace{0.2cm}\\

$(\exists_{+})$ & if 1.& $\lan \exists R.C,\lhd,n\ran\in\cL(x)$
%Jeff 17/07/05
% \Tr(R)
with $\Tr(R)$, $x$ is not indirectly blocked and\\
                      &    2.& $x$ has an
%Jeff 29/07/05
                      \rcn R {\rhd} {n_1}-neighbour $y$ with $\lan \exists R.C,\lhd,n\ran\not\in\cL(y)$ and\\
                      &    3.& $\tup{x,y}$
                      %R^*
                      conjugates with
                      $\lan R,\lhd,n\ran$\\
                      %a positive triple
                      %that connects $x$ and $y$\\
                      %in $\cL(\tup{x,y})$\\
                      & then & $\cL(y)\rightarrow\cL(y)\cup
                       \{\lan \exists R.C,\lhd,n\ran\}$\vspace{0.2cm}\\
\hline
\end{tabular}}
\end{center}
\caption{
%Jeff 17/07/05
%Tableaux expansion
\fkdsi completion rules} \label{tableau}
\end{table*}

\begin{definition}[Tableaux Algorithm]\label{SIta}
%Jeff 17/07/05
For an \fkdsi ABox \A, the algorithm
%initializes
initialises a forest $\Forest$ to contain (i)~a root node
$x_{a_i}$, for each individual $a_i\in\Individuals_\A$ occurring
in \A, labelled with $\cL(x)$ such that
% and additionally
$\{\lan C_i,\bow,n\ran\}\Sub\cL(x_{a_i})$  for each assertion of
the form $(a_i:C_i)\bow n$ in \A, and (ii)~an edge $\lan
x_{a_i},x_{a_j}\ran$,
%if \A contains
for each assertion $(\tup{a_i,a_j}:R_i)\bow n$ in \A, labelled
with $\cL(\lan x_{a_i},x_{a_j}\ran)$ such that $\{\lan
R_i,\bow,n\ran\}\Sub\cL(\lan x_{a_i},x_{a_j}\ran)$.
% for each assertion
%of the form $\lan (a_i,a_j):R_i\bow n\ran$ in \A.
Moreover, the algorithm expands \R by adding an axiom
$\Tr(\Inv(R))$ for each $\Tr(R)\in\R$. $\Forest$ is then expanded
by repeatedly applying the completion rules from Table
\ref{tableau}.
%George 13/07/05 added "Note that"
%Note that
%where we use
%the notation
%$R^*$ to denote either the role $R$ or the role returned by
%$\Inv(R)$.
 The completion forest is complete when, for some node $x$,
 $\cL(x)$ contains a clash, or none of the completion rules in Table~\ref{tableau}
 are  applicable.
The algorithm stops when a clash occurs; it answers `\A is
consistent \wrt \R' iff the completion rules can be applied in
such a way that they yield a complete and clash-free conpletion
forest, and `\A is inconsistent \wrt \R' otherwise.
\end{definition}

%George Remark on models 28/09/05
There are some remarks regarding Definition \ref{SIta}. The
expansion rules are based on the properties of the semantics
presented in Definition \ref{SItableau}. For example, consider the
$(\forall_\rhd)$-rule. Now, if $\lan \forall
R.C,\geq,0.7\ran\in\cL(x)$, and $\lan R,\geq, 0.6\ran\in\cL(\lan
x,y\ran)$, this means that the last triple violates property 7 of
Definition \ref{SItableau}. This property says that the membership
degree of the edge $\lan x,y\ran$ to the role $R$ should be lower
or equal than the degree $1-0.7$, otherwise the membership degree
of $y$ to $C$ should be greater or equal than $0.7$. Interpreted
to membership triples this means that if a triple of the form
$\lan R,\geq, n\ran$ exists in $\cL(\lan x,y\ran)$, then $n\leq
1-0.7$, or if the triple is of the form $\lan R,>,n\ran$, then
$n<1-0.7$. In order to discover if these restrictions are violated
the $(\forall_\rhd)$-rule compares the triples of the edge $\lan
x,y\ran$ with the artificial triple $\lan R,\leq, 0.3\ran$ against
conjugation. In the present case conjugation occurs, thus we
should add the triple $\lan C,\geq,0.7\ran$ to the label of $y$.
Similar arguments hold for the rest of the properties. Please note
that artificial triples are not added in the completion-forest but
are only used to perform checks on membership degrees.
%Jeff 17/07/05
%George 28/09/05
Secondly, in the above tableaux algorithm, we see that we are
dealing with finite number of membership degrees.
%Stoilos 13/07/06 because of the fuzzy operations used in the \fkdsi DL.
%, i.e., the G\"odel t-norm
% $(t(a,b)=\min(a,b))$, the   G\"odel t-conorm
%$(u(a,b)=\max(a,b))$  and the Lukasiewicz negation $(c(a)=1-a)$.
In fact, from Table \ref{tableau}, we can see that for an
arbitrary
%expression
fuzzy assertion  of the form
%$(x:D)\geq n$
$(x:D)\bow n$ either value $n$ or its complement $c(n)$
%are present
appear in the expansion of a node $x$ where $\lan
D,\bow,n\ran\in\cL(x)$. The finite property of the membership
degrees makes blocking possible in our algorithm.
%Stoilos 13/07/06
This property is a consequence of the fuzzy operations used in our
context, i.e. the G\"odel t-norm and t-conorm, the Lukasiewicz
complement and Kleene-Dienes fuzzy implication and it usually does
not hold for other combinations of fuzzy operations. Finding an
appropriate blocking condition when other norms are used in
combination with Description Logics that include transitive
relations is an open research issue. Finally, it is worth noting
that since we assume all concepts to be in their negation normal
form the $(\neg_\bow)$-rule only applies to concept names. But,
since we employ a rule for handling negated concepts this is not
absolutely necessary in fuzzy DLs. Hence, we are able to not
produce the NNF form of negated concepts and apply the
$(\neg_\bow)$-rule directly on them. This might be the base for
optimization, since we might be able to identify clashes earlier,
or for generalizations to other norm operations, since in that
case we might not be able to produce the NNF of negated concepts.
In either case, the proof of lemma \ref{SIsoundness} would require
a slight modification in order to correctly interpret negated
concepts.

\begin{example}
Let us see some examples of applications of expansion rules.

\begin{itemize}
    \item $(\forall_\geq)$: Let $\lan \forall
    R.C,\geq,0.7\ran\in\cL(x)$ and $\lan
    \Inv(R),>,0.3\ran\in\cL(\lan y,x\ran)$. According to the
    definition of an $R$-neighbour, $y$ is an
    $R_{>,0.3}$-neighbour, hence $\lan x,y\ran$ conjugates with
    $\lan R,\leq,0.3\ran$, and additionally $\lan C,\geq,0.3\ran\not\in\cL(y)$.
    Thus, we should add $\lan C,\geq,0.3\ran$ in
    $\cL(y)$.\vspace{0.1cm}

    \item $(\exists_\geq)$: Let $\lan \exists
    \Inv(R).C,\geq,0.7\ran\in\cL(x)$. Then create a new node $y$
    in the forest and set, $\lan \Inv(R),\geq,0.7\ran\in\cL(\lan
    x,y\ran)$, $\lan C,\geq,0.7\ran\in\cL(y)$.\vspace{0.1cm}

    \item $(\exists_+)$: Let $\lan \exists
    \Inv(R).C,<,0.5\ran\in\cL(x)$, $\lan
    \Inv(R),\geq,0.7\ran\in\cL(\lan
    x,y\ran)$ and $\Tr(R)$. $y$ is an
    $\Inv(R)_{\geq,0.7}$-neighbour of $x$, hence
    $\lan x,y\ran$ conjugates with $\lan
    \Inv(R),<,0.5\ran$, and additionally $\lan \exists
    \Inv(R).C,<,0.5\ran\not\in\cL(y)$. Hence, $\lan \exists
    \Inv(R).C,<,0.5\ran$ should be added in $\cL(y)$.
\end{itemize}

\diae
\end{example}

%Jeff 30/11/05
Now we can revisit example \ref{example:reduction} to see how the
procedure presented in this section can be used to determine the
%George 13/04/06 unsatisfiability
consistency of the ABox.

%Jeff 30/11/05: replace o_i with x_{\indv o_i} and replace Inv with ^-
\begin{example}
Recall that our fuzzy ABox was $\A=\{(\tup{ \indv o_1, \indv
o_2}:\mathsf{isPartOf})\geq 0.8$, $(\tup{ \indv o_2, \indv
o_3}:\mathsf{isPartOf})\geq 0.9$, $(\indv o_2:\mathsf{Body})\geq
0.85$ and $(\indv o_1:\mathsf{Arm})\geq 0.75\}$, and that we
wanted to test the
%George 13/04/06 satisfiability
consistency of the fuzzy ABox $\A'=\A\cup\{(\indv o_3:\exists
\mathsf{\Inv(isPartOf)}. \mathsf{Body}\sqcap \exists
\mathsf{\Inv(isPartOf)}.\mathsf{Arm})<0.75\}$, \wrt
$\R=\{\Tr(\mathsf{isPartOf})\}$. According to Definition
\ref{SIta} the algorithm initializes a completion-forest to
contain the following triples (note that we have a node $x_{\indv
o_i}$ for each individual $\indv o_i$):

\begin{center}
\begin{tabular}{cll}
$(1)$ & $\lan \mathsf{isPartOf},\geq,0.8\ran\in\cL(\lan x_{\indv  o_1},x_{\indv o_2}\ran)$ &\\
$(2)$ & $\lan \mathsf{isPartOf},\geq,0.9\ran\in\cL(\lan x_{\indv o_2},x_{\indv o_3}\ran)$ &\\
$(3)$ & $\lan \mathsf{Body},\geq,0.85\ran\in\cL(x_{\indv o_2})$ &\\
$(4)$ & $\lan \mathsf{Arm},\geq,0.75\ran\in\cL(x_{\indv o_1})$ &\\
$(5)$ & $\lan \exists \mathsf{isPartOf^-}. \mathsf{Body}\sqcap
\exists\mathsf{isPartOf^-}.\mathsf{Arm},<,0.75\ran\in\cL(x_{\indv o_3})$ &\\
\end{tabular}
\end{center}

\noindent Furthermore, the algorithm expands \R by adding the
axiom $\Tr(\con{isPartOf}^-)$. Please note that for simplicity we
have not expanded the concepts $\con{Arm}$ and $\con{Body}$ in the
membership triples. Subsequently, by applying expansion rules from
Table \ref{tableau} we have the following steps:

\begin{center}
\begin{tabular}{cll}
$(6)$ & $\lan
\exists\mathsf{isPartOf^-}.\mathsf{Body},<,0.75\ran\in\cL(x_{\indv
o_3})
 \mid \lan \exists\mathsf{isPartOf^-}.\mathsf{Arm},<,0.75\ran\in\cL(x_{\indv o_3})$ & $(\sqcap_<)$\\
\end{tabular}
\end{center}
Hence at this point we have two possible completion forests. For
the first one we have,

\begin{center}
\begin{tabular}{cll}
$(6_1)$ & $\lan\exists\mathsf{isPartOf^-}.\mathsf{Body},<,0.75\ran\in\cL(x_{\indv o_3})$ &\\
$(7_1)$ & $\lan\mathsf{Body},<,0.75\ran\in\cL(x_{\indv o_2})$ & $(\exists_<):(6_1),(2)$\\
$(8_1)$ & clash $(7_1)$ and $(3)$
\end{tabular}
\end{center}
while for the second possible completion-forest we have.
\begin{center}
\begin{tabular}{cll}
$(6_2)$ & $\lan \exists\mathsf{isPartOf^-}.\mathsf{Arm}),<,0.75\ran\in\cL(x_{\indv o_3})$ &\\
$(7_2)$ & $\lan\mathsf{Arm},<,0.75\ran\in\cL(x_{\indv o_2})$ & $(\exists_<):(6_2),(2)$\\
$(8_2)$ & $\lan \exists\mathsf{isPartOf^-}.\mathsf{Arm}),<,0.75\ran\in\cL(x_{\indv o_2})$ & $(\exists_+):(6_2),(2)$\\
$(9_2)$ & $\lan\mathsf{Arm},<,0.75\ran\in\cL(x_{\indv o_1})$ & $(\exists_<):(8_2),(1)$\\
$(10_2)$ & clash $(9_2)$ and $(4)$
\end{tabular}
\end{center}
Thus, since all possible expansions result to a  clash, $\A'$ is
%George 13/04/06 unsatisfiable
inconsistent
%and
the knowledge base entails the fuzzy assertion.

\end{example}

\begin{example}
Consider the fuzzy knowledge base $\kb=\tup{\T,\A,\R}$, with TBox
$\T=\{C\equiv \forall R^-.(\forall P^-.\neg A)\}$, ABox
$\A=\{(\indv a:A)\geq 0.8, (\tup{\indv a,\indv b}:P)\geq 0.8,
(\indv b:C)\geq 0.8, (\indv b:\exists R.C)\geq 0.8, (\indv
b:\forall R.(\exists R.C))\geq 0.8\}$ and RBox $\R=\{\Tr(R)\}$.
%Stoilos 13/07/06
First the algorithm expands \R by adding axiom $\Tr(R^-)$. Then,
in order to check the consistency of \A \wrt \T and \R the
algorithm initializes the following completion-forest:

\begin{center}
\begin{tabular}{cll}
$(1)$ & $\lan A,\geq,0.8\ran\in\cL(x_{\indv a})$ &\\
$(2)$ & $\lan C,\geq, 0.8\ran\in\cL(x_{\indv b})$ &\\
$(3)$ & $\lan \exists R.C,\geq,0.8\ran\in\cL(x_{\indv b})$ &\\
$(4)$ & $\lan \forall R.(\exists R.C),\geq,0.8\ran\in\cL(x_{\indv b})$ &\\
$(5)$ & $\lan P,\geq, 0.8\ran\in\cL(x_{\indv a}, x_{\indv b})$.
\end{tabular}
\end{center}

\noindent Then, we get the following application of expansion
rules,

\begin{center}
\begin{tabular}{cll}
$(6)$ & $\lan C,\geq,0.8\ran\in\cL(x_{\indv o_1}), \lan R,\geq, 0.8\ran\in\cL(x_{\indv b}, x_{\indv o_1})$ & $(\exists_\geq): (3)$\\
$(7)$ & $\lan \exists R.C,\geq,0.8\ran\in\cL(x_{\indv o_1})$ & $(\forall_\geq): (4)$\\
$(8)$ & $\lan \forall R.(\exists R.C),\geq,0.8\ran\in\cL(x_{\indv o_1})$ & $(\forall_+): (4)$\\
\end{tabular}
\end{center}
As we can see $\cL(x_{\indv b})=\cL(x_{\indv o_1})$, hence
$x_{\indv o_1}$ is blocked by $x_{\indv b}$. On the other hand it
is not indirectly blocked. Hence, since $\lan \forall R^-.(\forall
P^-.\neg A),\geq,0.8\ran\in\cL(x_{\indv o_1})$ (due to the
definition of $C$ in the TBox) we have the following application
of expansion rules,
\begin{center}
\begin{tabular}{cll}
$(9)$ & $\lan \forall P^-.\neg A,\geq,0.8\ran\in\cL(x_{\indv b})$& $(\forall_\geq): (6)$\\
$(10)$ & $\lan \neg A,\geq,0.8\ran\in\cL(x_{\indv a})$ & $(\forall_\geq): (9)$\\
$(11)$ & $\lan A,\leq,0.2\ran\in\cL(x_{\indv a})$ & $(\neg_\geq): (10)$\\
$(12)$ & clash $(11)$ and $(1)$
\end{tabular}
\end{center}
Please note that adding $\lan \forall P^-.\neg A,\geq,0.8\ran$ to
$\cL(x_{\indv b})$ causes the blocking of node $x_{\indv o_1}$ to
be broken since it no longer holds that $\cL(x_{\indv
b})=\cL(x_{\indv o_1})$. Hence, the notions of indirectly blocked
nodes and dynamic blocking are crucial in the presence of inverse
roles in order to correctly identify consistent and inconsistent
ABoxes. Also note that if the algorithm had chosen to expand
$x_{\indv o_1}$ (since this node is no more blocked) rather than
$x_{\indv b}$, then it would have created another node, say
$x_{\indv o_2}$, for which $\cL(x_{\indv o_1})=\cL(x_{\indv
o_2})$. Then again $\lan C,\geq, 0.8\ran$ would be added to
$x_{\indv o_1}$, since $x_{\indv o_2}$ would not be indirectly
blocked, the block on $x_{\indv o_2}$ would be broken, but then it
would hold that $\cL(x_{\indv b})=\cL(x_{\indv o_1})$. Hence
$x_{\indv o_1}$ would be permanently blocked while $x_{\indv o_2}$
indirectly blocked. Then the algorithm would have no other choice
but to identify the clash in node $x_{\indv a}$, as it is showed
in steps (9) to (12).
\end{example}

%Jeff 18/06/05
%\subsection{Soundness and Completeness}
\subsection{Decidability of \fkdsi}

The soundness and completeness of the algorithm will be
demonstrated by proving that for an \si ABox \A, it always
terminates and that it returns \emph{consistent} iff \A is
consistent.

%George 13/07/05 added "and RBox \R" and ", when started for \A and \R"
\begin{lemma}
(\textbf{Termination}) For each \fkd-\si ABox \A and RBox \R, the
tableaux algorithm terminates, when started for \A and \R.
\end{lemma}

\begin{Proof} Let $m=|sub(\A)|$,
%Jeff 29/07/05
$k=|\bR_\A|$ and $l$ be the number of different membership degrees
appearing in \A . Obviously $m$  and $l$ are
%George 13/07/05 added "and $l$"
linear in the length of \A. Termination is a consequence of the
following properties of the expansion rules:

\begin{enumerate}
    \item The expansion rules never remove nodes from the forest or
    concepts from node labels.

%Jeff 29/07/05
    \item Only the $(\exists_{\rhd})$- or the $(\forall_{\lhd})$-rule generate new
    nodes, and each generation is triggered when \tup{\exists
    R.C,\rhd,n} or  \tup{\forall
    R.C,\lhd,n} is in a node label where $\exists
    R.C$ or $\forall
    R.C$ is in $sub(\A)$. As no nodes can be removed, these rules
    will not be applied on the same label repeatedly.
    %Successors are only generated for triples of the form
    %$\lan \exists R.C,\rhd,n\ran$ and $\lan \forall R.C,\lhd,n
    %\ran$ and for any node only once.
    Since $sub(\A)$ contains at
    most $m$
%Jeff 29/07/05
  %  assertions, and hence also triples, of the form
  %  $\lan a:\exists R.C\rhd n_1\ran$ and $\lan a:\forall R.C\lhd
  %  n_2\ran$
$\exists R.C$  or  $m$ $\forall R.C$,
  the out-degree of the forest is bounded by $2ml$.

%Jeff 29/07/05
    \item Nodes are labelled with triples of the form \tup{C, \bow,
    n}, so there are at most $2^{8ml}$  different possible
    labellings for a pair of nodes.
%    Furthermore up to $2l$ different values can appear in the
%    system ($l$ for the original values, plus $l$ for their
%    negations) and no other values can.
Thus, if a
    path $p$ is of length at least
    %Jeff 29/07/05
    % $l*2^m$
   $2^{8ml}$ (note that concepts that cause
    non-termination interact either with a value $n$ or with it's
    negation, and not with both),
    then there exist
%Jeff 29/07/05
    %2
    two nodes $x,y$ on $p$ which
contain the same
%Jeff 29/07/05
%    positive assertions.
label. Since a path on which nodes are blocked
    cannot become longer, paths are of length at most
%Jeff 29/07/05
     $2^{8ml}$.
\end{enumerate}
\qed
\end{Proof}
As the previous lemma suggests the tableaux algorithm runs in
exponential space. This is due to a well-known problem inherited
from the crisp \si language \cite{Tobies01}. Consider for example
the following concepts taken from \citeA{Tobies01},
\[C\equiv\exists R.D\sqcap \forall R.(\exists R.D)\]
\[D\equiv (A_1\sqcup B_1)\sqcap(A_2\sqcup B_2)\sqcap\ldots\sqcap(A_n\sqcup B_n)\]
where $R$ is a transitive role. Now consider that we want to check
the consistency of the fuzzy ABox $\A=\{(a:C)\geq n\}$. Concept
$C$ causes the generation of $\rcn R {\geq} {n}$-successors $b_i$
for which it also holds that $(b_i:D)\geq n$. Now due to the
$\sqcup_\geq$-rule, which might choose to add either
$(b_i:A_i)\geq n$ or $(b_i:B_i)\geq n$, there are $2^n$ possible
ways of expanding $D$. Hence, the algorithm might create a path of
exponential depth before blocking applies. \citeA{Tobies01}
presents an optimized blocking technique that leads to a \Pspace
algorithm for \si. This technique involves a refined blocking
strategy as well as the modification of the tableaux expansion
rules. Investigating the applicability of this technique to \fkdsi
is an interesting open problem.

%George 13/07/05 added "and RBox \R" and "w.r.t. \R"
\begin{lemma}\label{SIsoundness}
(\textbf{Soundness}) If the expansion rules can be applied to an
\fkd-\si ABox \A and
%Jeff 29/07/05
an RBox \R such that they yield a complete
and clash-free completion-forest, then \A has a fuzzy tableau
w.r.t. \R.
\end{lemma}
\begin{Proof}
Let $\mathcal{F}_A$ be a complete and clash-free completion-forest
constructed by the tableaux algorithm for \A. The construction of
a fuzzy tableau $T=(\bS,\cL,\cE,\cV)$ is based on the construction
of a fuzzy model, presented by \citeA{Straccia01}:

For a set of triples of the form $\lan A,\geq,n_i\ran$, $i$ a
positive integer, that might exist within a set of triples
$\cL(x)$, the maximum value of $n_i$'s is chosen as a membership
degree of $x$ to the fuzzy set $A^\I$, i.e. the degree $\cL(x,A)$
in our case. If the maximum value participates in a triple of the
form $\lan A,>,n\ran$ a small factor $\epsilon$ is added to the
maximum. The existence of such a value is ensured by the
clash-freeness of \Forest. Please also note that without loss of
generality we can force all factors $\epsilon$ to be equal.
Furthermore, when no triple of the form $\lan
C,\rhd,n_i\ran\in\cL(x)$ exists, while only triples $\lan
C,\lhd,n_i\ran\in\cL(x)$ do, the membership degree is set to 0. In
cases where a value or existential restriction exists as well as a
non conjugated relation, special care to the choice of $\epsilon$
has to be made in order not to choose a high value that causes a
conjugation in the interpretation. At last, in order to interpret
concepts of the form $\neg A$, where $A$ is a concept name, we
first compute the maximum degree of a node to the concept $A$ and
then use it to compute $\neg A$. The function that returns the
maximum degree is denoted by $glb$ \cite{Straccia01}. Please note
that the labellings $\cL(s,C)$ refer to nodes of the fuzzy
tableau, while those of $\cL(x)$ to nodes of the
completion-forest. A fuzzy tableau can be defined as follows:

%George 08/09/05
\begin{center}{%\footnotesize
\begin{tabular}{rcl}
\bS & \hspace{0.1cm} = \hspace*{0.1cm} & \{$x\mid x$ is a node in \Forest, and $x$ is not blocked\},\vspace{0.1cm}\\
$\cL(x,\bot)$         & = &   $0$, for all $x\in\bS$,\vspace{0.1cm}\\
$\cL(x,\top)$         & = &   $1$, for all $x\in\bS$,\vspace{0.1cm}\\
$\cL(x,C)$            & = & $glb[\lan C,\bow,n_i\ran]$, for $\lan C,\bow,n_i\ran\in\cL(x)$ $x$ not blocked,\vspace{0.1cm}\\
$\cL(x,\neg A)$       & = & $1-\cL(x,A)$, for all $x$ in \Forest not blocked, with $\lan \neg A,\bow,n\ran\in\cL(x)$,\vspace{0.1cm}\\
$\cE(R,\lan x,y\ran)$ & = &\{$glb[\lan R^*,\bow,n_i\ran] \mid$ \parbox[t]{20em}{1. $y$ is an \rcn R {\bow} {n_i}-neighbour of $x$ \emph{or}\\
                                                                    2.$\lan R,\bow,n_i\ran\in\cL(\lan x,z\ran)$ and $y$ blocks $z$ \emph{or}\\
                                                                    3.$\lan \Inv(R),\bow,n_i\ran\in\cL(\lan y,z\ran)$ and $x$ blocks $z$\},}\vspace{0.1cm}\\
$\cV(a_i)$ & = & $x_{a_i}$, where $x_{a_i}$ is a root node,\\
\end{tabular}}
\end{center}
where $R^*$ represents either $R$ or $\Inv(R)$. It can be shown
that $T$ is a fuzzy tableau for \A \wrt \R:

\begin{enumerate}
    \item Property 1 of Definition \ref{SItableau} is satisfied
    due to the construction of $T$ and because \Forest is
    clash-free.\vspace{0.1cm}

    \item Property 2 of Definition \ref{SItableau} is satisfied
    because the $\neg$-rule does not apply and we force all
    factors $\epsilon$ to be equal. Let $\cL(x,\neg
    A)=n_1\geq n$. The definition of $T$ implies that $1-n \geq
    1-n_1=\cL(x,A)$.\vspace{0.1cm}

    \item Properties 3-6 of Definition \ref{SItableau} are
    satisfied because none of $\sqcup_{\bow}$ nor $\sqcap_{\bow}$
    apply to any $x\in\bS$. For example, let $\cL(x,C\sqcap
    D)=n_1\geq n$. The definition of $T$ implies that, either
    $\lan C\sqcap D,\geq,n_1\ran\in\cL(x)$ or $\lan
    C\sqcap D,>,n'\ran\in\cL(x)$, with $n_1=n'+\epsilon$.
    Completeness of \Forest implies that either $\lan
    C,\geq,n_1\ran\in\cL(x)$ and $\lan D,\geq,n_1\ran\in\cL(x)$ or
    $\lan C,>,n'\ran\in\cL(x)$ and $\lan
    D,>,n'\ran\in\cL(x)$. Hence, $\cL(s,C)=glb[\lan
    C,\bow,n_i\ran]\geq \cL(s,C\sqcap
    D)\geq n$, $\cL(s,D)=glb=[\lan C,\bow,n_i\ran]\geq
    \cL(s,C\sqcap D)\geq n$. The rest of
    properties follow in a similar way. \vspace{0.1cm}

    \item Property 7 in Definition \ref{SItableau} is satisfied.
    Let $x\in\bS$ with $\cL(x,\forall
    R.C)=n_1\geq n$ and $\cE(R,\lan x,y
    \ran) \ngeq^- 1-n$. The definition of $T$ implies that
    either $\lan \forall R.C,\geq,n_1\ran\in\cL(x)$
    or $\lan \forall R.C,>,n'\ran\in\cL(x)$ with $n_1=n'+\epsilon$.
    Moreover, since the $glb$ function does not
    create an unnecessary conjugation we have that either:
        \begin{enumerate}
            \item $y$ is an $R_{\rhd,r}$-neighbour of $x$
            \item $\lan R,\rhd,r\ran\in\cL(\lan x,z\ran)$, $y$
%George 13/07/05 added "thus $\cL(y)=\cL(z)$" and removed "the same positive...".
            blocks $z$ thus $\cL(y)=\cL(z)$, or
            \item $\lan \Inv(R),\rhd,r\ran\in\cL(\lan y,z\ran)$,
            $x$ blocks $z$, thus $\cL(x)=\cL(z)$.
        \end{enumerate}
    and $\lan R,\rhd,r\ran$ or $\lan \Inv(R),\rhd,r\ran$
    causes conjugation. Hence, in all 3 cases, the
    $\forall_{\rhd}$-rule ensures that either $\lan
    C,\geq,n_1\ran\in\cL(y)$ or $\lan C,>,n'\ran\in\cL(y)$.
    Thus, either $\cL(y,C)\geq n_1\geq n$, or $\cL(y,C)\geq
    n'+\epsilon=n_1\geq n$.
    The case with $\cL(x,\forall R.C)>n$ and $\cL(x,\exists
    R.C)\lhd n$, where the latter regards property 8, are
    shown in a similar way.\vspace{0.1cm}

    \item Property 9 in Definition \ref{SItableau} is satisfied.
    Let $x\in\bS$ with $\cL(x,\exists
    R.C)=n_1\geq n$. The definition of $T$ implies that
    either $\lan \exists R.C,\geq,n_1\ran\in\cL(x)$
    or $\lan \exists R.C,>,n'\ran\in\cL(x)$, with $n_1=n'+\epsilon$.
    Then the $\exists_{\rhd}$-rule ensures that there is either:
        \begin{enumerate}
            \item a predecessor $y$ such that $\lan
            \Inv(R),\geq,n_1\ran\in\cL(\lan y,x\ran)$
            and $\lan C,\geq,n_1\ran\in\cL(y)$ or dually with $>$ and
            $n'$. Because $y$ is a predecessor of $x$ it cannot be
            blocked, so $y\in\bS$, $\cE(R,\lan x,y
            \ran)\geq n_1\geq n$ and $\cL(y,C)\geq n_1$ or $\cE(R,\lan x,y
            \ran)\geq n'+\epsilon=n_1\geq n$ and $\cL(y,C)\geq n'+\epsilon=n_1\geq n$.

            \item a successor $y$ such that $\lan
            R,\geq,n\ran\in\cL(\lan x,y \ran)$, $\lan
            C,\geq,n\ran\in\cL(y)$ or dually with $>$ and $n'$. If
            $y$ is not blocked, then $y\in\bS$ and $\cE(R,\lan x,y
            \ran)\geq n_1$, $\cL(y,C)\geq n_1$ or $\cE(R,\lan x,y
            \ran)\geq n'+\epsilon$, $\cL(y,C)\geq
            n'+\epsilon=n_1$. Otherwise, $y$ is
            blocked by some $z$. Hence, $z\in\bS$ and $\lan
            R,\geq,n_1\ran\in\cL(\lan
            x,z\ran)$, $\lan C,\geq,n_1\ran\in\cL(z)$ or $\lan
            R,>,n'\ran\in\cL(\lan
            x,z\ran)$, $\lan C,>,n\ran\in\cL(z)$. In both cases
            $\cL(z,C)\geq n$ and $\cE(R,\lan x,z\ran)\geq n$.
        \end{enumerate}
    Similar proof applies for $\cL(x,\exists R.C)> n$ and also
    for Property 10 with $\cL(x,\forall R.C)\lhd n$.\vspace{0.1cm}

    \item Property 12 in Definition \ref{SItableau} is satisfied.
    Let $x\in\bS$ with $\cL(x,\forall
    R.C)=n_1\geq n$ and $\cE(R,\lan x,y
    \ran) \not\geq^- 1-n$. The definition of $T$ implies that
    either $\lan \forall R.C,\geq,n_1\ran\in\cL(x)$
    or $\lan \forall R.C,>,n'\ran\in\cL(x)$ with $n_1=n'+\epsilon$.
    Moreover, since the $glb$ function does not
    create an unnecessary conjugation we have that either:
        \begin{enumerate}
            \item $y$ is an $R_{\rhd,r}$-neighbour of $x$
            \item $\lan R,\rhd,r\ran\in\cL(\lan x,z\ran)$, $y$
%George 13/07/05 added "thus $\cL(y)=\cL(z)$" and removed "the same positive...".
            blocks $z$ thus $\cL(y)=\cL(z)$, or
            \item $\lan \Inv(R),\rhd,r\ran\in\cL(\lan y,z\ran)$,
            $x$ blocks $z$, thus $\cL(x)=\cL(z)$.
        \end{enumerate}
    and $\lan R,\rhd,r\ran$ or $\lan \Inv(R),\rhd,r\ran$
    causes conjugation. Hence, in all 3 cases, the
    $\forall_{+}$-rule ensures that either $\lan
    \forall R.C,\geq,n_1\ran\in\cL(y)$ or $\lan \forall R.C,>,n'\ran\in\cL(y)$.
    Thus, either $\cL(y,\forall R.C)\geq n_1\geq n$, or $\cL(y,\forall R.C)\geq
    n'+\epsilon=n_1\geq n$.
    The case with $\cL(x,\forall R.C)>n$ and $\cL(\exists
    R.C,\lhd,n\ran$, where the latter regards property 11, are
    shown in a similar way.\vspace{0.1cm}

    %George 13/07/05 changed \geq to \bow
    \item Property 13 in Definition \ref{SItableau} is satisfied
    because, if $\cE(R,\lan x,y \ran)\bow n$, then either:
        \begin{enumerate}
            \item $y$ is an $R_{\bow,n_1}$-neighbour of $x$, so $x$ is an
            $\Inv(R)_{\bow,n_1}$-neighbour of $y$.
            \item $\lan R,\bow,n_1\ran\in\cL(\lan x,z \ran)$, and $y$ blocks
            $z$, so $\lan \Inv(\Inv(R)),\bow,n_1\ran\in\cL(\lan x,z \ran)$
            \item $\lan \Inv(R),\bow,n_1\ran\in\cL(\lan y,z \ran)$ and $x$
            blocks $z$.
        \end{enumerate}
        In all 3 cases, $\cE(\Inv(R),\lan y,x \ran)\bow n$.

%    \item Property 13 in definition \ref{SItableau} is satisfied
%    because $\mathcal{F}_A$ is clash-free.

    \item Properties 14 and 15 are satisfied cause of the
    initialization of the completion-forest and the fact that the
    algorithm never blocks root nodes.
\end{enumerate}
\qed
\end{Proof}

%George 13/07/05 added "w.r.t \R" and several other
\begin{lemma}\label{SIcompleteness}
\textbf{(Completeness)} Let \A be an \fkd-\si ABox and \R an RBox.
If \A has a fuzzy tableau w.r.t. \R, then the expansion rules can
be applied in such a way that the tableaux algorithm yields a
complete and clash-free completion-forest for \A and \R.
\end{lemma}
\begin{Proof} Our proof of completeness is based on the proof for
crisp DLs presented by \citeA{Horrocks99} and \citeA{Horrocks00}.

Let $T=(\bS,\cL,\cE,\cV)$ be a fuzzy tableau for \A. Using $T$, we
trigger the application of the expansion rules such that they
yield a completion-forest $\Forest$ that is both complete and
clash-free.

Since we know that \A has a fuzzy tableau ($T$) we can steer the
application of rules such that they yield a complete and
clash-free completion-forest. \citeA{Horrocks99} and
\citeA{Horrocks00} define a mapping $\pi$ which maps nodes of
$\Forest$ to elements of \bS, and guide the application of the
non-deterministic rules $\sqcup_{\rhd}$ and $\sqcap_{\lhd}$. Our
method differs from the one used in crisp DLs \cite{Horrocks99} in
the following way. Using the membership degree of a node to a
concept, found in the fuzzy tableau, we create artificial triples
which are tested against conjugation with the candidate triples
that the non-deterministic rules can insert in the
completion-forest. The triples that don't cause a conjugation can
be added. The modified rules, which are used to guide such an
expansion, are presented in Table \ref{table:modrules}.

%George 26/05/05 I changed some wrongly inserted parenthesis "(" ")"
%in the second rule to right and left angles
\begin{table}[ht]
\begin{center}\small
\begin{tabular}{crl}
$(\sqcup'_{\rhd})$  & if 1.& $\lan C_1\sqcup C_2,\rhd,n\ran\in\cL(x)$, $x$ is not indirectly blocked, and\\
                        &    2.& $\{\lan C_1,\rhd,n\ran,\lan C_2,\rhd,n\ran\}\cap\cL(x)=\emptyset$\\
                        & then & $\cL(x)\rightarrow\cL(x)\cup\{C\}$ for some $C\in\{\lan C_1,\rhd,n\ran,\lan C_2,\rhd,n\ran\}$\\
                        &      & not conjugated with $\lan C_1,\leq,\cL(\pi(x),C_1)\ran$ or $\lan C_2,\leq,\cL(\pi(x),C_2)\ran$\vspace{0.1cm}\\

$(\sqcap'_{\lhd})$  & if 1.& $\lan C_1\sqcap C_2,\lhd, n\ran\in\cL(x)$, $x$ is not indirectly blocked, and\\
                        &    2.& $\{\lan C_1, \lhd, n\ran,\lan C_2, \lhd, n\ran\}\cap\cL(x)=\emptyset$\\
                        & then & $\cL(x)\rightarrow\cL(x)\cup\{C\}$ for some $C\in\{\lan C_1,\lhd,n\ran,\lan C_2,\lhd,n\ran\}$\\
                        &      & not conjugated with $\lan C_1,\geq,\cL(\pi(x),C_1)\ran$ or $\lan C_2,\geq,\cL(\pi(x),C_2)\ran$\vspace{0.1cm}\\

\end{tabular}
\end{center}
\caption{The $\sqcup'_{\rhd}$- and
$\sqcap'_{\lhd}$-rules}\label{table:modrules}
\end{table}

$\pi$ ensures
%George 13/07/05
that a new fuzzy assertion about the membership degree of a node
to a concept, created by a non-deterministic rule, is not more
restrictive than the one already known in the fuzzy tableau, thus
avoiding possible conjugations.
%the subset relation, as shown in
%\cite{Horrocks99,Horrocks00}, which thus keeps the
%completion-forest satisfiable without adding any unsatisfiable
%concept.
This together with the termination property ensure the
completeness of the algorithm. \qed
\end{Proof}

%George 26/05/05
%A fuzzy ABox \A \emph{entails} a fuzzy assertion $\lan \phi \geq
%n \ran$ iff $\A\cup \{\lan \phi < n\ran\}$ is inconsistent
%\cite{Straccia01}. Furthermore, an \fkd-\si concept $D$
%subsumes an \fkd-\si concept $C$ w.r.t. an empty $TBox$ \T (or
%a $TBox$ that can be reduced to an empty $TBox$) iff $\{\lan
%\phi:C\geq n\ran,\lan \phi:D<n\ran\}$ is inconsistent. Thus, most
%common inference problems of crisp DLs can be reduced to fuzzy
%DLs.

%Gior
\begin{theorem}\label{decision}
The tableaux algorithm is a decision procedure for the consistency
of \fkdsi $ABoxes$ and the satisfiability and subsumption of
\fkdsi concepts with respect to \emph{simple} terminologies.
\end{theorem}
Theorem \ref{decision} is an immediate consequence of lemmas
\ref{SItableau}, \ref{SIsoundness} and \ref{SIcompleteness}.
Moreover, as we discussed in section
%George 13/07/05 fixed \ref{}
\ref{sec:f-SI}, subsumption can be reduced to consistency checking
for $ABoxes$.

\section{Adding Role Hierarchies and Number Restrictions}\label{sec:f-SHIN}
In the current section we will provide the necessary extensions of
the reasoning algorithm presented in the previous section, in
order to provide reasoning support for the fuzzy DL language
\fkdshin. To achieve our goal we will extend the results of
section \ref{sec:transinvest} by also considering role
hierarchies, while we will also provide an investigation on the
number restrictions constructor.

In classical DLs, the results of transitive roles and value
restrictions obtained by \citeA{Sattler96}, were extended by
\citeA{Horrocks99} to also consider role hierarchies. More
precisely, they show that if $x\in(\forall R.C)^\I$, $\lan
x,y\ran\in P^\I$, $\Tr(P)$ and $P\sss R$, then $y\in(\forall
P.C)^\I$. In fuzzy DLs that also include role hierarchies we can
easily extend the results obtained in section
\ref{sec:transinvest}. Let $(\forall R.C)^\I(x)\geq c_a$,
$P^\I(x,y)=p$, $\Tr(P)$, and $c_a,p\in[0,1]$, and consider also
that $P\sss R$. Since $P$ is transitive, then $\forall
x,y\in\deltai$ and for some arbitrary $z\in\deltai$ it holds that,
$P\ifunc(x,y)\geq \min(P\ifunc(x,z),P\ifunc(z,y))$. Due to the
semantics of role inclusion axioms we have that $R\ifunc(x,y)\geq
\min(P\ifunc(x,z),P\ifunc(z,y))$. Then, if we work in a similar
way as in section \ref{sec:transinvest} we will get that,
$\max(c(P\ifunc(a,b)),(\all P C)\ifunc(b))\geq v_a$, which means
that either $c(P\ifunc(a,b))\geq v_a$ or $(\all P C)\ifunc(b)\geq
v_a$. A similar result can be obtained for the case where
$(\forall R.C)^\I(a)>n$. Hence, we get the following result:

\begin{corollary}\label{cor:trhierforall}
If $(\forall R.C)^\I(a)\rhd n$, and $\Tr(P)$ with $P\sss R$, then
in a \fkd-DL it holds that, $(\forall P.(\forall P.C))^\I(a)\rhd
n$.
\end{corollary}

Finally, for the case of negative assertions and existential
restrictions the following is easily obtained.

\begin{corollary}\label{cor:trhierexists}
If $(\exists R.C)^\I(a)\lhd n$, and $\Tr(P)$ with $P\sss R$, then
in a \fkd-DL it holds that, $(\exists P.(\exists P.C))^\I(a)\lhd
n$.
\end{corollary}

Now we will investigate fuzzy number restrictions. Although, from
in Table \ref{table:fkdsi-sem} it seems that the semantics of
number restrictions are quite complicated, we will see that
intuitively, they are quite similar to their crisp counterparts,
as long as we also consider membership degrees.

Consider for example the at-least restriction $(\geq pR)^\I(a)\geq
n$, where $a\in\deltai$. Then according to Table
\ref{table:fkdsi-sem} we have,
\[\sup_{b_1,\ldots,b_p\in\Delta^\I}\min^p_{i=1}\{R^\I(a,b_i)\}\geq
n.\]

\noindent This means that there must be at least $p$ pairs $\lan
a,b_i\ran$, for which $R^\I(a,b_i)\geq n$, holds. These semantics
are quite intuitive and similar with those of crisp number
restrictions. There one would require at least $p$ pairs for which
$R^\I(a,b_i)\geq 1$, which simply means more than $p$ pairs.
Similarly, we can work for $(\geq pR)^\I(a)>n$.

Consider now an at-most restriction of the form $(\leq
pR)^\I(a)\geq n$. Based on the semantics we have the inequation,
\[\inf_{b_1,\ldots,b_{p+1}\in\Delta^\I}\max^{p+1}_{i=1}\{1-R^\I(a,b_i)\}\geq
n.\]

\noindent This means that for all $p+1$ pairs $\lan a,b_i\ran$,
that can be formed, there is at least one pair for which
$c(R^\I(a,b_k))\geq n$, holds. We can also view this equation in a
different way which resembles that of crisp number restrictions.
From that perspective we can say that, there are at most $p$ pairs
$\lan a,b_i\ran$ for which $c(R^\I(a,b_i))<n$, holds. Similarly,
an at-most restriction of the form $(\leq pR)^\I(a)>n$ implies
that there are at-most $p$ pairs $\lan a,b_i\ran$, for which it
holds that $c(R^\I(a,b_i))\leq n$. Hence reasoning \wrt number
restrictions can be reduced to counting how many role assertions
$(\tup{a,b_i}:R)\geq n_i$ satisfy the above inequalities. If we
find that more than $p$ assertions satisfy these inequalities,
then we have to non-deterministically merge some of the individual
$b_i$, as is the case in the crisp \shin algorithm
\cite{Horrocks00}. Now, lets consider the extreme boundaries of 0
and 1, and apply our equation to the classical at-most
restriction, $a\in(\leq pR)^\I$. The fuzzy equivalent of this
assertions is $(\leq pR)^\I(a)\geq 1$, which implies that there
are at most $p$ $b_i\in\deltai$ such that,
$c(R^\I(a,b_i))<1\Rightarrow R^\I(a,b_i)>0$, holds. Since we are
only considering 0 and 1 the last inequality implies,
$R^\I(a,b_i)=1$, i.e. at-most $p$ successors of $a$ in $R^\I$.
%Hence, we see that fuzzy DLs are a \emph{sound} extension of crisp
%DLs, in the sense that at the extreme boundaries of 0 and 1 their
%semantics coincide.

Dually, we can also provide such intuitive meaning for the cases
which involve negative inequalities, like for example the cases of
$(\geq pR)^\I(a)\leq n_1$ or $(\leq pR)^\I(a)\leq n_2$. Applying
negation to the first equation we obtain, $(\neg (\geq
pR))^\I(a)\geq c(n_1)$, where $c$ is a fuzzy complement. Since the
min and max operations satisfy the De Morgan laws, this assertion
can be translated to $(\leq (p-1)R)^\I(a)\geq 1-n_1$, with $p\geq
1$, which is the negation normal form of the former assertion.
Similarly, the equation $(\leq pR)^\I(a)\leq n_2$ can be
transformed to the equivalent, $(\geq(p+1)R)^\I(a)\geq 1-n_2$.

Using the above results we can proceed in the definition of an
\fkdshin fuzzy tableau. Similarly to \ref{SItableau} we consider
all concepts to be in NNF. This can be achieved by using the
concept equivalences for number restrictions of section
\ref{sec:f-SI}. The definition of a fuzzy tableau for \fkdshin
first appeared by \citeA{Stoilos05d}, but here we have revised
that definition to better represent the properties of fuzzy
models. Before defining a fuzzy tableau for \fkdshin we extend the
definition of sub-concepts of a concept $D$ and an ABox \A.

\begin{definition}
For a fuzzy concept $D$ and a role hierarchy \R we define
$sub(D,\R)$ to be the smallest set of \fkdshin-concepts that
satisfies the following:
\begin{itemize}
    \item $D\in sub(D,\R)$,
    \item $sub(D,\R)$ is closed under sub-concepts of $D$, and
    \item if $\forall S.C\in sub(D,\R)$ and $R\sss S$, then $\forall R.C\in sub(D,\R)$
    \item if $\exists S.C\in sub(D,\R)$ and $R\sss S$, then $\exists R.C\in sub(D,\R)$
\end{itemize}
Finally, we define $sub(\A,\R)=\bigcup_{(a:D)\bow
n\in\A}sub(D,\R)$.
\end{definition}
When \R is clear from the context we will simply write $sub(\A)$.

\begin{definition}\label{SHINtableau}
If \A is an \fkdshin ABox, \R an \fkdshin RBox, $\bR_{\A}$ is the
set of roles occurring in \A and \R together with their inverses,
$\Individuals_\A$ is the set of individuals in \A, then a fuzzy
tableau $T$ for \A \wrt \R is defined as in Definition
\ref{SItableau} with the additional properties:

\begin{enumerate}

    \item[11'.] If $\cL(s,\exists R.C)\lhd n$, and $\Tr(P)$ with
    $P\sss R$, then $\cE(P, \lan s,t\ran)\lhd n$ or
    $\cL(t, \exists P.C)\lhd n$,\vspace{0.0cm}

    \item[12'.] If $\cL(s,\forall R.C)\rhd n$, and $\Tr(P)$ with
    $P\sss R$, then $\cE(P,\lan s,t\ran)\rhd^-1-n$ or
    $\cL(t,\forall P.C)\rhd n$,\vspace{0.0cm}

    \item[16.] If $\cE(R,\lan s,t\ran)\rhd n$ and $R\sss
    S$, then $\cE(S,\lan s,t\ran)\rhd n$,\vspace{0.0cm}

    \item[17.] If $\cL(s,\geq pR)\rhd n$, then $\sharp R^T(s,\rhd,n)\geq p$,\vspace{0.0cm}

    \item[18.] If $\cL(s,\leq pR)\lhd n$, then $\sharp R^T(s,\lhd^-,1-n)\geq p+1$,\vspace{0.0cm}

    \item[19.] If $\cL(s,\geq pR)\lhd n$, then $\sharp R^T_\neg(s,\lhd,n)\leq p-1$,\vspace{0.0cm}
%    conjugated with $\lan \lan s,t \ran, \lhd,n\ran$,\vspace{0.1cm}

    \item[20.] If $\cL(s,\leq pR)\rhd n$, then $\sharp R^T_\neg(s,\rhd^-,1-n)\leq p$,\vspace{0.0cm}

    \item[21.] If $a\ndoteq b\in\A$, then $\cV(a)\neq\cV(b)$

\end{enumerate}
where $R^T(s,\bow,n)=\{t\in\bS\mid \cE(R,\lan s,t\ran)\bow n\}$
returns the set of elements $t\in\bS$ that participate in $R$ with
some element $s$ with a degree, greater or equal, greater, lower
or equal or lower than $n$, and $R^T_\neg(s,\bow,n)=\{t\in\bS\mid
\cE(R,\lan s,t\ran)$\mbox{$\not\bow$} $n$\} returns those elements
that don't satisfy the given inequality.
\end{definition}

As in Definition \ref{SItableau}, we are based on the semantics of
the language and the observations made in the beginning of this
section about the properties of the value and existential
restrictions, when transitive roles and role hierarchies are
involved, and the semantic meaning of at-most and at-least number
restrictions. Thus, property 18 should be read as, if $\cL(s,\leq
pR)\geq n$ then there are at-most $p$ $t\in\bS$ such that
$\cE(R,\lan s,t\ran)\nleq 1-n$, i.e. $\cE(R,\lan s,t\ran)>1-n$,
and if $\cL(s,\leq pR)>n$, then there are at-most $p$ $t\in\bS$
such that $\cE(R,\lan s,t\ran)\not<1-n$.

\begin{lemma} \label{SHINsatisf}
An \fkdshin ABox \A is consistent w.r.t. \R, iff there exists a
fuzzy tableau for \A w.r.t. \R.
\end{lemma}
\begin{Proof}
The proof of the lemma is similar to that of lemma \ref{SIsatisf}
with some important technical details. For the ``if" direction, if
$T=(\bS,\cL,\cE,\cV)$ is a fuzzy tableau for \A w.r.t. \R, then a
model $\I=(\Delta,\cdot^\I)$ of \A and \R is constructed as
$\Delta^\I=\bS$, $a^\I=\cV(a)$, where $a\in\Individuals_\A$,
$\top^\I(s)=\cL(s,\top)$, $\bot^\I(s)=\cL(s,\bot)$ for all
$s\in\bS$, and $A^\I(s)=\cL(s,A)$, for all $s\in\bS$ and concept
names $A$, while for roles we have:

%\begin{table}[h]
\begin{center}
%\centering
{\hspace*{-15pt}
\begin{tabular}{rcl}
%$\Delta^\I$ & = & \bS\vspace{0.0cm}\\
%$a^\I$ & = & $\cV(a)$, $a\in\Individuals_\A$\vspace{0.0cm}\\
%$\top^\I(s)$ & = & $\cL(s,\top)$ for all $s\in\bS$\vspace{0.1cm}\\
%$\bot^\I(s)$ & = & $\cL(s,\bot)$ for all $s\in\bS$\vspace{0.1cm}\\
%$A^\I(s)$ & = & $\cL(s,A)$, for all $s\in\bS$ and concept names $A$\vspace{0.0cm}\\
$R^\I(s,t)$ & = &  $\left\{\begin{tabular}{ll} $R^+_{\cE}(s,t),$ & if $\Tr(R)$\vspace{0.1cm}\\
                                               $\undermax_{P\sss R,P\neq R}(R_\cE(s,t),P^\I(s,t))$ & otherwise\\
                            \end{tabular}\right.$\\
\end{tabular}}
\end{center}
%\end{table}
Observe that the interpretation of non-transitive roles is
recursive in order to correctly interpret those non-transitive
roles that have a transitive sub-role. From the definition of
$R^\I$ and property 12', if $R^\I(s,t)=n\in(0,1]$, then either
$\cE(R,\lan s,t\ran)=n$, or $\cE(R,\lan s,t\ran)=0$ and there
exist several paths $l\geq 1$ of the form,
\[\cE(P,\lan
s,s_{l_1}\ran)=p_{l_1}, \cE(P,\lan
s_{l_1},s_{l_2}\ran)=p_{l_2},\ldots,\cE(P,\lan
s_{l_m},t\ran)=p_{l_{m+1}}\]

\noindent with $\Tr(P)$, $P\sss R$ and $\cE(R,\lan
s,t\ran)=\max(0,\sup_l\{\min(p_{l_1},\ldots,p_{l_{m+1}})\})$.

Property 16 of $\I$ ensures that $\forall
s,t\in\deltai,P^\I(s,t)\leq R^\I(s,t)$ for all $P\sss R$. Again,
by induction on the structure of concepts we can we show that
$\cL(s,C)\bow n$ implies $C^\I(s)\bow n$ for any $s\in\bS$. Here,
we restrict our attention on to the cases that are different than
lemma \ref{SIsatisf}. Similarly to lemma \ref{SIsatisf} we also
restrict our attention to the inequalities $\geq$.

\begin{enumerate}

    \item[6'.] If $\cL(s,\forall R.C)\geq n$ and $R^\I(s,t)=p$,
    then either
    \begin{enumerate}
        \item $\cE(R,\lan s,t \ran)= p$, or

        \item $\cE(R,\lan s,t \ran)\neq p$. Then,
        there exist several paths $l\geq 1$ of the form,
        $\cE(P,\lan s,s_{l_1}\ran)=p_{l_1}$, $\cE(P,\lan
        s_{l_1},s_{l_2}\ran)=p_{l_2},\ldots,$ $\cE(P,\lan
        s_{l_m},t\ran)=p_{l_{m+1}},$ with $\Tr(P)$ and $P\sss R$.
        The membership degree $p$ of the pair $\lan s,t\ran$ to
        $(P^+)^\I$, would be equal to the maximum degree (since we
        cannot have infinite number of paths) of all the minimum
        degrees for each path. If that degree is such that it is
        not lower or equal than $1-n$ then there exists a path $k$
        where all degrees
        \[\cE(P,\lan
        s_{k_i},s_{k_{i+1}}\ran)=p_{k_i}, 0\leq i\leq k_m,
        s_{k_0}\equiv s, s_{k_{m+1}}\equiv t\]

        \noindent are not lower or
        equal than $1-n$, because all $p_{k_i}$'s would be greater
        or equal than the minimum degree of the path. Hence, due
        to property 11, we would have that $\cL(s_{k_i},\forall
        P.C)\geq n$, for all $1\leq i\leq k_m$.
    \end{enumerate}
    In case $p\leq 1-n$ we have then $\max(1-p,C^\I(t))\geq n$. In
    case $p\nleq 1-n$ then $\cL(t,C)\geq n$, so $C^\I(t)\geq n$
    and thus also $\max(1-p,C^\I(t))\geq n$. In both cases
    $(\forall R.C)^\I(s)\geq n$.\vspace{0.0cm}
    %The cases
    %$\cL(s,\forall R.C)> n$ and $\cL(s,\exists R.C)\lhd n$
    %can be shown similarly.\vspace{0.1cm}

    \item[7.] If $\cL(s,\geq pR)\geq n$ then we have, $\cE(R,\lan
    s,t_i\ran)\geq n$, $1\leq i\leq p$. By definition $R^\I(s,t_i)\geq n$,
    and thus
    \[n\leq\sup_{t_i\in\deltai}\{\ldots,\min^p_{i=1}\{R^\I(s,t_i)\},\ldots\}=(\geq
    pR)^\I(s).\]\vspace{0.0cm}
    %The cases $\cL(s,\geq pR)> n$ and $\cL(s,\leq
    %pR)\lhd n$ can be show similarly.

    \item[8.] If $\lan \leq pR,\geq,n\ran$ there are at
    most $p$ pairs $\lan s,t_i\ran$ for which, $\cE(R,\lan
    s,t_i\ran)\not\leq 1-n$, $1\leq i\leq p$.
    Thus in all $p+1$-tuples that can be
    formed there would be at least one pair $\lan s,t_{p+1}\ran$
    for which $\cE(R,\lan s,t_{p+1}\ran)\leq 1-n$ (even if
    $\cE(R,\lan s,t_{p+1}\ran)=0\leq 1-n$). Hence,
    $R^\I(s,t_{p+1})\leq 1-n\Rightarrow c(R^\I(s,t_{p+1}))\geq n$.
    Finally, we have that,
    \[n\leq\inf_{t_i\in\deltai}\{\ldots,\max(\max^p_{i=1}\{c(R^\I(s,t_i))\},
    c(R^\I(s,t_{p+1}))),\ldots\}=(\leq
    pR)^\I(s).\]
    %The cases $\cL(s,\leq pR)> n$ and
    %$\cL(s,\geq pR)\lhd n$ can be show similarly.
\end{enumerate}

For the converse, if \I=$(\Delta^\I,\cdot^\I)$ is a model for \A
w.r.t. \R, then a fuzzy tableau $T=(\bS,\cL,\cE,\cV)$ for \A and
\R is defined in exactly the same was an in lemma \ref{SIsatisf}.
Then,

\begin{enumerate}
    \item Properties 1-10 and 13 of Definition \ref{SItableau}
    and 16-20 in Definition \ref{SHINtableau} are satisfied as a
    direct consequence of the semantics of \fkdshin
    concepts.\vspace{0.0cm}

    \item Property 12' of Definition \ref{SHINtableau} is
    satisfied as a consequence of the semantics of transitive
    roles, role hierarchies and value restrictions that have
    been investigated in the beginning of the section. Hence,
    if $(\forall R.C)^\I(s)\geq n$, $P\sss R$ and $\Tr(P)$ then
    either $P^\I(s,t)\leq 1-n$, or $(\forall P.C)^\I(t)\geq n$
    holds, otherwise if $(\forall R.C)^\I(s)>n$, $P\sss R$ and
    $\Tr(P)$ then either $P^\I(s,t)<1-n$ or $(\forall
    P.C)^\I(t)>n$ holds. By definition of $T$ if $\cL(s,\forall
    R.C)\rhd n$, $P\sss R$ and $\Tr(P)$ then either $\cE(P,\lan
    s,t\ran)\rhd^- 1-n$ or $\cL(t,\forall P.C)\rhd n$. Similarly,
    for property 11' of Definition
    \ref{SHINtableau}.\vspace{0.0cm}

    \item $T$ satisfies Properties 14-15 of Definition
    \ref{SItableau} and Property 21 in Definition
    \ref{SHINtableau} because \I is a model of \A.

\end{enumerate}
\qed
\end{Proof}

\subsection{Constructing an \fkdshin Fuzzy Tableau}\label{sec:tableauxprocSHIN}
In this section we will show how the algorithm of \fkdsi,
presented in section \ref{sec:tableauxproc}, can be extended to
deal with \fkdshin ABoxes. There are a number of modifications
that need to be made, like the definition of $R$-neighbours, the
$(\forall_+)$- and $(\exists_+)$-rules, the blocking strategy, the
clash definition and the addition of rules for number
restrictions.

The most important modification from the algorithm of \fkdsi is
the blocking strategy. As it was noted by \citeA{Horrocks99} a DL
language that provides inverse roles, transitive role axioms, and
number restrictions lacks the \emph{finite-model} property; i.e.
there are \fkdshin-concepts that are satisfiable only in infinite
interpretations. This means that the usual blocking techniques
which create a cycle from the predecessor of a blocked node to the
blocking one, might fail to construct a correct tableau and due to
lemma \ref{SHINsatisf} a correct model. It is crucial to remark
here the difference between an infinite and a witnessed model, as
presented in remark \ref{rem:remark}. Although there are
\fkdshin-concept that are satisfiable in infinite interpretations,
these interpretations can still be witnessed \wrt the membership
degrees. The infinite or finite property of interpretations comes
from the constructs of the language, while the witnessed or
non-witnessed property comes from the continuity of the fuzzy
operators \cite{Hajeck05}.

Consider for example a node $x$ which contains some triple of the
form $\lan \leq 1R,\geq,1\ran$. If a successor of $x$, say $y$, is
blocked by some ancestor of $x$, say $z$, the dynamic blocking
techniques would create a cycle leading from $x$ back to $z$. But
this extra edge $\lan x,z\ran$ might violate the number
restriction on $x$. To overcome this problem the construction of
the tableau from the completion-forest is performed by repeatedly
copying the sub-tree underneath the node that causes blocking, $z$
in our case. Thus, we are able to obtain an infinite out of the
constructed finite forest. Furthermore, in order for copied nodes
to be satisfiable in their new locations an extra condition,
compared to dynamic blocking has to be employed. The new blocking
technique is called \emph{pair-wise} blocking \cite{Horrocks99};
i.e., blocking occurs when two nodes belong to the same set of
concepts, their predecessors also belong to the same set of
concepts and the edges that connect them are also equal. That way
unravelling is guaranteed.

\begin{definition}[\fkdshin Completion Forest]\label{SHINforest}
First we extend the definition of $R$-successors, predecessors and
neighbours. If nodes $x$ and $y$ are connected by an edge $\lan
x,y \ran$ with $\tup{P,\rhd,n}\in\cL(\tup{x,y})$, and $P\sss R$,
then $y$ is called an $R_{\rhd,n}$-\emph{successor} of $x$ and $x$
is called an $R_{\rhd,n}$-\emph{predecessor} of $y$. If $y$ is an
$R_{\rhd,n}$-successor or an $\Inv(R)_{\rhd,
n}$-\emph{predecessor} of $x$, then $y$ is called an $R_{\rhd,
n}$-neighbour of $x$.

For a role $R$, a node $x$ in \Forest, an inequality $\bow$ and a
membership degree $n\in[0,1]$ we define:
$R^{\Forest}_C(x,\bow,n)=\{y\mid$ $y$ is an $R_{\rhd',
n'}$-neighbour of $x$, and $\lan x,y\ran$ conjugates with $\lan
R,\bow,n\ran\}$. Intuitively, this set contains all R-neighbours
of $x$ that conjugate with a given triple.

A node $x$ is \emph{blocked} iff it is not a root node and it is
either directly or indirectly blocked. A node $x$ is
\emph{directly blocked} iff none of its ancestors is blocked, and
it has ancestors $x'$, $y$ and $y'$ such that:

\begin{enumerate}
    \item $y$ is not a root node,
    \item $x$ is a successor of $x'$ and $y$ a successor of $y'$,
    \item $\cL(x)=\cL(y)$ and $\cL(x')=\cL(y')$ and,
    \item $\cL(\lan x',x\ran)=\cL(\lan y',y\ran)$.
\end{enumerate}
\noindent In this case we say that $y$ blocks $x$. A node $y$ is
indirectly blocked iff one of its ancestors is blocked, or it is a
successor of a node $x$ and $\cL(\lan x,y\ran)=\emptyset$.

For a node $x$, $\cL(x)$ is said to contain a clash if it contains
an \fkdsi clash, or if it contains,

\begin{itemize}
    \item some triple $\lan \leq
    pR,\rhd,n\ran$ and $x$ has $p+1$ $R_{\rhd_i,n_i}$-neighbours
    $y_0,\ldots,y_p$, $\lan x,y_i\ran$ conjugates with
    $\lan R,\rhd^-,1-n\ran$ and $y_i\neq y_j$, $n_i,n\in[0,1]$,
    for all $0\leq i<j\leq p$, or

    \item some triple $\lan \geq
    pR,\lhd,n\ran$ and $x$ has $p$ $R_{\rhd_i,n_i}$-neighbours
    $y_0,\ldots,y_{p-1}$, $\lan x,y_i\ran$ conjugates with $\lan
    R,\lhd,n\ran$ and $y_i\neq y_j$, $n_i,n\in[0,1]$, for all
    $0\leq i<j\leq p-1$.

\end{itemize}
%%George 10/02/06
%Moreover, for an edge $\lan x,y\ran$, $\cL(\lan x,y\ran)$ is said
%to contain a clash iff there exist two conjugated triples in
%$\cL(\lan x,y\ran)$, or if $\cL(\lan x,y\ran)\cup
%\{\lan\Inv(R),\rhd,n\ran\mid \lan R,\rhd,n\ran\in\cL(\lan
%y,x\ran)\}$, where $x,y$ are root nodes, contains two conjugated
%triples.
\end{definition}

\begin{table*}[th]
\begin{center}
\footnotesize \hspace*{-25pt}
\begin{tabular}{crl}
\hline
\hspace*{15pt}Rule\hspace{15pt} & \hspace*{25pt}  & \hspace*{90pt}Description      \\
\hline
$(\forall'_{+})$ & if 1.& $\lan \forall S.C,\rhd,n\ran\in\cL(x)$, $x$ is not indirectly blocked, and\\
                      &    2.& there is some $R$, with $\Tr(R)$, and $R\sss S$,\\
                      &    3.& $x$ has a $R_{\rhd',n'}$-neighbour $y$ with, $\lan \forall R.C,\rhd,n\ran\not\in\cL(y)$, and\\
                      &    4.& $\lan x,y\ran$ conjugates with $\lan R,\rhd^-,1-n\ran$\\
                      & then & $\cL(y)\rightarrow\cL(y)\cup \{\lan \forall R.C,\rhd,n\ran\}$,\vspace{0.1cm}\\
%\hline
$(\exists'_{+})$ & if 1.& $\lan \exists S.C,\lhd,n\ran\in\cL(x)$, $x$ is not indirectly blocked and\\
                      &    2.& there is some $R$, with $\Tr(R)$, and $R\sss S$,\\
                      &    3.& $x$ has a $R_{\rhd,n'}$-neighbour $y$ with, $\lan \exists R.C,\lhd,n\ran\not\in\cL(y)$, and\\
                      &    4.& $\lan x,y\ran$ conjugates with $\lan R,\lhd,n\ran$\\
                      & then & $\cL(y)\rightarrow\cL(y)\cup \{\lan \exists R.C,\lhd,n\ran\}$,\vspace{0.1cm}\\

$(\geq_{\rhd})$ & if 1.& $\lan \geq pR,\rhd,n\ran\in\cL(x)$, $x$ is not blocked,\\
                      &    2.& there are no $p$ $R_{\rhd,n}$-neighbours $y_1,\ldots,y_p$ of $x$ with $y_i\neq y_j$ for $1\leq i<j \leq p$\\
%                      &    3.& with $y_i\neq y_j$ for $1\leq i<j \leq p$\\
                      & then & create $p$ new nodes $y_1,\ldots,y_p$, with $\cL(\lan x,y_i\ran)=\{\lan R,\rhd,n\ran\}$ and $y_i\neq y_j$ for $1\leq i<j \leq p$\vspace{0.1cm}\\
%\hline
$(\leq_{\lhd})$ & if 1.& $\lan \leq pR,\lhd,n\ran\in\cL(x)$, $x$ is not blocked,\\
                       & then & apply $(\geq_{\rhd})$-rule for the triple $\lan \geq (p+1)R,\lhd^-,1-n\ran$\vspace{0.1cm}\\
%\hline
$(\leq_{\rhd})$ & if 1.& $\lan \leq pR,\rhd,n\ran\in\cL(x)$, $x$ is not indirectly blocked,\\
                      &    2.& $\sharp R^{\Forest}_C(x,\rhd^-,1-n)>p$, there are two of them $y$, $z$, with no $y\not\doteq z$ and\\
                      &    3.& $y$ is neither a root node nor an ancestor of $z$\\
                      & then & 1. $\cL(z)\rightarrow\cL(z)\cup \cL(y)$ and\\
                      &      & 2. if $z$ is an ancestor of $x$ \\
                      &      &  \begin{tabular}{lrcl} then & $\cL(\lan z,x\ran)$ & $\longrightarrow$ & $\cL(\lan z,x \ran)\cup \Inv(\cL(\lan x,y \ran))$\\
                                    else & $\cL(\lan x,z\ran)$ & $\longrightarrow$ & $\cL(\lan x,z \ran)\cup \cL(\lan x,y \ran)$\\
                                \end{tabular}\\
                      &      & 3. $\cL(\lan x,y\ran)\longrightarrow \emptyset$ and set $u\not\doteq z$ for all $u$ with $u\not\doteq y$\vspace{0.1cm}\\
%\hline
$(\geq_{\lhd})$ & if 1.& $\lan \geq pR,\lhd,n\ran\in\cL(x)$, $x$ is not indirectly blocked,\\
                       & then & apply $(\leq_{\rhd})$-rule for the triple $\lan \leq (p-1)R,\lhd^-,1-n\ran$\vspace{0.1cm}\\
%\hline
$(\leq_{r_{\rhd}})$ & if 1.& $\lan \leq pR,\rhd,n\ran\in\cL(x)$,\\
                      &  2.& $\sharp R^{\Forest}_C(x,\rhd^-,1-n)>p$, there are two of them %$R$-neighbours
                      $y$, $z$, both root nodes, with no $y\not\doteq z$ and\\
                      & then & 1. $\cL(z)\rightarrow\cL(z)\cup \cL(y)$ and\\
                      &      & 2. For all edges $\lan y,w \ran$:\\
                      &      & ~i. if the edge $\lan z,w\ran$ does not exist, create it with $\cL(\lan z,w\ran)=\emptyset$\\
                      &      & ~ii. $\cL(\lan z,w\ran)\longrightarrow \cL(\lan z,w\ran)\cup \cL(\lan y,w\ran)$\\
                      &      & 3. For all edges $\lan w,y \ran$:\\
                      &      & ~i. if the edge $\lan w,z\ran$ does not exist, create it with $\cL(\lan w,z\ran)=\emptyset$\\
                      &      & ~ii. $\cL(\lan w,z\ran)\longrightarrow \cL(\lan w,z\ran)\cup \cL(\lan w,y\ran)$\\
                      &      & 4. Set $\cL(y)=\emptyset$ and remove all edges to/from $y$\\
                      &      & 5. Set $u\not\doteq z$ for all $u$ with $u\not\doteq y$ and set $y\doteq z$\vspace{0.1cm}\\
%\hline
$(\geq_{r_{\lhd}})$ & if 1.& $\lan \geq pR,\lhd,n\ran\in\cL(x)$,\\
                      & then & apply $(\leq_{r_{\rhd}})$-rule for the triple $\lan \leq (p-1)R,\lhd^-,1-n\ran$\vspace{0.0cm}\\
\hline
\end{tabular}
\end{center}
\caption{Additional tableaux rules for \fkdshin}
\label{table:fkdshin-rules}
\end{table*}

\begin{definition}[\fkdshin Tableaux Algorithm]
The initialisation of a forest (\Forest) for an \fkdshin ABox \A
is similar to the initialisation of a forest for an \fkdsi ABox
\A, with the difference, that equalities and inequalities need to
be considered. More precisely, we also initialise the relation
$\ndoteq$ as $x_{a_i}\ndoteq x_{a_j}$ if $a_i\ndoteq a_j\in\A$ and
the relation $\doteq$ to be empty. The latter is used to keep
track the nodes that are merged due to the application of a rule
for number restrictions. Finally, the algorithm expands \R by
adding axioms $\Inv(R)\sqsubseteq \Inv(S)$ for each $R\sqsubseteq
S\in\R$. \Forest is then expanded by repeatedly applying the
completion rules from Tables \ref{tableau} and
\ref{table:fkdshin-rules}. Note that in Table
\ref{table:fkdshin-rules} we abuse the syntax and use the notation
$\Inv(\cL(\lan x,y\ran))$ to indicate the set of triples obtained
from $\cL(\lan x,y\ran)$ by applying function $\Inv$ to the role
$R$ of each triple $\lan R,\bow,n\ran\in \cL(\lan x,y\ran)$.
\end{definition}

\begin{example}
Now, let us see some examples of the new expansion rules.

\begin{itemize}
    \item $(\forall_+)$: Let $\lan \forall
    S.C,>,0.6\ran\in\cL(x)$, $\lan
    \Inv(P),\geq,0.7\ran\in\cL(\lan y,x\ran)$ with $\Tr(R)$
    and $P\sqsubseteq R\sqsubseteq S$. Then, there is role
    $R$, with $R\sss S$, and $y$ is an
    $R_{\geq,0.7}$-neighbour of $x$, since $y$ is an
    $\Inv(R)_{\geq,0.7}$-predecessor of $x$,
    $\lan x,y\ran$ conjugates with $\lan
    \Inv(R),<,0.4\ran$, and $\lan \forall
    R.C,>,0.6\ran\not\in\cL(y)$. Hence, $\lan \forall
    R.C,>,0.6\ran$ should be added in $\cL(y)$.\vspace{0.0cm}

    \item $(\leq_\geq)$: Let $\lan \leq 2S,\geq,0.7\ran\in\cL(x)$,
    $\lan S,>,0.7\ran\in\cL(\lan x,y_1\ran)$, $\lan
    S,>,0.8\ran\in\cL(\lan x,y_2\ran)$ and $\lan
    P,\geq,0.4\ran\in\cL(\lan x,y_3\ran)$ with $P\sss S$. Hence,
    $x$ has 3 $S_{\rhd' n'}$-neighbours all conjugated with $\lan
    S,\geq^-,1-0.7\ran\equiv\lan S,\leq,0.3\ran$ and none an
    ancestor of $x$. Hence we have to non-deterministically merge
    two of them. If we replace the triple $\lan
    S,>,0.7\ran\in\cL(\lan x,y_1\ran)$ with $\lan
    S,>,0.2\ran$ the rule is no more applicable. That is because
    although $y_1$ is an $S_{\rhd' n'}$-neighbour of $x$, $\lan
    x,y_1\ran$ does not conjugate anymore with $\lan S,\leq,0.3\ran$.
    Intuitively, this means that the connection between $x$ and
    $y_1$ is too weak and thus does not contradict the at-most
    restriction on $x$.
\end{itemize}
\diae
\end{example}

As it is obvious the algorithm can be used in order to perform
reasoning for the weaker language \fkd-\shif (\fkd-\shi plus
\emph{functional number restrictions} \citeR{Horrocks99}. \shif is
obtained from \shin by allowing only cardinalities $0$ and $1$ in
at-most and at-least restriction. It is worth noting that, without
counting datatypes, \shif is the logical underpinning of the OWL
Lite ontology language \cite{Horrocks03c}.

%On the other hand, \citeA{Stoilos06ate} note that when the
%classical reasoning algorithm is applied to concepts of the form
%$\neg C\sqcup D$ it actually performs \emph{case analysis} on all
%concepts and individuals that appear in the ABox. More precisely,
%for all $a$

\subsection{Decidability of \fkdshin}\label{SHINdecidability}
The proof of termination, soundness and completeness of \fkdshin
is slightly more involved than that of \fkdsi. This is mainly due
to the requirement to apply the unravelling process on a
constructed finite completion forest.

\begin{lemma}[Termination]\label{lem:SHINtermination}
Let \A be an \fkdshin ABox \A and \R an RBox. The tableaux
algorithm terminates when started for \A and \R.
\end{lemma}

\begin{Proof}
Let $m=|sub(\A)|$, $k=|\bR_\A|$, $p_{\max}=\max\{p\mid \geq pR\in
sub(\A)\}$ and $l$ be the number of different membership degrees
appearing in \A. The termination of our algorithm is a consequence
of the same properties that ensure termination in the case of the
crisp \shin language \cite{Horrocks00}. In brief we have the
following observations. Firstly, the only rules that remove nodes
or concepts from the node labels are the rules $\leq_{\rhd}$,
$\geq_{\lhd}$, $\leq_{r_\rhd}$ and $\geq_{r_\lhd}$, which either
expand them or set them to $\varnothing$, which means that nodes
will be blocked and will remain blocked forever. Secondly, the
expansion rules ($\exists_{\rhd}$, $\geq_{\rhd}$ and the dual ones
for negative inequalities) can only be applied once for each node
for the same reasons as in the \shin case \cite{Horrocks00}. Since
$sub(\A)$ contains at most $m$ concepts $\exists R.C$ and $\forall
R.C$, the out-degree of the tree is bounded by $2lmp_{\max}$.
Finally, there is a finite number of possible labellings for a
pair of nodes and an edge, since concepts are taken from $sub(\A)$
and the number of membership degrees is finite. Thus, there are at
most $2^{8mlk}$ possible labellings for a pair of nodes and an
edge. Hence, if a path $p$ is of length at least $2^{8mlk}$, the
pair-wise blocking condition implies that there are 2 nodes $x,y$
on $p$ such that $y$ directly blocks $x$.\qed
\end{Proof}

\begin{lemma}\label{SHINsoundness}
(\textbf{Soundness}) If the expansion rules can be applied to an
\fkdshin ABox \A and RBox \R, such that they yield a complete and
clash-free completion forest, then \A has a fuzzy tableau w.r.t.
\R.
\end{lemma}
\begin{Proof}
Let \Forest be a complete and clash-free completion forest
constructed by the tableaux algorithm for \A. Since the \shin
language does not have the finite model property \cite{Horrocks99}
we have to unravel a possibly blocked tree in order to obtain an
infinite tableau. The constructions of such fuzzy tableau works as
follows. An individual in \bS corresponds to a \emph{path} in
\Forest. Moving down to blocked nodes and up to blocking ones we
can define infinite such paths. More precisely, a \emph{path} is a
sequence of pairs of nodes of \Forest of the form
$p=[\frac{x_0}{x'_0},\ldots,\frac{x_n}{x'_n}]$. For such a path we
define $\Tail(p):=x_n$ and $\Tail'(p):=x'_0$. With
$[p\mid\frac{x_{n+1}}{x'_{n+1}}]$, we denote the path
$[\frac{x_0}{x'_0},\ldots,\frac{x_n}{x'_n},\frac{x_{n+1}}{x'_{n+1}}]$.
The set $\Paths(\Forest)$ is defined inductively as follows:

\begin{itemize}
    \item For root nodes $x_{a_i}$ of \Forest,
    $[\frac{x_{a_i}}{x_{a_i}}]\in\Paths(\Forest)$, and

    \item For a path $p\in\Paths(\Forest)$ and a node $z$ in
    \Forest:
    \begin{itemize}
        \item if $z$ is a successor of $\Tail(p)$ and $z$ is
        neither blocked not a root node, then
        $[p\mid\frac{z}{z}]\in\Paths(\Forest)$, or
        \item if for some node $y$ in \Forest, $y$ is a successor
        of $\Tail(p)$ and $z$ blocks $y$, then
        $[p\mid\frac{z}{y}]\in\Paths(\Forest)$
    \end{itemize}
\end{itemize}

Please node that since root nodes are never blocked, nor are they
blocking other nodes the only place where they occur in a path is
in the first place. Moreover, if $p\in\Paths(\Forest)$, then
$\Tail(p)$ is not blocked; $\Tail(p)=\Tail'(p)$ iff $\Tail'(p)$ is
not blocked and at last $\cL(\Tail(p))=\cL(\Tail'(p))$.

Membership degrees are defined exactly as in the case of \fkdsi.
Then, a fuzzy tableau can be defined as in the case of \fkdsi with
the following differences:

\begin{center}{\small
\begin{tabular}{rcl}
\bS                                             & = &   $\Paths(\Forest)$,\vspace{0.1cm}\\
%$\cL(p,\bot)$                                   & = &   $0$, for all $p\in\bS$,\vspace{0.0cm}\\
%$\cL(p,\top)$                                   & = &   $1$, for all $p\in\bS$,\vspace{0.0cm}\\
%$\cL(p,C)$                                      & = &   $glb[\lan C,\bow,n_i\ran]$, for $\lan C,\bow,n_i\ran\in\cL(\Tail(p))$,\vspace{0.0cm}\\
%$\cL(p,\neg A)$                                 & = &   $1-\cL(p,A)$, $\lan \neg A,\bow,n\ran\in\cL(\Tail(p))$,\vspace{0.0cm}\\
$\cE(R,\lan p,[p|\frac{x}{x'}]\ran)$            & = &   $glb[\lan R,\bow,n\ran], \lan R,\bow,n\ran\in\cL(\lan \Tail(p),x'\ran)$\vspace{0.1cm}\\
$\cE(R,\lan [q|\frac{x}{x'}],q\ran)$            & = &   $glb[\lan \Inv(R),\bow,n\ran], \lan \Inv(R),\bow,n\ran\in\cL(\lan \Tail(q),x'\ran)$\vspace{0.1cm}\\
$\cE(R,\lan [\frac{x}{x}],[\frac{y}{y}]\ran)$   & = &   $glb[\lan R^*,\bow,n\ran]$, $x,y$ root nodes and $y$ $R$-neighbour of $x$,\vspace{0.1cm}\\

$\cV(a_i)$ & = & $\left\{\begin{tabular}{l} $[\frac{x_{a_i}}{x_{a_i}}]$ if $x_{a_i}$ is a root node in \Forest with $\cL(x_{a_i})\neq\emptyset$\vspace{0.1cm}\\
                                             $[\frac{x_{a_j}}{x_{a_j}}]$ if $\cL(x_{a_j})=\emptyset$ and $x_{a_j}$ is a root node, \\
                                             \hspace{24pt}with $\cL(x_{a_j})\neq\emptyset$ and $x_{a_i}\doteq x_{a_j}$\\
                         \end{tabular}\right.$\\

\end{tabular}}
\end{center}
It can be shown that $T$ is a fuzzy tableau for \A \wrt \R:

\begin{enumerate}

    \item Properties 1-6 %of Definition \ref{SItableau} %are
%    satisfied because \Forest is clash-free and due to the
%    construction of T. For example, let $\cL(p,\neg A)=n_1\geq n$.
%    The definition of $T$ implies that $1-n \geq
%    1-n_1=\cL(p,A)$.\vspace{0.1cm}
    and Property 13 of Definition \ref{SItableau} are
    satisfied due to the same reasons as in the proof of lemma
    \ref{SIsoundness}\vspace{0.1cm}.
    %because none of the $\sqcup_{\rhd}$,
%    $\sqcap_{\rhd}$, $\sqcup_{\lhd}$ or $\sqcap_{\lhd}$
%    rules apply to any node in \Forest, and $\Tail(p)$ is not
%    blocked. For example, let $\cL(p,C\sqcap D)=n_1\geq n$. The
%    definition of $T$ implies that, either $\lan C\sqcap
%    D,\geq,n_1\ran\in\cL(\Tail(p))$ or $\lan C\sqcap
%    D,>,n'\ran\in\cL(\Tail(p))$, with $n_1=n'+\epsilon$.
%    Completeness of \Forest implies that either $\lan
%    C,\geq,n_1\ran\in\cL(\Tail(p))$ and $\lan
%    D,\geq,n_1\ran\in\cL(\Tail(p))$ or $\lan
%    C,>,n'\ran\in\cL(\Tail(p))$ and $\lan
%    D,>,n'\ran\in\cL(\Tail(p))$. Hence, $\cL(s,C)\geq
%    \cL(s,C\sqcap D)\geq n$, $\cL(s,D)\geq \cL(s,C\sqcap D)\geq
%    n$. Property 5 follows for similar reasons.

    \item For property 7, let $p,q\in\bS$ with $\cL(p,\forall
    R.C)=n_1\geq n$ and $\cE(R,\lan p,q\ran)\nleq 1-n$, i.e.
    $\cE(R,\lan p,q\ran)>1-n$. The definition of $T$ implies
    that either $\lan \forall R.C,\geq,n_1\ran\in\cL(\Tail(p))$ or
    $\lan \forall R.C,>,n'\ran\in\cL(\Tail(p))$ with
    $n_1=n'+\epsilon$. If $q=[p|\frac{x}{x'}]$, then $x'$ is an
    $R$-successor of $\Tail(p)$ and, since $glb$ does not create
    unnecessary conjugations we have that $\lan
    R,\rhd,r\ran\in\cL(\lan \Tail(p),x'\ran)$ conjugates with
    $\lan R,\leq,1-n\ran$. Hence, due to completeness of \Forest
    we have either $\lan C,\geq,n_1\ran\in\cL(x')$ or $\lan
    C,>,n'\ran\in\cL(x')$. From the definition of
    $\Paths(\Forest)$ we have that $\cL(x')=\cL(x)=\cL(q)$. If
    $p=[q|\frac{x}{x'}]$, then $x'$ is an $\Inv(R)$-successor of
    $\Tail(q)$ and again, the definition of $glb$ implies that
    $\lan \Inv(R),\rhd,r\ran\in\cL(\lan \Tail(q),x'\ran)$
    conjugates with $\lan\Inv(R),\leq,1-n\ran$. Thus, due to
    completeness of \Forest, either $\lan
    C,\geq,n_1\ran\in\cL(\Tail(q))=\cL(q)$ or $\lan
    C,>,n'\ran\in\cL(\Tail(q))=\cL(q)$. If $p=[\frac{x}{x}]$ and
    $q=[\frac{y}{y}]$ for two root nodes $x,y$ then $y$ is an
    $R$-neighbour of $x$, and since the $\forall_{\rhd}$-rule does
    not apply we have that wither $\lan
    C,\geq,n_1\ran\in\cL(y)=\cL(q)$ or $\lan
    C,>,n'\ran\in\cL(y)=\cL(q)$. Similar proof holds for
    $\cL(p,\forall R.C)>n$ and for property 8, of definition
    \ref{SItableau} and for the modified properties 11' and 12' of
    definition \ref{SHINtableau}.\vspace{0.1cm}

    \item For property 9 of Definition \ref{SItableau}, assume
    that $\cL(p,\exists R.C)=n_1\geq n$ and let $\Tail(p)=x$. The
    definition of $T$ implies that either $\lan \exists
    R.C,\geq,n_1\ran\in\cL(x)$ or $\lan \exists
    R.C,>,n'\ran\in\cL(x)$, with $n_1=n'+\epsilon$. We have to
    show that there is some $q\in\bS$ such that $\cE(R,\lan
    p,q\ran)\geq n_1\geq n$ and $\cL(q,C)\geq n_1\geq n$. Since
    the $\exists_{\rhd}$-rule is not
    applicable there is some $y$ in \Forest with either $\lan
    C,\geq,n_1\ran\in\cL(y)$ or $\lan
    C,>,n'\ran\in\cL(y)$. Now there are two possibilities:
        \begin{enumerate}
            \item If $y$ is a successor of $x$, then $y$ can
            either be a root node or not. In case $y$ is a root
            node so is $x$, since it is a predecessor of $y$, so
            $p=[\frac{x}{x}]$ and $q=[\frac{y}{y}]\in\bS$. In case
            $y$ is not a root node if $y$ is not blocked, then
            $q=[p|\frac{y}{y}]\in\bS$; if $y$ is blocked by some
            $z$ then, $q=[p|\frac{z}{y}]\in\bS$.

            \item $x$ is an $\Inv(R)$-successor of $y$. Since
            $x$ is a successor of $y$ we distinguish the cases of $x$
            being a root node or not.
            If $x$ is a root then so is $y$, hence
            $q=[\frac{y}{y}]\in\bS$. If $x$
            is not a root node then either $p=[q|\frac{x}{x'}]$,
            with $\Tail(q)=y$, or $p=[q|\frac{x}{x'}]$, with
            $\Tail(q)=u\neq y$, $x$ blocks $x'$ and $u$ is a
            predecessor of $x'$. By the definition
            of pair-wise blocking we have that $\cL(y)=\cL(u)$ and
            $\cL(\lan y,x\ran)=\cL(\lan u,x'\ran)$.
        \end{enumerate}
    In any of these cases, $\cE(R,\lan
    p,q\ran)\geq n_1\geq n$, $\cL(q,C)\geq n_1\geq n$.
    Similar proof applies for $\cL(p,\exists
    R.C)> n$ and for property 10.\vspace{0.1cm}

    \item Property 16 in definition \ref{SHINtableau} is
    satisfied due to the definition of $R$-successor that
    takes into account the role hierarchy.\vspace{0.1cm}

    \item For Property 17 assume that $\cL(p,\geq
    mR)=n_1\geq n$. The definition of $T$ implies
    that either $\lan \geq mR,\geq,n_1\ran\in\cL(x)$
    or $\lan \geq mR,>,n'\ran\in\cL(x)$, with $n_1=n'+\epsilon$.
    This means that there are $m$
    individuals $y_1,\ldots,y_m$ in \Forest such that each $y_i$
    is an $R_{\geq,n'}$- or $R_{>,n'}$-neighbour of
    $x$. We have to show that for each of these $y_i$s,
    there is a path $q_i$, such that $\cE(R,\lan
    p,q_i\ran)\geq n_1$, and $q_i\neq q_j$ for all
    $1\leq i<j\leq m$. The proof is similar with the one given by
    \citeA{Horrocks00}. It is based on the fact that in case
    where some $z$ blocks several $y_i$s, then the construction
    of the paths distinguishes between these $y_i$s be seting
    $q_i=[p|\frac{z}{y_i}]$, thus ensuring the existence of
    different paths in $T$. Thus, for each $y_i$ there is a
    different path $q_i$ in \bS with $\cE(R,\lan p,q_i\ran)\geq
    n_1\geq n$, or $\cE(R,\lan p,q_i\ran)\geq n'+\epsilon\geq n$
    and $\sharp R^T(p,\geq,n)\geq m$. Similarly for $\cL(p,\geq
    mR)>n$ and for property 18.\vspace{0.1cm}

    \item For Property 19 in definition \ref{SHINtableau} suppose
    that there exists $p\in\bS$ with $\cL(p,\leq
    mR)=n_1\geq n$ and $\sharp R^T_\neg(p,\leq,1-n)>m$.
    We have to show that this implies $\sharp
    R^{\Forest}_C(\Tail(p),\leq,1-n)>m$, in the
    completion-forest, thus contradicting either clash-freeness or
    completeness of \Forest. More precisely, one has to show that the
    construction does not create more conjugated paths for $T$
    than those that exist in \Forest. This can only be the case
    if for some node $y$ the construction creates two distinct
    paths of the form $q_i=[p|\frac{y_i}{y}]$. As shown by
    \citeA{Horrocks00}, the proof relies on the fact that
    the function $\Tail'$ is injective on the paths of $T$, i.e.
    for $q_1$ and $q_2$, $\Tail'(q_1)=y=\Tail'(q_2)$ implies that
    $q_1=q_2$. Hence, such paths cannot be distinct. Similar
    observations hold for $\cL(p,\leq mS)>n$ and for property 20.
    \vspace{0.1cm}

    \item Properties 14 and 15 of Definition \ref{SItableau} are
    satisfied cause of the initialization of the completion-forest
    and the fact that the algorithm never blocks root nodes.
    Furthermore, for each root node $x_{a_i}$ whose label and edges
    are removed by the $\leq_{r_{\rhd}}$-rule, there is another
    root node $x_{a_j}$ with $x_{a_i}=x_{a_j}$ and $\{\lan
    C,\rhd,n\ran|(a_i:C)\rhd
    n\in\A\}\subseteq\cL(x_{a_j})$.\vspace{0.1cm}

    \item Property 21 of Definition \ref{SHINtableau} is satisfied
    because the $\leq_{r_{\rhd}}$-rule does not identify two root
    nodes $x_{a_i}$, $x_{a_j}$ when $x_{a_i}\neq x_{a_j}$ holds.
\end{enumerate}
\qed
\end{Proof}

\begin{lemma}[Completeness]\label{SHINcompleteness}
Let \A be an \fkdshin fuzzy ABox and \R an RBox. If \A has a fuzzy
tableau w.r.t. \R, then the expansion rules can be applied to \A
and \R in such a way that the tableaux algorithm yields a complete
and clash-free completion-forest.
\end{lemma}
\begin{Proof}
The proof is quite similar with the proof of lemma
\ref{SIcompleteness}. In the new algorithm we have some new
non-deterministic rules, but again the existence of fuzzy tableau
for \A \wrt \R can help us steer the application of those
non-deterministic rules. In the following table we show the
modified rule $\leq'_{\rhd}$. The rest of the non-deterministic
rules can be guided by modifying them in a similar way.

\begin{table}[th]
\begin{center}\small
\begin{tabular}{crl}
$(\leq'_{\rhd})$ & if 1.& $\lan \leq pR,\rhd,n\ran\in\cL(x)$, $x$ is not indirectly blocked,\\
                      &      & $\sharp R^{\Forest}_C(x,\rhd^-,1-n)>p$, there are two of them %$R$-neighbours
                      $y$, $z$, with no $y\not\doteq z$ and\\
%                      &    2.& there are $p+1$ $R$-neighbours $y_1,\ldots,y_{p+1}$ connected to $x$ with a triple $\lan P^*,\rhd', n_i\ran$, $P\sss R$,\\
%                      &    3.& which is conjugated with $\lan P^*,\rhd^-,1-n\ran$, and there are two of them $y$, $z$, with no $y\not\doteq z$ and\\
                      &    3.& $y$ is neither a root node nor an ancestor of $z$ and $\pi(y)=\pi(z)$\\
                      & then & 1. $\cL(z)\rightarrow\cL(z)\cup \cL(y)$ and\\
                      &      & 2. if $z$ is an ancestor of $x$ \\
                      &      & \begin{tabular}{lrcl}
                                then & $\cL(\lan z,x\ran)$ & $\longrightarrow$ & $\cL(\lan z,x \ran)\cup \Inv(\cL(\lan x,y \ran))$\\
                                else & $\cL(\lan x,z\ran)$ & $\longrightarrow$ & $\cL(\lan x,z \ran)\cup \cL(\lan x,y \ran)$\\
                               \end{tabular}\\
                      &      & 3. $\cL(\lan x,y\ran)\longrightarrow \emptyset$ and set $u\not\doteq z$ for all $u$ with $u\not\doteq y$\\
%                      &      & 4. Set $u\not\doteq z$ for all $u$ with $u\not\doteq y$\vspace{0.0cm}\\
\end{tabular}
\end{center}
\caption{The $\leq'_{\rhd}$-rule}\label{Nmodified}
\end{table}
\qed
\end{Proof}

\begin{theorem}\label{SHINdecision}
The tableaux algorithm is a decision procedure for the consistency
problem of \fkdshin ABoxes and the satisfiability and subsumption
of \fkdshin-concepts with respect to \emph{simple} terminologies.
\end{theorem}
%Theorem \ref{SHINdecision} is an immediate consequence of lemmas
%\ref{SHINtableau}, \ref{SHINsoundness} and \ref{SHINcompleteness}.
%Moreover, as we discussed in section \ref{sec:f-SI}, subsumption
%can be reduced to consistency checking for ABoxes.

We will conclude this section by investigating the complexity of
the proposed algorithm.
\begin{lemma}\label{lem:sublength}
For an \fkdshin ABox \A and a role hierarchy \R,
$sub(\A,\R)=\Oonly(|\A|\times |\R|)$.
\end{lemma}
\begin{Proof}
The proof is quite similar with the one presented by
\citeA{Tobies01}. Since $sub(\A,\R)$ contains all concepts $C$
such that $(a:C)\bow n\in\A$ and is closed under sub-concepts of
$C$, it contains $\Oonly(|\A|)$ concepts. Additionally, we have to
add a concept $\forall R.C$ or $\exists R.C$ to $sub(\A,\R)$ if
$\forall S.R\in sub(\A,\R)$ or $\exists S.R\in sub(\A,\R)$ and
$R\sss S$ and then close $sub(\A,\R)$ again under sub-concepts and
$\sim$. This may yield at most two concept for every concept in
$sub(\A,\R)$ and role in \R. Thus, $sub(\A,\R)=\Oonly(2|\A|\times
|\R|)$.\qed
\end{Proof}

\begin{lemma}
The \fkdshin-algorithm runs in 2-\Nexptime.
\end{lemma}
\begin{Proof}
Let \A be a \fkdshin ABox and \R an RBox. Let $m=sub(\A)$,
$k=|\bR_\A|$, $p_{\max}$ the maximum number $p$ that occurs in a
number restriction and $l$ the number of different membership
degrees appearing in \A. Following \citeA{Tobies01} we set
$n=|\A|+|\R|$, then due to lemma \ref{lem:sublength} it holds that
$m=\Oonly(2|\A|\cdot|\R|)=\Oonly(n^2)$, $k=\Oonly(|\A|+|\R|)$,
$p_{\max}=\Oonly(2^{|\A|})=\Oonly(2^n)$ and
$l=\Oonly(|\A|)=\Oonly(n)$. In the proof of lemma
\ref{lem:SHINtermination} we have shown that paths in a
completion-forest for \A become no longer than $2^{8mlk}$ and that
the out-degree is bounded by $2lmp_{\max}$. Hence, the \fkdshin
algorithm will construct a forest with no more than
\[(2lmp_{\max})^{2^{8mlk}}=\Oonly((2n\cdot n^2\cdot 2^n)^{2^{8n^2\cdot n\cdot n}})=\Oonly(2^{n\cdot 2^{8n^4}})=\Oonly(2^{2^{n^5}})\]
nodes.\qed
\end{Proof}
Hence, the \fkdshin algorithm is of the same theoretical
complexity as the \shin algorithm \cite{Tobies01}.

Concluding our presentation on the issue of reasoning with
expressive fuzzy DLs we comment on how to handle GCIs in the
\fkdsi and \fkdshin languages. As it is noted by
\citeA{Horrocks99}, \shin is expressive enough to
\emph{internalize} GCIs into a single concept, hence reducing
reasoning with GCIs to concept satisfiability. The idea behind
internalization is that the semantic restrictions imposed by an
axioms of the form $C\sqsubseteq D$ can be encoded within a
concept of the form $\neg C\sqcup D$. As it was remarked by
\citeA{Stoilos06ate} this reduction of concept inclusions does not
hold for f$_{KD}$-DLs, since the semantics of the axiom
$C\sqsubseteq D$ are different than that of the concept $\neg
C\sqcup D$. Hence, the internalization method proposed by
\citeA{Horrocks99} for the \shin language cannot be applied in the
\fkdshin language.

\citeA{Stoilos06ate} and \citeA{Li06a} propose techniques by which
we can handle GCIs in \fkd-DLs. \citeA{Stoilos06ate} use the DL
language \fkdalc in order to present their technique, while
\citeA{Li06a} use the language \fkd-\shi. These procedures can be
applied in the cases of \fkdsi and \fkdshin, since they are
independent of the underlying DL formalism. Roughly speaking these
techniques are performed in three steps. In the first step the
ABox is \emph{normalized}, by replacing each assertion of the form
$(\indv a:C)>n$ and $(\indv a:C)<n$ by assertions $(\indv a:C)\geq
n+\ell$ and $(\indv a:C)\leq n-\ell$, respectively, where $\ell$
is a small number from $[0,1]$. Obviously, in a normalized ABox
only assertions with inequalities $\geq$ and $\leq$ are present.
In the second step the set of \emph{relative} membership degrees
is constructed: $X^\A=\{0,0.5,1\}\cup \{n,1-n\mid \phi\bow n\}$,
where obviously $\bow\in\{\geq,\leq\}$. Finally, a tableaux
expansion rule is employed to transfer the semantic restrictions
imposed by each GCI $C\sqsubseteq D\in\T$ into fuzzy assertions of
the ABox. More precisely, for each $C\sqsubseteq D\in\T$, node $x$
in \Forest and degree $n\in X^\A$, the algorithm adds either $\lan
C,\leq,n-\ell\ran$ or $\lan D,\geq,n\ran$ to $\cL(x)$. We remark
here that the rule proposed by \citeA{Li06a} is slightly
different.

As noted by \citeA{Stoilos06ate}, tableaux algorithms need to be
slightly changed in order to handle GCIs. First, due to the
normalization step, degrees are now taken from the interval
$[-\ell,1+\ell]$, thus clash definitions $\lan \bot,>,n\ran$ and
$\lan \top,<,n\ran$ are removed since no assertion with $>$ and
$<$ exist anymore and the clashes $\lan C,<,0\ran$ and $\lan
C,>,1\ran$ are replaced by $\lan C,\leq,-\ell\ran$ and $\lan
C,\geq,1+\ell\ran$, respectively. The termination of the algorithm
is not affected since again the set of membership degrees is
finite (taken from the set $X^\A$), but the practical complexity
increases dramatically since we have a non-deterministic choice
for each axiom $C\sqsubseteq D\in\T$ and degree $n\in X^\A$. The
proof of soundness is not affected much and as it was showed by
\citeA{Stoilos06ate} the $glb$ function is replaced by a simple
$\max$, due to the lack of assertions with inequalities $>$ and
$<$, while the non-deterministic rule for handling subsumptions
can also be modified to be guided, in order to provide us with a
proof for completeness.

\begin{example}
Let the knowledge base $\kb=\tup{\{\geq 1R\sqsubseteq
C\},\{(\tup{\indv a,\indv b}:R)\geq 0.6, (\indv a:C)<0.6\}}$.
Intuitively, the concept axioms states that the \emph{domain} of
the role $R$ is concept $C$.\footnote{A domain axiom can also be
stated as $\exists R.\top\sqsubseteq C$, but we use the above form
in order to show how the algorithm behaves with number
restrictions.} Obviously, the knowledge base is unsatisfiable
since the concept axiom suggests that $\forall
x^\I\in\deltai,\sup_{b^\I_1}\min^1_{i=1}(R^\I(x^\I,b^\I_i))=R^\I(x^\I,c^\I)\leq
C^\I(x^\I)$, for some arbitrary $c^\I\in\deltai$, but the ABox
assertions state that there exists $\indv a^\I\in\deltai$ and
$b^\I\in\deltai$ such that $R^\I(\indv a^\I,b^\I)\geq
0.6>C^\I(\indv a^\I)$. The above concept inclusion axioms is a
GCI, hence we have to use a technique for GCIs.

First, we apply the normalization step in the original ABox,
obtaining the normalized one: $\{(\tup{\indv a,\indv b}:R)\geq
0.6, (\indv a:C)\leq 0.6-\ell\}$. Secondly, we collect the set of
relative membership degrees:
$X^\A=\{0,0.5,1\}\cup\{0.4,0.4+\ell,0.6-\ell,0.6\}$.

Then, the algorithm initializes a completion-forest to contain the
following nodes with the respective triples:

\begin{center}
\begin{tabular}{cll}
$(1)$ & $\lan R,\geq,0.6\ran\in\cL(\lan x_{\indv a},x_{\indv b}\ran)$ &\\
$(2)$ & $\lan C,\leq,0.6-\ell\ran\in\cL(x_{\indv a})$ &\\
\end{tabular}
\end{center}

Then the algorithm expands the completion forest by using the
rules from Tables \ref{tableau} and \ref{table:fkdshin-rules} and
with the additional rule presented by \citeA{Stoilos06ate}. This
rule applied on the axiom $\geq 1R\sqsubseteq C$ adds either $\lan
\geq 1R,\leq,n-\ell\ran$ or $\lan C,\geq,n\ran$ in $\cL(x_{\indv
a})$, for each $n\in X^\A$. Hence, at some point the algorithm
chooses $0.6\in X^\A$ and adds either $\lan \geq
1R,\leq,0.6-\ell\ran$, or $\lan C,\geq,0.6\ran$. In the former
case $\lan \geq 1R,\leq,0.6-\ell\ran\in\cL(x_{\indv a})$ and
$x_{\indv a}$ has 1 $R_{\geq, 0.6}$-neighbour $x_{\indv b}$, and
$\lan x_{\indv a}, x_{\indv b}\ran$ conjugated with $\lan
R,\leq,0.6-\ell\ran$, while in the latter case $\{\lan
C,\leq,0.6-\ell\ran,\lan C,\geq,0.6\ran\}\subseteq \cL(x_{\indv
a})$, hence $\cL(x_{\indv a})$ contains a pair of conjugated
triples and thus a clash. We conclude that all possible expansions
result to a clash, thus the knowledge base is unsatisfiable.

\diae
\end{example}

\section{Related Work}\label{sec:relwork}
There have been many efforts in the past to extend description
logics with fuzzy set theory
\cite{Yen91,Tresp98,Straccia01,Holldobler02,Sanchez06,Straccia05,Hajeck05,Li06b}.
The first effort was presented by \citeA{Yen91}. In his extension,
explicit membership functions over a domain were used as well as
\emph{membership manipulators}, like ``very" or ``moreOrLess", in
order to alter membership functions and define new concepts from
already defined ones. A later approach was presented by
\citeA{Tresp98}, where membership manipulators also appear.
Regarding reasoning algorithms Yen described a structural
subsumption algorithm for a rather small DL language while Tresp
and Molitor a tableaux calculus for $\mathcal{ALC}_{FM}$ (\alc
extended with fuzzy set theory and the membership manipulator
constructor). %\citeA{Tresp98} used equalities for fuzzy
%assertions.
The application of the tableaux rules creates a set of equations
and inequations which are later solved with an optimization
method. Moreover, when determining a subsumption or entailment
relation between two concepts, with respect to a (KB), the
assertions of the KB were considered of a crisp form (i.e. $a$
belongs to $C$ to a degree of 1). After the application of the
reasoning algorithm and the solution of the equations the minimum
value of the solution set is taken as the degree that the KB
entails the crisp assertion or that a concept is subsumed by
another.

A fuzzy extension of the \alc language was also considered by
Straccia \citeyear{Straccia01,Straccia98}. Reasoning algorithms
for the problem of crisp entailment and subsumption were provided,
and were based on tableaux calculus. The algorithm was proved to
be PSPACE-complete. Moreover, complete reasoning algorithms for
fuzzy \alc were provided by \citeA{Holldobler02}, where membership
manipulators (linguistic hedges) were also used on primitive
concepts. This approach was later extended by \citeA{Holldobler05}
to allow linguistic hedges also on complex concepts. The languages
presented are called $\mathcal{ALC}_{FH}$ and
$\mathcal{ALC}_{FLH}$ (\alc plus linguistic hedges and linear
linguistic hedges, respectively). In all these approaches also the
$\min$-$\max$ norms and the Kleene-Dienes implication were used to
perform the fuzzy set theoretic operations.

%Furthermore, there are some efforts to extend the semantics of
%\shoindp \cite{Straccia05} and \alcq \cite{Sanchez04}, the latter
%including fuzzy quantifiers, but with no reasoning algorithms.
%Moreover, in \cite{Straccia05} no investigation was provided for
%the effect of adding fuzzines to transitive relations, on the
%semantics of the language.

%George 21/10/05
Approaches towards more expressive DLs, are presented by
\citeA{Sanchez04,Sanchez06}, \citeA{Straccia05},
\citeA{Straccia05d} and \citeA{Stoilos05d}. The language
considered by \citeA{Sanchez04} is \alcq (\alc plus qualified
number restrictions, \citeR{Tobies01}). The authors also include
\emph{fuzzy quantifiers} which is a novel approach to fuzzy DLs,
and the norm operations are the same with the ones used here.
Subsequently, \citeA{Sanchez06} propose a procedure to calculate
the satisfiability interval for a fuzzy concept. Due to the
presence of fuzzy quantifiers it is not clear how other inference
problems, like entailment and subsumption can be solved.
\citeA{Straccia05} considered the semantics of fuzzy \shoindp,
which is the DL counterpart of the OWL DL language. In his
approach generalized norm operations were used for the semantics,
while no reasoning algorithms were provided as well as no
investigation of the properties of value and existential
restrictions, when transitive relations and role hierarchies
participate in such concepts. Furthermore, the semantics of number
restrictions were not analyzed. The approach by Straccia was used
by \citeA{Stoilos05d}, in order to provide the abstract syntax and
semantics to concept and role descriptions and axioms of the fuzzy
OWL language. Additionally, \citeA{Stoilos05d} present a method to
translate a fuzzy OWL ontology to a fuzzy \shoin knowledge base,
thus reasoning in fuzzy OWL can be reduced to reasoning in
expressive fuzzy DLs. At last the language considered by
\citeA{Straccia05d} is \alcd (\alc plus concrete domains), where
additionally a reasoning algorithm based on an optimization
technique was presented. The norm operations used are the ones we
used in the current paper, plus the Lukasiewicz t-norm,
$t(a,b)=\max(0,a+b-1)$, t-conorm, $u(a,b)=\min(1,a+b)$ and fuzzy
implication $\J(a,b)=\min(1,1-a+b)$. An approach towards fuzzy DLs
with concrete domains has been also presented by \citeA{Liu04} for
modelling the selection of research and development projects.

In all previous approaches reasoning with respect to simple and
acyclic TBoxes was considered. \citeA{Stoilos06ate} propose a
method to perform reasoning \wrt general and/or cyclic TBoxes in
the language \fkdalc. This method applies a preprocessing step on
the ABox, called \emph{normalization} and then it extends the
classical \fkdalc algorithm \cite{Straccia01} with an additional
rule, in order to deal with general and cyclic axioms. Moreover,
\citeA{Li06a} extend the fuzzy tableau of \fkdshi proposed by
\citeA{Stoilos05c} with an additional rule to also handle with
general and cyclic TBoxes in the language \fkdshi. Interestingly,
the technique used by \citeA{Stoilos06ate} is different than that
presented by \citeA{Li06a}.

It also is worth noting the works of \citeA{Bonatti03}, where the
complexity of fuzzy DL languages is investigated. Furthermore,
\citeA{Hajeck05} investigates properties of the fuzzy \alc
language, when arbitrary continuous norm operations are used and
provides interesting results. More precisely, Hajek shows that the
problems of concept satisfiability and subsumption are decidable
for the Lukasiewicz fuzzy \alc (\falc{L}), while in product fuzzy
\alc (\falc{P}) and G\"odel fuzzy \alc (\falc{G}) only witnessed
satisfiability and subsumption are decidable. For unrestricted
models both \falc{P} and \falc{G} lack the finite model property
\cite{Hajeck05}. This is accomplished by reducing these problems
to the problem of propositional satisfiability of fuzzy
propositional logic. These results where further extended to fuzzy
DLs with truth constants, i.e. to ABox consistency, again by
\citeA{Hajek06}. Moreover, \citeA{Straccia04} present a technique
by which an f$_{KD}$-\alch knowledge base can be reduced to a
crisp \alch knowledge base. Hence, reasoning in a fuzzy KB can be
performed by using existing and optimized DL systems. Then,
\citeA{Bobillo06} extended this technique to be able to reduce a
\fkdshoin KB to a crisp \shoin KB. At last, \citeA{Li05} and
\citeA{Li06b}, also use the idea of the reduction in order to
annotate the concepts and roles of the crisp languages $\alcn$ and
$\alcq$, respectively with degrees denoted as sub-scripts in the
syntax of concepts and roles and provide reasoning for the
languages f$_{KD}$-$\alcn$ and f$_{KD}$-$\alcq$.

In all previous approaches, reasoning algorithms for rather
inexpressive fuzzy DLs, i.e. fuzzy-\alc extended with concept
modifiers or concrete domains or number restrictions or qualified
number restrictions or general TBoxes were presented. As far as we
know this is the first presentation of a reasoning algorithm for
such complex fuzzy DL languages. In order to achieve our goal we
have provided an investigation on the semantics of the extended
language when fuzzy transitive relations and role hierarchies are
considered in value and existential restrictions or of the number
restrictions constructor. The aim of such an investigation is to
discover if properties of the classical \si and \shin languages,
like the propagation of value restrictions or counting of
$R$-neighbours also apply in the fuzzy case. We have found that
apart from value restrictions also existential restrictions have
to be propagated. Additionally, we have shown that the membership
degree of these concepts in their new nodes is the same as that in
their source nodes. Moreover, we have seen that role hierarchies
can be smoothly integrated as in the classical case. Additionally,
the analysis of the semantics of number restrictions have shown
that despite how complex their semantics might appear, regarding
reasoning they can also be efficiently handled, as in the
classical case. Furthermore, we have investigated the
applicability of blocking strategies, like dynamic blocking and
pair-wise blocking, which are used in the crisp \si and \shin
language to ensure the termination of the proposed algorithms. We
have seen that the properties of the norm operations used ensure
that such blocking conditions can also be applied. Based on these
investigations, we were able to provide a tableaux reasoning
algorithm, to decide the key inference problems of very expressive
fuzzy DLs, and we have proved their soundness, completeness and
termination.

%Last but not least, we have extended a reduction procedure
%proposed in the literature \cite{Straccia04}, which transforms
%\fkdsi reasoning into the reasoning of the crisp \shi language.

\section{Conclusions and Future Work}\label{sec:conclution}
%Stoilos 13/07/06
Making applications capable of coping with vagueness and
imprecision will result in the creation of systems and
applications which will provide us with high quality results and
answers to complex user defined tasks. To this direction we have
to extend with fuzzy set theory the underlying logical formalisms
that they use in order to represent knowledge and perform
reasoning tasks. DL is a logical formalism that has gained a lot
of attention the last decade, cause of its decidability, the
powerful reasoning tools that have been implemented and the
well-defined model-theoretic semantics.
%Most real life application, that use description logics, are based
%on complex DL languages like \shoindp, which is hard to deal with
%even in the classical case.

Towards extending DLs with fuzzy set theory we have presented two
very expressive fuzzy DLs, \fkdsi and \fkdshin. We have
investigated the properties of the semantics that result by adding
fuzziness to very expressive fuzzy DLs, i.e. to fuzzy DLs that
allow for transitive and inverse roles, role hierarchies and
number restrictions and we have provided sound, complete and
terminating reasoning algorithms for both of these formalisms.
Even though handling fuzziness in such expressive languages seems
quite difficult and reasoning was not previously known, we show
that \fkdsi and \fkdshin with the $\min$, $\max$ norms are still
decidable. We have shown that the techniques used in the classical
case can also be applied in the extended frameworks, but this can
only happen after closely investigating the properties of the
languages and after proving that these techniques also work in
this new setting. In the current paper we have not addressed
\emph{nominals} (\Oonly) \cite{Horrocks05a}. Note that in the
fuzzy DL literature, there are proposals for crisp nominals
\cite{Stoilos05d} and fuzzy nominals \cite{Bobillo06}. Thus, the
nominal constructor is not yet a mature notion in fuzzy DLs and
more research is needed in order to find appropriate semantics for
them, considering also the issue from the application point of
view.

As far as future directions are concerned, these will include the
extension of the algorithm of \fkdshin, in order to provide
reasoning support for the fuzzy DL \fkdshoiq. \shoiq extends \shin
with \emph{qualified} number restrictions \cite{Tobies01}, which
are very important in real life applications \cite{Rector97}, and
with nominals. Thus, we also intend to compare the properties of
the different proposals for nominals in fuzzy DLs. Again, although
we expect that similar notions as in the classical \shoiq language
can be applied to \fkdshoiq, we need to investigate them in the
new setting and prove that they work. Furthermore, additional
research effort can be focused on the investigation of the
reasoning problem for the f-$\si$ and f-\shin languages, extended
with other norm operations. Regarding f-\shin, this might be a
very difficult problem since counting on number restrictions might
not be possible anymore.

\acks{This work is  supported by the FP6 Network of Excellence EU
project Knowledge Web (IST-2004-507482). Giorgos Stoilos, Giorgos
Stamou and Vassilis Tzouvaras were also partially funded by the
European Commission under FP6 Integrated Project X-Media
(FP6-026978).}

\vskip 0.2in
%\bibliography{FuzzyOWL}

\begin{thebibliography}{}

\bibitem[\protect\BCAY{Alejandro, Belle,\ \BBA\ Smith}{Alejandro
  et~al.}{2003}]{Alejandro03}
Alejandro, J., Belle, T., \BBA\ Smith, J. \BBOP2003\BBCP.
\newblock \BBOQ Modal keywords, ontologies and reasoning for video
  understanding\BBCQ\
\newblock In {\Bem Proceedings of the International Conference on Image and
  Video Retrieval}.

\bibitem[\protect\BCAY{Athanasiadis, Mylonas, Avrithis,\ \BBA\
  Kollias}{Athanasiadis et~al.}{2007}]{Athanasiadis07}
Athanasiadis, T., Mylonas, P., Avrithis, Y., \BBA\ Kollias, S.
\BBOP2007\BBCP.
\newblock \BBOQ Semantic image segmentation and object labeling\BBCQ\
\newblock {\Bem IEEE Transactions on Circuits and Systems for Video
  Technology}, {\Bem 17\/}(3), 298--312.

\bibitem[\protect\BCAY{Baader}{Baader}{1990}]{Baader90}
Baader, F. \BBOP1990\BBCP.
\newblock \BBOQ {Augmenting Concept Languages by Transitive Closure of Roles:
  An Alternative to Terminological Cycles}\BBCQ\
\newblock Research report\ RR-90-13.
\newblock An abridged version appeaered in Proc. IJCAI-91,pp.446-451.

\bibitem[\protect\BCAY{Baader, McGuinness, Nardi,\ \BBA\
  Patel-Schneider}{Baader et~al.}{2002a}]{Baader03a}
Baader, F., McGuinness, D., Nardi, D., \BBA\ Patel-Schneider, P.
  \BBOP2002a\BBCP.
\newblock {\Bem The Description Logic Handbook: Theory, implementation and
  applications}.
\newblock Cambridge University Press.

\bibitem[\protect\BCAY{Baader, Horrocks,\ \BBA\ Sattler}{Baader
  et~al.}{2002b}]{Baader02}
Baader, F., Horrocks, I., \BBA\ Sattler, U. \BBOP2002b\BBCP.
\newblock \BBOQ {Description Logics for the Semantic Web}\BBCQ\
\newblock {\Bem KI -- K{\"u}nstliche Intelligenz}, {\Bem 16\/}(4), 57--59.

\bibitem[\protect\BCAY{Bechhofer, van Harmelen, Hendler, Horrocks, McGuinness,
  Patel-Schneider,\ \BBA\ eds.}{Bechhofer et~al.}{2004}]{Bechhofer04}
Bechhofer, S., van Harmelen, F., Hendler, J., Horrocks, I.,
McGuinness, D.~L.,
  Patel-Schneider, P.~F., \BBA\ eds., L. A.~S. \BBOP2004\BBCP.
\newblock \BBOQ \uppercase{OWL} web ontology language reference\BBCQ\
\newblock \BTR.

\bibitem[\protect\BCAY{{Benitez}, {Smith},\ \BBA\ {Chang}}{{Benitez}
  et~al.}{2000}]{Benitez00}
{Benitez}, A.~B., {Smith}, J.~R., \BBA\ {Chang}, S.
\BBOP2000\BBCP.
\newblock \BBOQ {MediaNet: a multimedia information network for knowledge
  representation}\BBCQ\
\newblock In {\Bem Proc. SPIE Vol. 4210, p. 1-12, Internet Multimedia
  Management Systems, John R. Smith; Chinh Le; Sethuraman Panchanathan; C.-C.
  J. Kuo; Eds.}, \BPGS\ 1--12.

\bibitem[\protect\BCAY{Berners-Lee, Hendler,\ \BBA\ Lassila}{Berners-Lee
  et~al.}{2001}]{Tim01}
Berners-Lee, T., Hendler, J., \BBA\ Lassila, O. \BBOP2001\BBCP.
\newblock \BBOQ The semantic web\BBCQ\
\newblock {\Bem Scientific American}.

\bibitem[\protect\BCAY{Bobillo, Delgado,\ \BBA\ G\'omez-Romero}{Bobillo
  et~al.}{2006}]{Bobillo06}
Bobillo, F., Delgado, M., \BBA\ G\'omez-Romero, J. \BBOP2006\BBCP.
\newblock \BBOQ A crisp representation for fuzzy shoin with fuzzy nominals and
  general concept inclusions\BBCQ\
\newblock In {\Bem Proc. of the 2nd International Workshop on Uncertainty
  Reasoning for the Semantic Web (URSW 06), Athens, Georgia}.

\bibitem[\protect\BCAY{Bonatti\ \BBA\ Tettamanzi}{Bonatti\ \BBA\
  Tettamanzi}{2005}]{Bonatti03}
Bonatti, P.\BBACOMMA\  \BBA\ Tettamanzi, A. \BBOP2005\BBCP.
\newblock \BBOQ Some complexity results on fuzzy description logics\BBCQ\
\newblock In V.~{Di Ges\`u}, F.~Masulli, A.~P.\BED, {\Bem WILF 2003
  International Workshop on Fuzzy Logic and Applications}, LNCS~2955, Berlin.
  Springer Verlag.

\bibitem[\protect\BCAY{Calvanese, De~Giacomo, Lenzerini, Nardi,\ \BBA\
  Rosati}{Calvanese et~al.}{1998}]{Calvanese98}
Calvanese, D., De~Giacomo, G., Lenzerini, M., Nardi, D., \BBA\
Rosati, R.
  \BBOP1998\BBCP.
\newblock \BBOQ Description logic framework for information integration\BBCQ\
\newblock In {\Bem Proc. of the 6th Int. Conf. on the Principles of Knowledge
  Representation and Reasoning (KR'98)}, \BPGS\ 2--13.

\bibitem[\protect\BCAY{Chen, Fellah,\ \BBA\ Bishr}{Chen et~al.}{2005}]{Chen05}
Chen, H., Fellah, S., \BBA\ Bishr, Y. \BBOP2005\BBCP.
\newblock \BBOQ Rules for geospatial semantic web applications\BBCQ\
\newblock In {\Bem W3C Workshop on Rule Languages for Interoperability}.

\bibitem[\protect\BCAY{Ding\ \BBA\ Peng}{Ding\ \BBA\ Peng}{2004}]{Ding04}
Ding, Z.\BBACOMMA\  \BBA\ Peng, Y. \BBOP2004\BBCP.
\newblock \BBOQ {A Probabilistic Extension to Ontology Language OWL}\BBCQ\
\newblock In {\Bem Proceedings of the 37th Hawaii International Conference On
  System Sciences (HICSS-37).}, \BPG~10, Big Island, Hawaii.

\bibitem[\protect\BCAY{Dubois\ \BBA\ Prade}{Dubois\ \BBA\
  Prade}{2001}]{Dubois01}
Dubois, D.\BBACOMMA\  \BBA\ Prade, H. \BBOP2001\BBCP.
\newblock \BBOQ Possibility theory, probability theory and many-valued logics:
  A clarification\BBCQ\
\newblock {\Bem Ann. Math. Artif. Intell.}, {\Bem 32\/}(1-4), 35--66.

\bibitem[\protect\BCAY{Fagin}{Fagin}{1998}]{Fagin98}
Fagin, R. \BBOP1998\BBCP.
\newblock \BBOQ Fuzzy queries in multimedia database systems\BBCQ\
\newblock In {\Bem Proc. Seventeenth ACM Symp. on Principles of Database
  Systems}, \BPGS\ 1--10.

\bibitem[\protect\BCAY{Giugno\ \BBA\ Lukasiewicz}{Giugno\ \BBA\
  Lukasiewicz}{2002}]{Giugno02}
Giugno, R.\BBACOMMA\  \BBA\ Lukasiewicz, T. \BBOP2002\BBCP.
\newblock \BBOQ {P-SHOQ(D)}: A probabilistic extension of shoq(d) for
  probabilistic ontologies in the semantic web\BBCQ\
\newblock In {\Bem JELIA '02: Proceedings of the European Conference on Logics
  in Artificial Intelligence}, \BPGS\ 86--97, London, UK. Springer-Verlag.

\bibitem[\protect\BCAY{Hajek}{Hajek}{1998}]{Hajek98}
Hajek, P. \BBOP1998\BBCP.
\newblock {\Bem Metamathematics of fuzzy logic}.
\newblock Kluwer.

\bibitem[\protect\BCAY{Hajek}{Hajek}{2005}]{Hajeck05}
Hajek, P. \BBOP2005\BBCP.
\newblock \BBOQ Making fuzzy description logic more general\BBCQ\
\newblock {\Bem Fuzzy Sets and Systems}, {\Bem 154\/}(1), 1--15.

\bibitem[\protect\BCAY{Hajek}{Hajek}{2006}]{Hajek06}
Hajek, P. \BBOP2006\BBCP.
\newblock \BBOQ Computational complexity of $t$-norm based propositional fuzzy
  logics with rational truth constants\BBCQ\
\newblock {\Bem Fuzzy Sets and Systems}, {\Bem 157\/}(13), 677--682.

\bibitem[\protect\BCAY{H{\"o}lldobler, Khang,\ \BBA\ St{\"o}rr}{H{\"o}lldobler
  et~al.}{2002}]{Holldobler02}
H{\"o}lldobler, S., Khang, T.~D., \BBA\ St{\"o}rr, H.-P.
\BBOP2002\BBCP.
\newblock \BBOQ A fuzzy description logic with hedges as concept
  modifiers\BBCQ\
\newblock In {\Bem Proceedings InTech/VJFuzzy'2002}, \BPGS\ 25--34.

\bibitem[\protect\BCAY{H\"olldobler, Nga,\ \BBA\ Khang}{H\"olldobler
  et~al.}{2005}]{Holldobler05}
H\"olldobler, S., Nga, N.~H., \BBA\ Khang, T.~D. \BBOP2005\BBCP.
\newblock \BBOQ The fuzzy description logic $\alc_{FLH}$\BBCQ\
\newblock In {\Bem International workshop on Description Logics}. CEUR.

\bibitem[\protect\BCAY{Hollunder}{Hollunder}{1994}]{Hollunder94}
Hollunder, B. \BBOP1994\BBCP.
\newblock \BBOQ An alternative proof method for possibilistic logic and its
  application to terminological logics\BBCQ\
\newblock In {\Bem Proceedings of the 10th Annual Conference on Uncertainty in
  Artificial Intelligence (UAI-94)}, \BPGS\ 327--335, San Francisco, CA. Morgan
  Kaufmann Publishers.

\bibitem[\protect\BCAY{Hollunder, Nutt,\ \BBA\ Schmidt-Schaus}{Hollunder
  et~al.}{1990}]{Hollunder90}
Hollunder, B., Nutt, W., \BBA\ Schmidt-Schaus, M. \BBOP1990\BBCP.
\newblock \BBOQ Subsumption algorithms for concept description languages\BBCQ\
\newblock In {\Bem European Conference on Artificial Intelligence}, \BPGS\
  348--353.

\bibitem[\protect\BCAY{Horrocks, Patel-Schneider,\ \BBA\ van Harmelen}{Horrocks
  et~al.}{2003}]{Horrocks03c}
Horrocks, I., Patel-Schneider, P.~F., \BBA\ van Harmelen, F.
\BBOP2003\BBCP.
\newblock \BBOQ From \shiq and \uppercase{RDF} to \uppercase{OWL}: The making
  of a web ontology language\BBCQ\
\newblock {\Bem Web Semantics}, {\Bem 1}.

\bibitem[\protect\BCAY{Horrocks\ \BBA\ Sattler}{Horrocks\ \BBA\
  Sattler}{1999}]{Horrocks99}
Horrocks, I.\BBACOMMA\  \BBA\ Sattler, U. \BBOP1999\BBCP.
\newblock \BBOQ A description logic with transitive and inverse roles and role
  hierarchies\BBCQ\
\newblock {\Bem Journal of Logic and Computation}, {\Bem 9}, 385--410.

\bibitem[\protect\BCAY{Horrocks\ \BBA\ Sattler}{Horrocks\ \BBA\
  Sattler}{2005}]{Horrocks05a}
Horrocks, I.\BBACOMMA\  \BBA\ Sattler, U. \BBOP2005\BBCP.
\newblock \BBOQ A tableaux decision procedure for $\mathcal{SHOIQ}$\BBCQ\
\newblock In {\Bem Proc. 19th Int.\ Joint Conf.\ on Artificial Intelligence
  (IJCAI~05)}.

\bibitem[\protect\BCAY{Horrocks, Sattler,\ \BBA\ Tobies}{Horrocks
  et~al.}{1999}]{Horrocks99j}
Horrocks, I., Sattler, U., \BBA\ Tobies, S. \BBOP1999\BBCP.
\newblock \BBOQ Practical reasoning for expressive description logics\BBCQ\
\newblock In {\Bem Proceedings of the 6th International Conference on Logic for
  Programming and Automated Reasoning {(LPAR'99)}}, \lowercase{\BNUM}\ 1705 in
  LNAI, \BPGS\ 161--180. Springer-Verlag.

\bibitem[\protect\BCAY{Horrocks, Sattler,\ \BBA\ Tobies}{Horrocks
  et~al.}{2000}]{Horrocks00}
Horrocks, I., Sattler, U., \BBA\ Tobies, S. \BBOP2000\BBCP.
\newblock \BBOQ {Reasoning with Individuals for the Description Logic}
  \shiq\BBCQ\
\newblock In MacAllester, D.\BED, {\Bem CADE-2000}, \lowercase{\BNUM}\ 1831 in
  LNAI, \BPGS\ 482--496. Springer-Verlag.

\bibitem[\protect\BCAY{Kandel}{Kandel}{1982}]{Kandel82}
Kandel, A. \BBOP1982\BBCP.
\newblock {\Bem A Fuzzy Techniques in Pattern Recognition}.
\newblock Wiley.

\bibitem[\protect\BCAY{Klement, Mesiar,\ \BBA\ Pap}{Klement
  et~al.}{2004}]{Klement04a}
Klement, E.~P., Mesiar, R., \BBA\ Pap, E. \BBOP2004\BBCP.
\newblock \BBOQ Triangular norms. position paper {I}: basic analytical and
  algebraic properties\BBCQ\
\newblock {\Bem Fuzzy Sets and Systems}, {\Bem 143}, 5--26.

\bibitem[\protect\BCAY{Klir\ \BBA\ Yuan}{Klir\ \BBA\ Yuan}{1995}]{Klir95}
Klir, G.~J.\BBACOMMA\  \BBA\ Yuan, B. \BBOP1995\BBCP.
\newblock {\Bem Fuzzy Sets and Fuzzy Logic: Theory and Applications}.
\newblock Prentice-Hall.

\bibitem[\protect\BCAY{Koller, Levy,\ \BBA\ Pfeffer}{Koller
  et~al.}{1997}]{Koller97}
Koller, D., Levy, A., \BBA\ Pfeffer, A. \BBOP1997\BBCP.
\newblock \BBOQ {{P-CLASSIC}: A tractable probabilistic Description
  Logic}\BBCQ\
\newblock In {\Bem Proceedings of the 14th National Conference on Artificial
  Intelligence (AAAI-97).}, \BPGS\ 390--397.

\bibitem[\protect\BCAY{Krishnapuram\ \BBA\ Keller}{Krishnapuram\ \BBA\
  Keller}{1992}]{Krishnapuram92}
Krishnapuram, R.\BBACOMMA\  \BBA\ Keller, J. \BBOP1992\BBCP.
\newblock \BBOQ Fuzzy set theoretic approach to computer vision: An
  overview\BBCQ\
\newblock In {\Bem IEEE International Conference on Fuzzy Systems}, \BPGS\
  135--142.

\bibitem[\protect\BCAY{Larsen\ \BBA\ Yager}{Larsen\ \BBA\
  Yager}{1993}]{Larsen93}
Larsen, H.\BBACOMMA\  \BBA\ Yager, R. \BBOP1993\BBCP.
\newblock \BBOQ The use of fuzzy relational thesauri for classificatory problem
  solving in information restrieval and exprert systems\BBCQ\
\newblock {\Bem IEEE Trans. in System, Man, and Cybernetics}, {\Bem 23\/}(1),
  31--41.

\bibitem[\protect\BCAY{Li, Xu, Lu,\ \BBA\ Kang}{Li et~al.}{2006a}]{Li06a}
Li, Y., Xu, B., Lu, J., \BBA\ Kang, D. \BBOP2006a\BBCP.
\newblock \BBOQ Discrete tableau algorithms for $\mathcal{FSHI}$\BBCQ\
\newblock In {\Bem Proceedings of the International Workshop on Description
  Logics (DL 2006), Lake District, UK}.

\bibitem[\protect\BCAY{Li, Xu, Lu,\ \BBA\ Kang}{Li et~al.}{2006b}]{Li06b}
Li, Y., Xu, B., Lu, J., \BBA\ Kang, D. \BBOP2006b\BBCP.
\newblock \BBOQ Reasoning technique for extended fuzzy \alcq\BBCQ\
\newblock In {\Bem ICCSA (2)}, \BPGS\ 1179--1188.

\bibitem[\protect\BCAY{Li, Xu, Lu, Kang,\ \BBA\ Wang}{Li et~al.}{2005}]{Li05}
Li, Y., Xu, B., Lu, J., Kang, D., \BBA\ Wang, P. \BBOP2005\BBCP.
\newblock \BBOQ Extended fuzzy description logic \alcn\BBCQ\
\newblock In {\Bem Proceedings of the 9th International Conference on Knowledge
  Based Intelligent Information and EngineeringSystems (KES-05)}, \BPGS\
  896--902.

\bibitem[\protect\BCAY{Liu, Tian,\ \BBA\ Ma}{Liu et~al.}{1994}]{Liu04}
Liu, O., Tian, Q., \BBA\ Ma, J. \BBOP1994\BBCP.
\newblock \BBOQ A fuzzy description logic approach to model management in r\&d
  project selection\BBCQ\
\newblock In {\Bem Proceedings of the 8th Pacific Asian Conference on
  Information Systems (PACIS-04)}.

\bibitem[\protect\BCAY{McGuiness}{McGuiness}{2003}]{McGuiness03a}
McGuiness, D. \BBOP2003\BBCP.
\newblock \BBOQ Configuration\BBCQ\
\newblock In Baader, F., Calvanese, D., McGuinness, D., Nardi, D., \BBA\
  Patel-Schneider, P.~F.\BEDS, {\Bem The Description Logic Handbook: Theory,
  Implementation, and Applications}, \BPGS\ 388--405. Cambridge University
  Press.

\bibitem[\protect\BCAY{Meghini, Sebastiani,\ \BBA\ Straccia}{Meghini
  et~al.}{2001}]{Meghini01}
Meghini, C., Sebastiani, F., \BBA\ Straccia, U. \BBOP2001\BBCP.
\newblock \BBOQ A model of multimedia information retrieval\BBCQ\
\newblock {\Bem Journal of the ACM}, {\Bem 48\/}(5), 909--970.

\bibitem[\protect\BCAY{Mostert\ \BBA\ Shields}{Mostert\ \BBA\
  Shields}{1957}]{Mostert57}
Mostert, P.\BBACOMMA\  \BBA\ Shields, A. \BBOP1957\BBCP.
\newblock \BBOQ On the structure of semigroups on a compact manifold with
  boundary\BBCQ\
\newblock {\Bem The Annals of Mathematics}, {\Bem 65\/}(1), 117--143.

\bibitem[\protect\BCAY{Navara}{Navara}{2000}]{Navara00}
Navara, M. \BBOP2000\BBCP.
\newblock \BBOQ Satisfiability in fuzzy logic\BBCQ\
\newblock {\Bem Neural Network World}, {\Bem 10\/}(5), 845--858.

\bibitem[\protect\BCAY{Nebel}{Nebel}{1990}]{Nebel90}
Nebel, B. \BBOP1990\BBCP.
\newblock \BBOQ Terminological reasoning is inherently intractable\BBCQ\
\newblock {\Bem Journal of Artificial Intelligence}, {\Bem 43}, 235--249.

\bibitem[\protect\BCAY{Oguntade\ \BBA\ Beaumont}{Oguntade\ \BBA\
  Beaumont}{1982}]{Oguntade82}
Oguntade, O.\BBACOMMA\  \BBA\ Beaumont, P. \BBOP1982\BBCP.
\newblock \BBOQ Ophthalmological prognosis via fuzzy subsets\BBCQ\
\newblock {\Bem Fuzzy Sets and Systems}, {\Bem 7\/}(2), 123--179.

\bibitem[\protect\BCAY{Pan}{Pan}{2004}]{Pan04}
Pan, J.~Z. \BBOP2004\BBCP.
\newblock {\Bem Description Logics: Reasoning Support for the Semantic Web}.
\newblock Ph.D.\ thesis, School of Computer Science, The University of
  Manchester, Oxford Rd, Manchester M13 9PL, UK.

\bibitem[\protect\BCAY{Rector\ \BBA\ Horrocks}{Rector\ \BBA\
  Horrocks}{1997}]{Rector97}
Rector, A.~L.\BBACOMMA\  \BBA\ Horrocks, I. \BBOP1997\BBCP.
\newblock \BBOQ Experience building a large, re-usable medical ontology using a
  description logic with transitivity and concept inclusions\BBCQ\
\newblock In {\Bem Proceedings Workshop on Ontological Engineering, AAAI Spring
  Symposium, Stanford CA.}, \BPGS\ 100--107. Hanley and Belfus, Inc.,
  Philadelphia, PA.

\bibitem[\protect\BCAY{S\'anchez\ \BBA\ Tettamanzi}{S\'anchez\ \BBA\
  Tettamanzi}{2004}]{Sanchez04}
S\'anchez, D.\BBACOMMA\  \BBA\ Tettamanzi, A. \BBOP2004\BBCP.
\newblock \BBOQ Generalizing quantification in fuzzy description logic\BBCQ\
\newblock In {\Bem Proceedings 8th Fuzzy Days in Dortmund}.

\bibitem[\protect\BCAY{Sanchez\ \BBA\ Tettamanzi}{Sanchez\ \BBA\
  Tettamanzi}{2006}]{Sanchez06}
Sanchez, D.\BBACOMMA\  \BBA\ Tettamanzi, A.~A. \BBOP2006\BBCP.
\newblock \BBOQ Fuzzy quantification in fuzzy description logics\BBCQ\
\newblock In Sanchez, E.\BED, {\Bem Capturing Intelligence: Fuzzy Logic and the
  Semantic Web}. Elsevier.

\bibitem[\protect\BCAY{Sattler}{Sattler}{1996}]{Sattler96}
Sattler, U. \BBOP1996\BBCP.
\newblock \BBOQ A concept language extended with different kinds of transitive
  roles\BBCQ\
\newblock In {\Bem KI '96: Proceedings of the 20th Annual German Conference on
  Artificial Intelligence}, \BPGS\ 333--345. Springer-Verlag.

\bibitem[\protect\BCAY{Stoilos, Stamou, Tzouvaras, Pan,\ \BBA\
  Horrocks}{Stoilos et~al.}{2005a}]{Stoilos05c}
Stoilos, G., Stamou, G., Tzouvaras, V., Pan, J., \BBA\ Horrocks,
I.
  \BBOP2005a\BBCP.
\newblock \BBOQ The fuzzy description logic f-\shin\BBCQ\
\newblock In {\Bem Proceedings of the International Workshop on Uncertainty
  Reasoning for the Semantic Web}.

\bibitem[\protect\BCAY{Stoilos, Stamou, Tzouvaras, Pan,\ \BBA\
  Horrocks}{Stoilos et~al.}{2005b}]{Stoilos05}
Stoilos, G., Stamou, G., Tzouvaras, V., Pan, J., \BBA\ Horrocks,
I.
  \BBOP2005b\BBCP.
\newblock \BBOQ A fuzzy description logic for multimedia knowledge
  representation\BBCQ\
\newblock In {\Bem Proc. of the International Workshop on Multimedia and the
  Semantic Web}.

\bibitem[\protect\BCAY{Stoilos, Stamou, Tzouvaras, Pan,\ \BBA\
  Horrocks}{Stoilos et~al.}{2005c}]{Stoilos05d}
Stoilos, G., Stamou, G., Tzouvaras, V., Pan, J., \BBA\ Horrocks,
I.
  \BBOP2005c\BBCP.
\newblock \BBOQ Fuzzy \uppercase{OWL}: Uncertainty and the semantic web\BBCQ\
\newblock In {\Bem Proc. of the International Workshop on \uppercase{OWL}:
  Experiences and Directions}.

\bibitem[\protect\BCAY{Stoilos, Straccia, Stamou,\ \BBA\ Pan}{Stoilos
  et~al.}{2006}]{Stoilos06ate}
Stoilos, G., Straccia, U., Stamou, G., \BBA\ Pan, J.
\BBOP2006\BBCP.
\newblock \BBOQ General concept inclusions in fuzzy description logics\BBCQ\
\newblock In {\Bem Proceedings of the 17th International Conference on
  Artificial Intelligence (ECAI 06)}, \BPGS\ 457--461. IOS Press.

\bibitem[\protect\BCAY{Straccia}{Straccia}{2001}]{Straccia01}
Straccia, U. \BBOP2001\BBCP.
\newblock \BBOQ Reasoning within fuzzy description logics\BBCQ\
\newblock {\Bem Journal of Artificial Intelligence Research}, {\Bem 14},
  137--166.

\bibitem[\protect\BCAY{Straccia}{Straccia}{2005a}]{Straccia05d}
Straccia, U. \BBOP2005a\BBCP.
\newblock \BBOQ Description logics with fuzzy concrete domains\BBCQ\
\newblock In {\Bem 21st Conf. on Uncertainty in Artificial Intelligence
  (UAI-05)}, Edinburgh.

\bibitem[\protect\BCAY{Straccia}{Straccia}{2005b}]{Straccia05}
Straccia, U. \BBOP2005b\BBCP.
\newblock \BBOQ Towards a fuzzy description logic for the semantic web\BBCQ\
\newblock In {\Bem Proceedings of the 2nd European Semantic Web Conference}.

\bibitem[\protect\BCAY{Straccia}{Straccia}{1998}]{Straccia98}
Straccia, U. \BBOP1998\BBCP.
\newblock \BBOQ A fuzzy description logic\BBCQ\
\newblock In {\Bem AAAI '98/IAAI '98: Proceedings of the fifteenth
  national/tenth conference on Artificial intelligence/Innovative applications
  of artificial intelligence}, \BPGS\ 594--599. American Association for
  Artificial Intelligence.

\bibitem[\protect\BCAY{Straccia}{Straccia}{2004}]{Straccia04}
Straccia, U. \BBOP2004\BBCP.
\newblock \BBOQ Transforming fuzzy description logics into classical
  description logics\BBCQ\
\newblock In {\Bem Proceedings of the 9th European Conference on Logics in
  Artificial Intelligence (JELIA-04)}, \lowercase{\BNUM}\ 3229 in Lecture Notes
  in Computer Science, \BPGS\ 385--399, Lisbon, Portugal. Springer Verlag.

\bibitem[\protect\BCAY{Sugeno}{Sugeno}{1985}]{Sugeno85}
Sugeno, M. \BBOP1985\BBCP.
\newblock {\Bem Industrial Applications of Fuzzy Control}.
\newblock North-Holland.

\bibitem[\protect\BCAY{Tarski}{Tarski}{1956}]{Tars56}
Tarski, A. \BBOP1956\BBCP.
\newblock {\Bem {Logic, Semantics, Metamathemetics: Papers from 1923 to 1938}}.
\newblock Oxford University Press.

\bibitem[\protect\BCAY{Tobies}{Tobies}{2001}]{Tobies01}
Tobies, S. \BBOP2001\BBCP.
\newblock {\Bem {Complexity Results and Practical Algorithms for Logics in
  Knowledge Representation}}.
\newblock Ph.D.\ thesis, Rheinisch-Westf\"alischen Technischen Hochschule
  Aachen.
\newblock {URL http://lat.inf.tu-dresden.de/research/phd/Tobies-PhD-2001.pdf }.

\bibitem[\protect\BCAY{Tresp\ \BBA\ Molitor}{Tresp\ \BBA\
  Molitor}{1998}]{Tresp98}
Tresp, C.\BBACOMMA\  \BBA\ Molitor, R. \BBOP1998\BBCP.
\newblock \BBOQ A description logic for vague knowledge\BBCQ\
\newblock In {\Bem In proc of the 13th European Conf. on Artificial
  Intelligence (ECAI-98)}.

\bibitem[\protect\BCAY{Vardi}{Vardi}{1997}]{Vardi97}
Vardi, M.~Y. \BBOP1997\BBCP.
\newblock \BBOQ Why is modal logic so robustly decidable?\BBCQ\
\newblock In {\Bem DIMACS Series in Discrete Mathematics and Theoretical
  Computer Science}, \BPGS\ 149--184.

\bibitem[\protect\BCAY{Yen}{Yen}{1991}]{Yen91}
Yen, J. \BBOP1991\BBCP.
\newblock \BBOQ Generalising term subsumption languages to fuzzy logic\BBCQ\
\newblock In {\Bem In Proc of the 12th Int. Joint Conf on Artificial
  Intelligence (IJCAI-91)}, \BPGS\ 472--477.

\bibitem[\protect\BCAY{Zadeh}{Zadeh}{1965}]{Zadeh65}
Zadeh, L.~A. \BBOP1965\BBCP.
\newblock \BBOQ Fuzzy sets\BBCQ\
\newblock {\Bem Information and Control}, {\Bem 8}, 338--353.

\bibitem[\protect\BCAY{Zimmermann}{Zimmermann}{1987}]{Zimmermann87}
Zimmermann, H. \BBOP1987\BBCP.
\newblock {\Bem Fuzzy Sets, Decision Making, and Expert Systems}.
\newblock Kluwer, Boston.

\end{thebibliography}
%\bibliographystyle{theapa}

\end{document}